\documentclass[10pt,twocolumn,letterpaper]{article}

\usepackage[pagenumbers]{cvpr} 

\usepackage{booktabs}
\usepackage{multirow}
\usepackage{graphicx}
\usepackage{amssymb}
\usepackage{multicol}
\usepackage{makecell}
\usepackage{colortbl}
\usepackage[table]{xcolor}
\usepackage{array}

\usepackage{pifont}
\usepackage{xcolor}
\usepackage{graphicx}
\usepackage{url}
\usepackage{cuted}
\usepackage[titletoc]{appendix}
\usepackage{url}
\usepackage{tabularx}
\usepackage{booktabs}
\usepackage{titletoc}
\usepackage[T1]{fontenc}

\usepackage[numbers,sort&compress]{natbib}

\newcommand{\bmark}{%
    \textcolor{blue!65!black}{
        {\ding{52}\rotatebox[origin=c]{-9.2}{\kern-0.7em\ding{55}}}%
    }%
}
\newcommand{\xmark}{%
    \textcolor{red!80!black}{\ding{55}}%
}
\newcommand{\cmark}{%
    \textcolor{green!50!black}{\ding{51}}%
}

\definecolor{cvprblue}{rgb}{0.21,0.49,0.74}
\usepackage[pagebackref,breaklinks,colorlinks,allcolors=cvprblue]{hyperref}

\def\confName{CVPR}
\def\confYear{2027}

\title{MemRR: Benchmarking Cross-Modal Referential Reasoning in Long-Term Multimodal Memory}

\author{
  \textbf{Kaiyan Lei}\textsuperscript{1,3},\hspace{0.5mm}
  \textbf{Jiayi Wang}\textsuperscript{1}, \hspace{0.5mm}
  \textbf{Yuxuan chen}\textsuperscript{2,3}, \hspace{0.5mm}
  \textbf{Kun Shao}\textsuperscript{3$^\dag$}, \hspace{0.5mm}
  \textbf{Xule Liu}\textsuperscript{3}, \\
  \textbf{Chao Li}\textsuperscript{3}, \hspace{0.5mm}
  \textbf{Jian Luan}\textsuperscript{3}, \hspace{0.5mm}
  \textbf{Xu-Yao Zhang}\textsuperscript{1$^\dag$}
  \\
  \textsuperscript{1}Institute of automation, Chinese academy of science,
  \textsuperscript{2}the University of Hong Kong \\
  \textsuperscript{3}Xiaomi Corporation\\ \\
}

\begin{document}
\maketitle



\newcommand\blfootnote[1]{%
  \begingroup
  \renewcommand\thefootnote{}\footnote{#1}%
  \addtocounter{footnote}{-1}%
  \endgroup
}
\blfootnote{$^\dagger$Corresponding author.}

\begin{abstract}
With the advancement of multimodal agents, existing benchmarks primarily focus on fact retrieval and aggregation, overlooking whether agents can maintain persistent associations among objects or events, and resolve evolving references across long-term interactions. 
To address this gap, we introduce MemRR, a new benchmark for cross-modal referential reasoning in long-term multimodal memory, comprising 10 scenarios and 2,271 QA pairs.
MemRR extends the fact retrieval paradigm toward cross-modal entity association and dynamic semantic reasoning, requiring agents to resolve references and establish persistent associations across referents. 
It has three dimensions: reference resolution, referential understanding and semantic inference, and seven subcategories, assessing the abilities to identify the referents, understand relationships and 
reason over explicit or implicit visual evidence and cross-turn context.
In benchmarking evaluations, the best-performing MLLM achieves a Judge score of about 0.6, revealing substantial challenges in cross-session evidence integration and referential reasoning. Memory agents struggle with referential reasoning tasks, with overall scores below 0.3, highlighting that the limitations of similarity-based retrieval and demonstrating that multimodal information retention is insufficient for effective referential reasoning. These findings motivate structural multimodal memory with long-context referential reasoning. Datasets and code are released at \url{https://anonymous.4open.science/r/MemRR_benchmarking-0E78/}.
\end{abstract}

\section{Introduction}

\begin{table*}[t]
    \centering
    \begin{minipage}[c]{0.63\textwidth}
    \centering
    \captionsetup{position=above}
    \captionof{table}{Comparison between MemRR with existing conversational benchmarks. \cmark: Satisfies; \xmark: Dose not satisfy; \bmark: Insufficient. For details of each dimension, see \S\ref{3.4}. }
    \resizebox{\textwidth}{!}{
\begin{tabular}{lcccccc|ccccccc}
\toprule
\multirow{2}{*}{Benchmarks}
& \multicolumn{6}{c|}{Data Format}
& \multicolumn{3}{c|}{Reference Resolusion} & \multicolumn{2}{c|}{Referential understaning} & \multicolumn{2}{c}{Semantic Inference} \\

\cmidrule(lr){2-7}
\cmidrule(lr){8-14}

& Year
& \makecell{Multi\\-modal}
& \makecell{Multi\\-Sess.}
& \makecell{Max L}
& \makecell{A.Turns}
& \makecell{A.Imgs}


& \makecell{RR-IR}
& \makecell{RR-ER}
& \makecell{RR-SR}
& \makecell{RU-RU}
& \makecell{RU-TCU}
& \makecell{SI-BI}
& \makecell{SI-PI} \\

\midrule

MemoryBank\cite{zhong2024memorybank}
& 2023 & \xmark & \cmark & -
& 3.77 & -
& \xmark & \xmark & \xmark
& \xmark & \xmark & \xmark & \xmark \\

LoCoMo\cite{maharana2024evaluating}
& 2024 & \cmark & \cmark & 10K
& 10.81 & 88.1
& \bmark & \cmark & \cmark
& \xmark & \bmark & \xmark & \xmark \\

LongMemEval\cite{wu2024longmemeval}
& 2025 & \xmark & \cmark & 1.5M
& 5.19 & -
& \bmark & \cmark & \bmark
& \cmark & \bmark & \xmark & \bmark \\

MemAgentBench\cite{hu2025evaluating}
& 2025 & \xmark & \cmark & 115K
& 9.55 & -
& \xmark & \xmark & \xmark
& \xmark & \xmark & \xmark & \xmark \\

MMDialog\cite{feng2023mmdialog}
& 2023 & \cmark & \xmark & -
& 4.6 & 2.59
& \xmark & \xmark & \xmark
& \bmark & \bmark & \xmark & \xmark \\

Mem-Gallery\cite{bei2026mem}
& 2026 & \cmark & \cmark & 80K
& 16.5 & 50.2
& \cmark & \bmark & \bmark
& \bmark & \bmark & \xmark & \xmark \\

MemEye\cite{guo2026memeye}
& 2026 & \cmark & \cmark & 71K
& 3.5 & 54.75
& \bmark & \xmark & \cmark
& \cmark & \xmark & \xmark & \xmark \\

MEMLENS\cite{ren2026memlens}
& 2026 & \cmark & \cmark & 256K
& 4.9 & 95.15
& \cmark & \bmark & \xmark
& \cmark & \bmark & \xmark & \xmark \\

VISUALMEM\cite{nguyen2026personal}
& 2026 & \cmark & \xmark & 131K
& 5.92 & 172.3
& \bmark & \xmark & \bmark
& \xmark & \xmark & \xmark & \xmark \\

\midrule

MemRR
& 2026 & \cmark & \cmark & 1.1M
& 22.8 & 111.3
& \cmark & \cmark & \cmark
& \cmark & \cmark & \cmark & \cmark \\

\bottomrule
\end{tabular}
}
    \label{tab_1}
\end{minipage}
\hfill
\begin{minipage}[c]{0.36\textwidth}
    \centering
    \includegraphics[width=\linewidth]{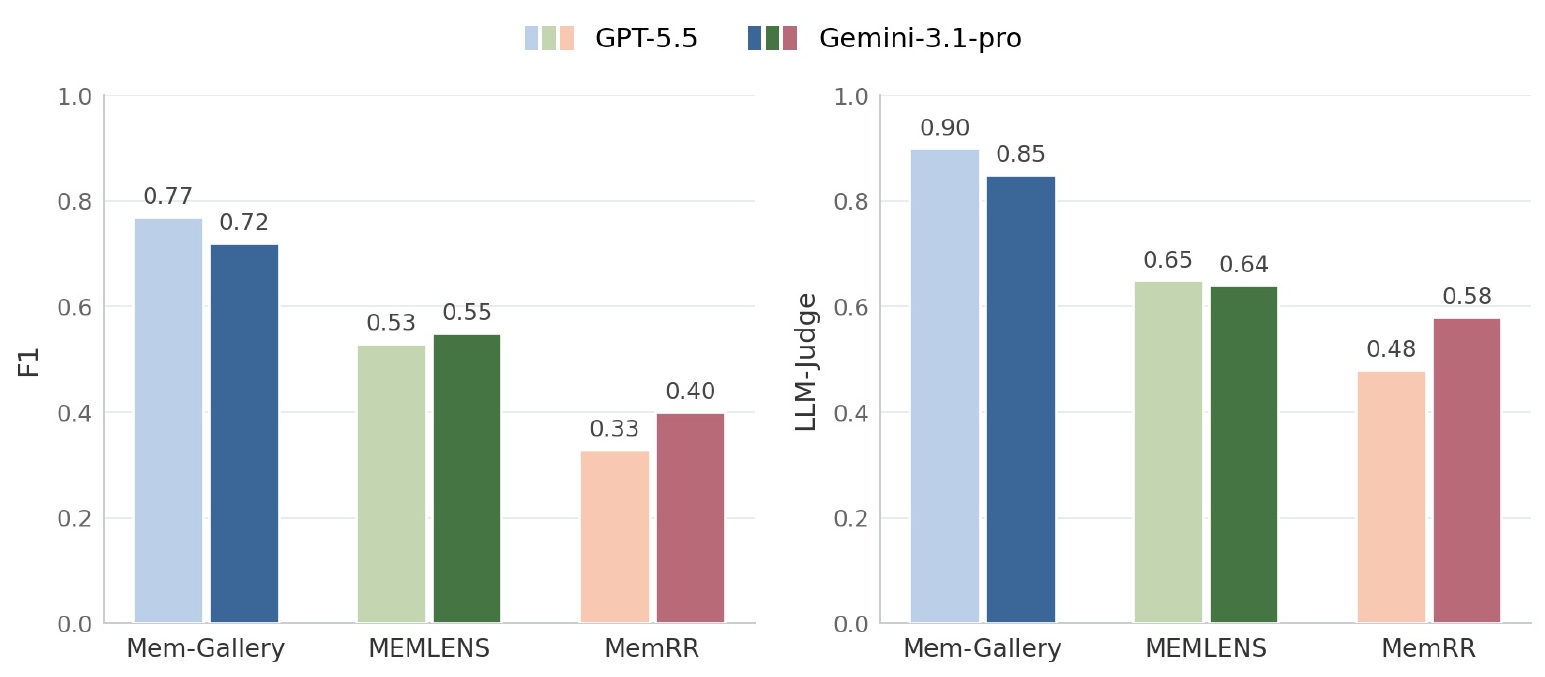}
    \captionof{figure}{Overall task difficulty of MemRR compared with Mem-Gallery \cite{bei2026mem} and MEMLENS \cite{ren2026memlens}, by evaluating MLLM's performance under full-context setting on each dataset. }
    \label{fig2}
\end{minipage}
\end{table*}

With the advancement of multimodal large language models (MLLMs), enabling agents to retain and leverage information from long-term history has become essential for personalized and persistent interactive systems \cite{guo2026seeing,hu2025memory,wu2025human,zhang2025survey}.
To evaluate this capability, recent studies \cite{ahmed2026multi} have introduced various long-term conversational memory benchmarks, advancing research on memory retrieval and reasoning. However, existing evaluation paradigms predominantly focus on retrieving and aggregating historical facts, representing memory as factual knowledge or dialogue snippets. Consequently, long-term interactions are largely modeled as the accumulation of discrete historical information.

Real-world multimodal interactions, however, involve more than factual accumulation. Users frequently refer to previously encountered visual entities, scenes, and events through diverse expressions, including visual attributes, spatial relationships, evolving semantic roles, etc. 
For example, a user may revisit a store discussed in an earlier session by referring to it as ``the store with the green sign," although its visual details were never explicitly described, and also purchase a sandwich. Moreover, the attributes or semantic roles evolve dynamically over time. The same store may later be referenced as ``the place where I bought a sandwich last month.'' Resolving such references requires recovering implicit visual evidence and maintaining persistent associations among entities and events, rather than relying solely on semantic similarity or keyword matching.
Thus, \textit{\textbf{multimodal memory must extend beyond historical fact retrieval to support persistent cross-modal reference modeling and reference-grounded semantic reasoning.}} As summarized in Table \ref{tab_1}, these capabilities remain insufficiently addressed by existing benchmarks.

A further limitation lies in the shallow utilization of visual modality. As shown in Fig.\ref{fig1}, existing benchmarks typically employ images as visual scenarios, global semantic context, or object exhibition, with limited modeling of instance-level referring correspondences between visual entities and expressions. Consequently, some tasks remain solvable through image captions or keywords, without requiring persistent visual entity tracking or fine-grained cross-modal reference resolution. \textit{\textbf{The inclusion of visual modalities alone does not entail substantial cross-modal memory and reasoning, motivating benchmarks that explicitly evaluate persistent entity associations and resolve cross-modal references across long-term interactions.}}

To close this gap, we introduce \textbf{MemRR}, a new benchmark for cross-modal referential reasoning in long-term multimodal memory. MemRR extends the fact retrieval paradigm toward cross-modal entity association and dynamic semantic reasoning. Cross-turn references include explicit or implicit visual attributes, relationships, or semantic roles, covering scenes, events, existing or implicit entities. Referents establish associations and semantic evolution temporally through diverse referring patterns. For example, a plant may be successively referred to across sessions as ``this blue pot"$\rightarrow$``the one I just bought"$\rightarrow$``the one on the balcony"$\rightarrow$``the one I pruned." Despite variations in expression, the underlying entity remains consistent.

To evaluate referential reasoning in long-term multimodal interactions, we define three complementary levels (\S\ref{3.4}), which are interdependent yet distinguishable: \textbf{Reference Resolution}, which assess the ability to identify referents; \textbf{Referential Understanding}, which further assess the ability to understand relationships and temporal evolution among referents; \textbf{Semantic Inference}, which assess the ability to infer user intent and behavior from referring expressions. 
It comprises 10 scenarios and 2,271 QA pairs, with the length of about 1.1M tokens per scenario. We further compare MemRR with two existing benchmarks by evaluating GPT-5.5 and Gemini-3.1-Pro under full-context settings using F1 and LLM-Judge metrics (Fig. \ref{fig2}). Results demonstrate that MemRR presents greater reasoning challenges, highlighting the demands of persistent cross-modal association and long-term semantic reasoning.

We conduct extensive evaluations of 7 frontier MLLMs and 10 memory agents, yielding three key findings. (1) Frontier MLLMs under full-context settings outperform memory agents, despite their greater context requirements, exposing limitations in existing agents on long-term referential reasoning. 
(2) Cross-session evidence integration and reasoning remains challenging even for advanced MLLMs, particularly in modeling complex semantic and association, tracking evolving semantics.
(3) Effective referential reasoning imposes demands on structured memory organization beyond similarity-based retrieval. Multimodal modality retention alone does not yield reasoning advantages.
These findings highlight the need for structured multimodal memory and reference-aware reasoning. 
Overall, our main contributions are follows:


\begin{itemize}
    \item Novel Perspective. We extend fact retrieval paradigm widely used in multimodal memory to cross-modal referential reasoning, highlighting the need to maintain evolving associations among visual entities and events.

    \item Benchmark Construction. We introduce MemRR, comprising 10 scenarios, 2,271 QA pairs, with three main evaluation levels: reference resolution, referential understanding, and semantic inference.

    \item Benchmarking \& Findings. we reveal key advantages and
limitations of existing methods based on extensive benchmarking and in-depth analysis, revealing actionable insights for future research.
\end{itemize}

\section{Related Works}

\subsection{Long-term Memory Benchmarks}

With the development of large language models (LLMs) toward persistent interactive agents, more works begin to evaluate the memory ability to retain, retrieve, and reason.
Early studies mainly focused on text-based memory \cite{feng2021multidoc2dial,bai2024longbench,zhang2024bench,zhong2024memorybank,du2026memguide}, while subsequent multimodal benchmarks extend evaluation to lengthy documents \cite{cao2024mmlongbench,wang2025mmlongbench,he2026memoryarena}, multi-images \cite{xue2025mmrc,ye2024mplug,liu2024mmdu}, and long videos \cite{song2024moviechat,wu2024longvideobench,zhou2025mlvu}, without explicitly modeling multi-session memory.
Among dialogue-centric benchmarks, some works focus on text retrieval \cite{wu2024longmemeval,bian2026realmem} and update \cite{hu2025evaluating}. LoCoMo \cite{maharana2024evaluating} introduces images, but still primarily emphasizes text evidence rather than fine-grained visual perception.
Recent multimodal benchmarks, including Mem-Gallery \cite{bei2026mem} and MEMLENS \cite{ren2026memlens}, extend evaluation to multi-session visual-textual interactions, focusing on image-text retrieval, with images predominantly serve as visual context or entity depictions. MemEye \cite{guo2026memeye} and VISUALMEM \cite{nguyen2026personal} focus mainly on visual memory extraction. Despite these advances, existing benchmarks provide limited coverage of constructing cross-turn referential chains and maintaining the evolution of semantic roles to evaluate cross-modal reasoning.

\subsection{Memory-Augmented LLM Agents}

Memory agents primarily rely on external memory mechanisms to persistently store, organize, and leverage historical interactions. Text-only agents have evolved from dynamical extraction \cite{chhikara2025mem0,packer2023memgpt}, agentic organization \cite{xu2025mem,yu2025memagent}, graph structure \cite{jain2025augustus,fisher2025neural,gu2024hipporag,li2026graphmemix} and hierarchical management \cite{kang2025memory,du2025rethinking} toward semantic compression and reasoning-guided retrieval \cite{liu2026simplemem,ji2026memory}.
Recent multimodal works have further incorporated visual and other heterogeneous information into long-term memory. These approaches enhance multimodal memory storage and utilization through modular memory management \cite{wang2025mirix,liu2025memverse}, agentic hybric systems \cite{feng2026m2a,sun2026pmmc}, multimodal atomic unit \cite{liu2026omni}, cross-modal retrieval optimization \cite{kang2026v,tian2026beyond,yan2026adamem}, heterogeneous information retrieval \cite{yeo2026universalrag,guo2025rag}, multiple temporal scales \cite{yeo2026worldmm}. Beyond external memory design, emerging works \cite{bo2025agentic,long2025seeing,du2025memory,nie2026personavlm} investigates training-based approaches to strengthen memory formation and reasoning, through reinforcement learning to optimize multimodal episodic and semantic memory.

\section{The MemRR Benchmark}

\subsection{Task Formulation}

MemRR evaluates an agent's ability to utilize multimodal memories from long-term interactions. Each sample comprises a chronologically ordered, multi-session dialogue history $D=\{S_1,\ldots,S_N\}$, a set of queries $\hat{Q}=\{\hat{q_1}, \ldots,\hat{q_n}\}$, and their corresponding ground-truth answers $\hat{A}=\{\hat{a_1}, \ldots,\hat{a_n}\}$. Each session simulates a sequence of events, with dialogue turns capturing continuous daily activities and knowledge-seeking interactions. Given the $D$ and $\hat{q_i}$, an agent produces an answer $a_i=f(D,\hat{q_i})$, which is evaluated against $\hat{a_i}$. Each query is further annotated with a set of evidence sessions $E(q_i)=\{S_{i_1},\ldots,S_{i_k}\}$ that contain the information necessary for answering. Queries requires resolving references, and reasoning over explicit or implicit visual evidence and cross-turn context.

\begin{figure*}
    \centering
    \includegraphics[width=\linewidth]{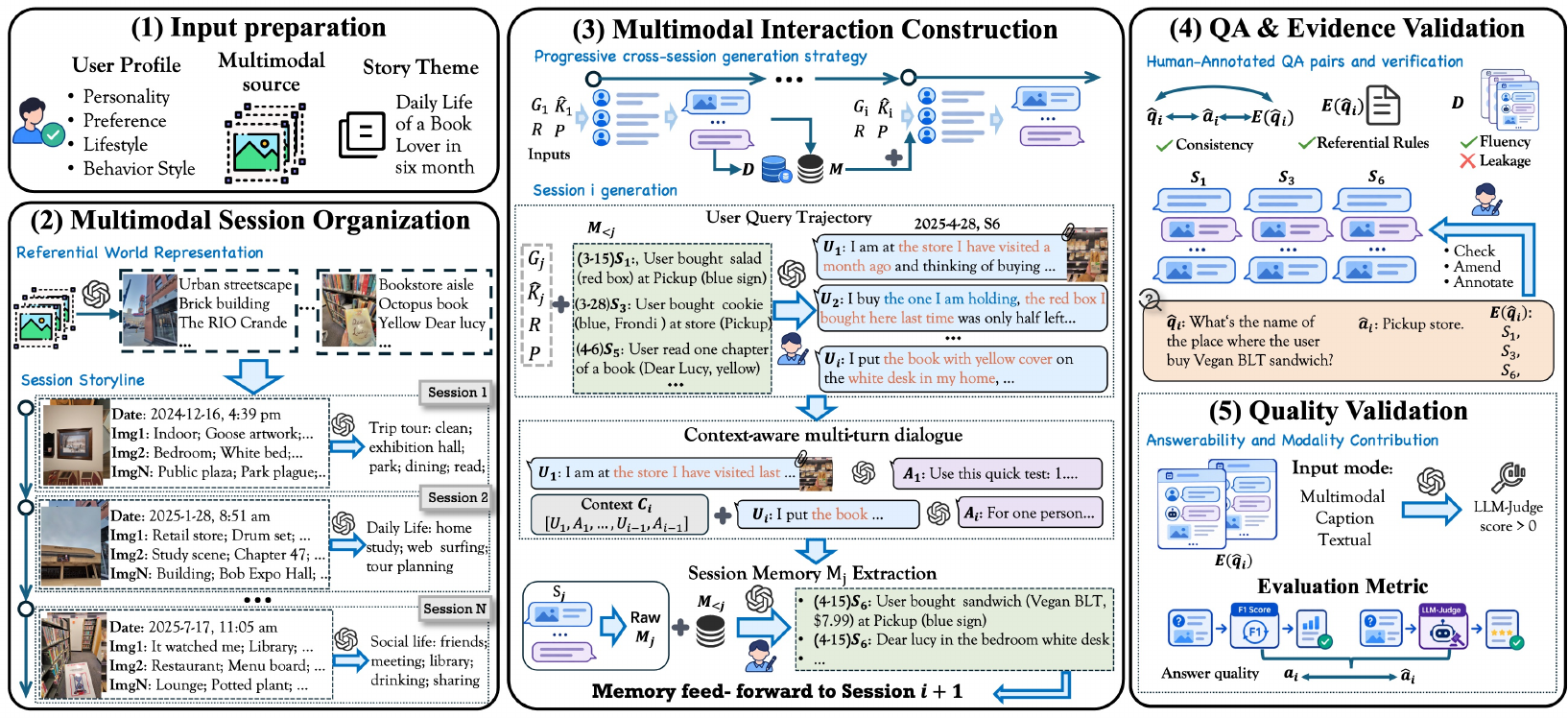}
    \caption{The overall construction pipeline of our proposed MemRR.}
    \label{fig_3}
\end{figure*}

\subsection{Cross-modal reference principles}

Conversation construction is guided by
two complementary dimensions: \textit{reference forms} and
\textit{referential semantic evolution}. The former focuses on the reference forms and analysis clues, while the latter focuses on the connections and changes in semantics across subsequent interactions.

\textbf{Reference Forms.} We consider two fundamental forms of reference: \textit{intra-turn reference}, which simulates referents within the current visual context through visual-grounded information or semantic-attribute, e.g., ``the object to the left of the white box''; \textit{cross-turn reference}, which simulates the non-repetitive re-referencing of existing and unmentioned referents using implicit attributes. Cross-turn reference occur within or across sessions, and relying on temporal, attribute, relation, ordinal- or sequence-based index, e.g., "the box I bought at Pickup for the first time."

\textbf{Referential Semantic Evolution.} Referential semantics dynamically evolve as interactions progress, such as attributes, relationships, states, and user knowledge, ensuring dynamic memory updates and maintain semantic consistency over time. We define three major patterns: \textit{relational establishment}, which establishes new connections between historical referents and current content through comparison, modification, or substitution, e.g., ``the box in my hand costs half as much as the small one''; \textit{temporal evolution}, which updates the attributes, constraints, or intentions associated with a referent over time while preserving its identity continuity, e.g., ``the red cereal box finished two days ago''; \textit{semantic continuation}, which extends ongoing discourse by referring to, supplementing, correcting, or negating former statements, thereby incrementally refining the conversational context, e.g., ``I was mistaken earlier, I meant this book instead.''

\subsection{Construction Pipeline}

Fig. \ref{fig_3} illustrates the overall pipeline. Instead of organizing around image-text relevance or fact retrieval, MemRR centers data construction on visual entities and their referential relationships. 
It progressively builds structured visual knowledge, integrates them into temporally and semantically coherent interactions, and constructs reasoning tasks involving cross-turn and cross-session evidences.

\textbf{Input preparation.}
The inputs are initialized with three inputs $(T,I,P)$: a \textit{Story Theme $T$}, which defines the general guidance of events within a time span; an \textit{Image Collection $I$}, material \cite{jiang-etal-2025-memory,jia2025omnispatial,zhou2025roborefer,galappaththige2025multi,yu2016modeling} that containing diverse entities and spatial or semantic cues; and a \textit{User Profile $P$}, encoding the personality, preferences, lifestyle, and behavioral patterns to guide personalized interaction generation.

\textbf{Multimodal Session Organization.}
Given $I$, we first construct a \textit{Referential World Representation} through visual analysis. It captures scene-level descriptions, salient entities and their attributes, spatial relationships, and OCR information. It also identifies referring expressions, potential events, and reasoning opportunities, yielding a structured visual representation $K$ as referential anchors. Based on $K$, images are organized into sessions according to shared semantic and referential contexts, yielding
a session-level set $\hat{K} = \{\hat{K_1},\hat{K_2},\ldots,\hat{K_S}\}$.
This process considers cross-image associations and event continuity. 
For each session, a storyline $G_i$ specifies its central topic, key entities, and events. The resulting storylines $G=\{G_1,G_2,\ldots,G_S\}$ are chronologically organized to establish coherent interactions with cross-session semantic dependencies.

\textbf{Multimodal Interaction Construction.}
To establish evolving semantic associations and long-range contextual dependencies, as shown in Fig. \ref{fig_3}(3), we adopt a progressive cross-session generation strategy, complemented by intra-session query planning and recalled memory incorporation to ensure coherence and long-term semantic continuity. 

For the $j$-th session, memory $M_{<j}$ is selectively incorporated into current dialogue materials to support cross-session references using connection establishment and semantic evolution.
Within $S_j$, we set a planner to plan the user's complete narrative thread, which determines interaction intentions and reference patterns across dialogue turns:

\begin{equation}
    Q_j =
    \mathcal{F}_{\mathrm{plan}}
    (G_j,\hat{K_j},M_{<j},\mathcal{R},P),
\end{equation}

where $Q_j=\{q_{j,1},q_{j,2},\ldots,q_{j,L_j}\}$ denotes the trajectory of user queries within $S_j$, and $L_s$ is the length. 
The predefined referential rules $\mathcal{R}$ govern the reference diversity and semantic evolution patterns, while $M_{<j}$ provides memory worth referring to construct temporally and semantically grounded references.
For example, a memory item that ``the user visited the `Pickup' store on March 20, 2025'', which can be reactivated through
temporal, spatial, or event-related references in subsequent sessions.

Given $G_j$ and $Q_j$, we generate the natural and coherent multi-turn user-assistant dialogue:

\begin{equation}
    S_j =
    \mathcal{F}_{\mathrm{dialogue}}
    (G_j,\hat{K_j},Q_j,M_{<j},P).
\end{equation}

$S_j$ preserves multiple reference expressions, contextual dependencies, and semantic transitions. Finally, relevant user-related session-level memory is generated by:

\begin{equation}
    M_j = \mathcal{F}_{\mathrm{merge}}(M_{<j}, \mathcal{F}_{\mathrm{mem}}(S_j,\hat{K_j})),
\end{equation}

which is accumulated for subsequent session generation.
The complete long-term conversation is obtained by chronological concatenation:
\begin{equation}
    D = S_1 \oplus S_2 \oplus \cdots \oplus S_N,
\end{equation}
where $\oplus$ denotes chronological concatenation.

\textbf{QA Construction and Evidence Validation.}
To ensure high-quality QA pairs that requiring cross-turn or cross-session evidence integration, grounding references in visual cues, and referential reasoning, rather than explicit fact retrieval, QA pairs are primarily constructed and annotated manually.
For each $\hat{q_i}$, we annotate its ground-truth answer $\hat{a_i}$ and supporting evidence sessions $E(q_i)$. Human verification and revisions consists of three rounds: (1) checking the reasoning process and consistency among $\hat{q_i}$, $E(q_i)$, and $\hat{a_i}$; (2) validating entity associations and reference resolutions conform to the predefined referential rules; (3) examining dialogue fluency and potential answer leakage.

\begin{figure}[t]
    \centering

    \begin{minipage}[c]{0.45\columnwidth}
        \centering
        \includegraphics[width=\linewidth]{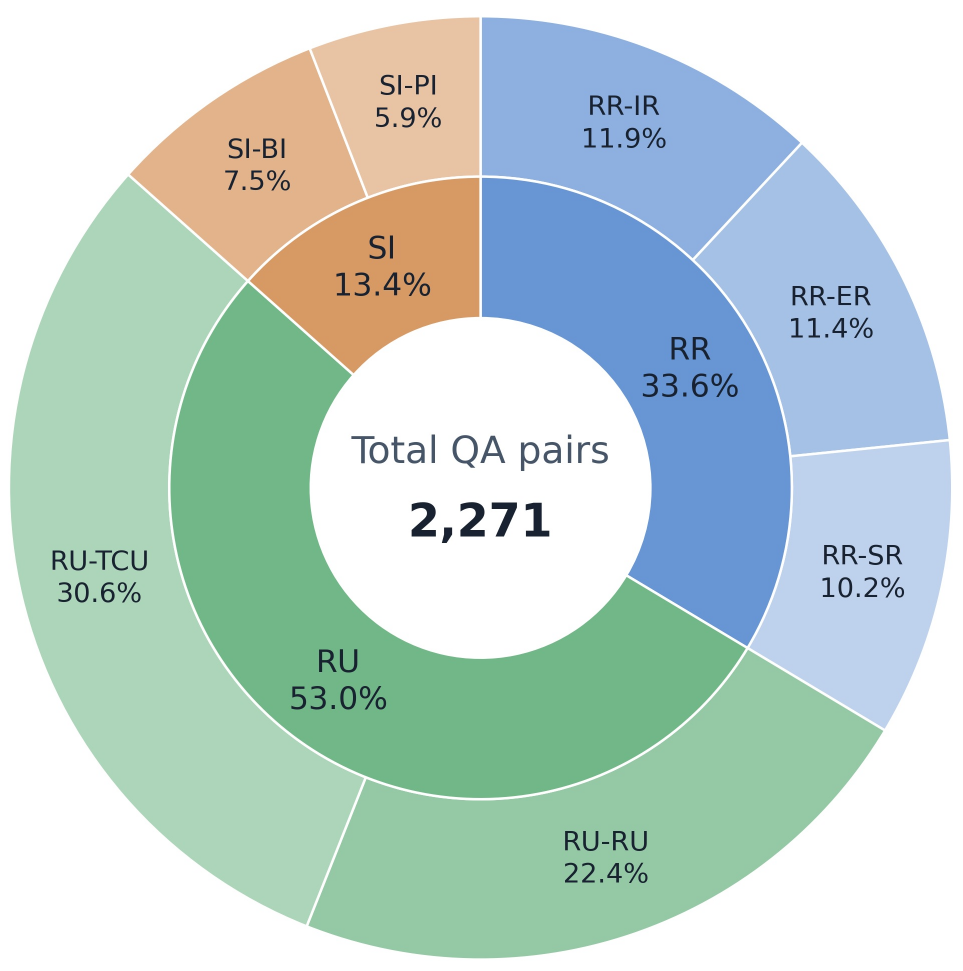}
        \label{fig_4}
        
    \end{minipage}
    \hfill
    \begin{minipage}[c]{0.4\columnwidth}
        \centering
        \resizebox{\linewidth}{!}{
        \begin{tabular}{lc}
            \toprule
            Statistic & Value \\
            \midrule
            Scenarios & 10 \\
            QA pairs & 2271 \\
            Total turns & 4638 \\
            Avg.sessions & 20.3 \\
            Avg.total tokens & $\sim$1.1M \\
            Avg.images/session & 5.5 \\
            Avg.turns/session & 22.8 \\
            Evidence sessions & 1$\sim$12 \\
            \bottomrule
        \end{tabular}
        }
    \end{minipage}
    \vspace{-1pt}
    \caption{Overview of the dataset statistics and distribution.}
    \label{fig:dataset_overview}
\end{figure}

\begin{table*}[t]
    \centering
    \caption{Cross-modality ablation on the visual necessity. Table shows the answerability ratio, which is the proportion of results where the judge score is greater than zero under given evidence sessions.}
    \label{tab:comparison}
    \small
    \begin{tabular}{c|cccccccc}
        \toprule
         & RR-IR & RR-ER & RR-SR & RU-RU& RU-TCU&SI-BI&SI-PI&Overall \\
        \midrule
        Full Multimodal Evidence Context  & 92.02\% & 88.79\% & 91.79\% & 88.81\% & 77.78\% & 95.77\% & 96.52\% & 86.93\%  \\
        Full Evidence Context with Image Caption  & 52.94\% & 56.95\% & 57.73\% &62.79\% & 43.49\% &74.83\% &88.70\% &56.69\% \\
        Full Text Evidence Context & 31.93\% & 41.26\% &35.05\% &53.20\% &33.33\% &54.55\% & 71.26\% & 38.76\%\\
        \bottomrule
    \end{tabular}
\end{table*}

\subsection{Evaluation Dimensions}
\label{3.4}

MemRR establishes multimodal memory evaluation framework centered on cross-modal referential reasoning from three interdependent yet distinguishable perspectives: \textit{reference resolution}, \textit{referential understanding}, and \textit{semantic inference}, progressively evaluating an agent's ability to resolve referents, understand evolving semantic structures, and infer latent user knowledge from accumulated referential expressions.

\textbf{Reference Resolution.}
This dimension focuses on resolving referents through fine-grained alignment among expressions, visual entities, and semantic context. It requires integrating cross-modal evidence across dialogue turns to identify referenced entities, events, and their contexts. It comprises three sub-dimensions:

\begin{itemize}
    \item \textbf{Instance Resolution (RR-IR).}
    Evaluates the ability to identify referenced visual entities and recover their identities, attributes, and states. Since many references originate from visual scenes, it requires establishing cross-modal correspondences between linguistic references and visual instances, and integrating non-lexically repeated evidence distributed across multiple turns.

    \item \textbf{Event Resolution (RR-ER).}
    Evaluates the ability to identify historical events and recover their key components, including activities, participants, and outcomes. It requires integrating temporally distributed evidence, tracing semantic progression of local interactions, and associating events with relevant entities and facts.

    \item \textbf{Situational Resolution (RR-SR).}
    Evaluates the ability to recover the situational context of referenced entities or events, including their temporal and spatial locations and environmental context, through the integration of distributed linguistic and visual evidence. 
\end{itemize}

\textbf{Referential Understanding.}
Beyond resolving references, this dimension evaluates the ability to model structured relationships and temporal dependencies among referents and user knowledge. It focuses on performing integrated reasoning based on explicit or implicit visual evidence and context across turns and sessions, to maintain relational structures and temporal consistency in long-term memory. It comprises two sub-dimensions:

\begin{itemize}
    \item \textbf{Relational Understanding (RU-RU).}
    Evaluates the ability to identify and reason about relationships among entities and events, including spatial, temporal, hierarchical, and comparative relations, as they are progressively established and updated throughout long-term interactions. 
    It requires integrating and reasoning over cross-modal evidence within or across sessions to structurally organize long-term memory and perform relational reasoning.

    \item \textbf{Temporal Consistency Understanding (RU-TCU).}
    Evaluates the ability to maintain consistency of referents while tracking the temporal evolution of their attributes, states, and user knowledge. As the same referent be revisited across sessions through various visual descriptions or semantic roles, it requires integrating cross-session evidences to perform long-range reference resolution, event consolidation, temporal ordering, numerical aggregation, and attribute accumulation, in order to reconstruct the long-term evolution trajectories of referents.
\end{itemize}

\textbf{Semantic Inference.}
Rather than focusing on resolving references and understanding relationships, it emphasizes inferring user preferences and behavioral intentions from accumulated referring expressions. It comprises two sub-dimensions:

\begin{itemize}
    \item \textbf{Behavioral Inference (SI-BI).}
    Evaluates the ability to infer users' underlying goals and prospective behavioral tendencies by reasoning over accumulated activities and patterns within or across sessions. It requires going beyond recovering related referents to anticipate likely subsequent actions or decisions.

    \item \textbf{Preference Inference (SI-PI).}
    Evaluates the ability to derive user preferences from dispersed and implicit behavioral evidence. Rather than relying on explicit preference statements, it requires identifying relevant entities or events and abstracting recurring behavioral patterns into higher-level preference representations.
\end{itemize}

\subsection{Statics and Quality Validation}

\textbf{Statistics.} As shown in Fig. \ref{fig:dataset_overview}, MemRR comprises 10 user scenarios, 4638 dialogue turns, and 2,271 QA pairs. Each scenario contains an average of 111.3 images and 463.8 dialogue turns, with about 1.1M tokens of multimodal context. Each QA pair requires evidence from 1$\sim$12 sessions and targets visual entities or events, posing challenges for referential reasoning. Details are shown in Appendix \ref{appendix_A}. We use GPT-5.6-sol for data generation and adopt human verification as the primary quality control mechanism.

\textbf{Data Quality and Answerability.} To assess QA reliability, we provide GPT-5.6-sol and Gemini-3.1-Pro with the full multimodal context from the corresponding evidence sessions, denoted by $E(q_i)=\{S_1,\ldots,S_k\}$. As reported in the 1st row of Table \ref{tab:comparison}, we measure the proportion of QA pairs receiving positive LLM-Judge scores. 86.93\% of QA pairs are answerable under the full evidence setting, while the remaining 13\% remain more challenging.

\textbf{Modality Contribution.} 
We further investigate the contribution of visual modality by replacing images in $E(q_i)$ with their captions. This substitution reduces the answerable proportion to 56.69\%, indicating that conventional captions preserve scene-level semantics and salient entities but fail to convey sufficient fine-grained visual details. 
Specifically, 30.24\% of the QA pairs require visual details that cannot be adequately recovered from captions. Retaining only textual context without visual inputs further reduces the answerable proportion to 38.76\%. This corresponds to a 48.17\% drop relative to the full multimodal evidence setting, demonstrating that textual context alone cannot recover substantial information grounded in visual observations, such as entity attributes, spatial configurations, and cross-modal associations. These results highlight the critical role of fine-grained visual information in long-term multimodal memory and cross-modal reference reasoning.

\begin{table*}[t]
\centering
\caption{Overall performance on MemRR. All memory agents are implemented with Gemini-2.5-Flash as base model. Best and second-best are marked in bold and underline: Frontier MLLMs are ranked separately, while agentic-memory methods are ranked together.}
\label{tab:main_results}

\setlength{\tabcolsep}{4pt}
\renewcommand{\arraystretch}{1.15}

\resizebox{\textwidth}{!}{
\begin{tabular}{c|cc|cc|cc|cc|cc|cc|cc|cc}
\toprule


\multirow{3}{*}{\textbf{Method Type}}
& \multicolumn{6}{c|}{Referential Resolution}
& \multicolumn{4}{c|}{Referential Understanding}
& \multicolumn{4}{c|}{Semantic Inference} 
& \\
& \multicolumn{2}{c|}{\textbf{RR-IR}}
& \multicolumn{2}{c|}{\textbf{RR-ER}}
& \multicolumn{2}{c|}{\textbf{RR-SR}}
& \multicolumn{2}{c|}{\textbf{RU-RU}}
& \multicolumn{2}{c|}{\textbf{RU-TCU}}
& \multicolumn{2}{c|}{\textbf{SI-BI}}
& \multicolumn{2}{c|}{\textbf{SI-PI}}
& \multicolumn{2}{c|}{\textbf{Overall}} \\

& F1 & LLM
& F1 & LLM
& F1 & LLM
& F1 & LLM
& F1 & LLM
& F1 & LLM
& F1 & LLM
& F1 & LLM \\

\midrule


\rowcolor{green!8}
\multicolumn{17}{c}{\textit{\textbf{Frontier MLLMs}}} \\

Claude-Sonnet-4.6 \cite{claude2026}
& 0.1940 & 0.2561 
& 0.1328 & 0.1983 	
& 0.2084 & 0.3115 
& 0.2085 & 0.2550 	
& 0.1259 & 0.1562 	
& 0.1290 & 0.2538 
& 0.1548 & 0.5014 
& 0.1635 & 0.2318   \\

Gemini-3.1-Pro-Preview \cite{gemini32026}
& \underline{0.4655} & \underline{0.7098}
& \textbf{0.4181} & \textbf{0.5944} 
& \textbf{0.4898} & \underline{0.7147}
& \textbf{0.4270} & \underline{0.6128 }
& \textbf{0.3074} & \textbf{0.4497 }	
& \underline{0.3898} & \underline{0.6418} 
& 0.2755 & 0.5955 	
& \textbf{0.3922} & \underline{0.5814}   \\

GPT-5.5 \cite{singh2025openai}
& 0.3958 & 0.5457 
& 0.3536 & 0.5119 
& 0.3879 & 0.5886 	
& 0.3924 & 0.5038 
& 0.2253 & 0.2987  
& 0.3188 & 0.5726 
& 0.2644 & \underline{0.7740}
& 0.3242 & 0.4679 \\

Kimi-K3 \cite{team2026kimi}
& \textbf{0.4888} & \textbf{0.7409} 
& \underline{0.4171} & \underline{0.5928}
& \underline{0.4859} & \textbf{0.7206}  
& \underline{0.4133} & \textbf{0.6300} 
& \underline{0.2926} & \underline{0.3952}
& \textbf{0.4050} & \textbf{0.7166} 
& \textbf{0.3238} & \textbf{0.8428} 
& \underline{0.3912} & \textbf{0.5891}  \\

Qwen3.7-Plus \cite{qwen32026}
& 0.3674 & 0.4886
& 0.2912 & 0.4488 
& 0.3465 & 0.5312 	
& 0.3011 & 0.4196 
& 0.1487 & 0.2436 
& 0.2448 & 0.4758
& 0.2363 & 0.6035
& 0.2563 & 0.3922  \\

Deepseek-V4-Pro \cite{deepseekai2026deepseek}
& 0.2142 & 0.3451 
& 0.1983 & 0.3220 
& 0.1936 & 0.3039 
& 0.2487 & 0.3365 
& 0.1338 & 0.2228 
& 0.2230 & 0.4148
& \underline{0.2800} & 0.6825 
& 0.1954 & 0.3160  \\

GLM-5.2 \cite{glm2026}
& 0.1760 & 0.3431 	
& 0.1540 & 0.3276 	
& 0.2316 & 0.3688 	
& 0.1758 & 0.3264 
& 0.0950 & 0.1951 
& 0.1952 & 0.4766 
& 0.2483 & 0.7402 
& 0.1546 & 0.3140  \\

\midrule
\rowcolor{gray!10}
Gemini-2.5-Flash \cite{gemini22025}
& 0.3288 & 0.4658 
& 0.3006 & 0.3854 
& 0.3463 & 0.5152 	
& 0.3339 & 0.4172 	
& 0.2031 & 0.2467 	
& 0.2398 & 0.4590 	
& 0.2587 & 0.6172 	
& 0.2807 & 0.3868  \\


\midrule
\rowcolor{blue!10}
\multicolumn{17}{c}{\textit{\textbf{Text-based Agentic-Memory methods}}} \\

NaiveRAG
& 0.1517 & 0.2596 	
& 0.1089 & 0.1806 	
& 0.1247 & 0.2493 	
& 0.1434 & 0.1841 	
& 0.0577 & 0.1211 	
& 0.1508 & 0.3513 
& 0.2328 & \underline{0.6562}
& 0.1147 & 0.2094  \\

A-Mem \cite{xu2025mem}
& 0.1424 & 0.2281 	
& 0.1028 & 0.1647 	
& 0.1482 & 0.2596 
& 0.1254 & 0.1632 
& 0.0614 & 0.1270 	
& 0.1664 & 0.3744 
& \textbf{0.2440} & 0.6230 
& 0.1138 & 0.2019  \\

SimpleMem \cite{liu2026simplemem}
& \underline{0.1759} & 0.1727 
& \underline{0.1250} & 0.2018 
& 0.1532 & 0.2082 
& \textbf{0.2328} & \underline{0.2093}
& 0.0853 & \underline{0.1366}
& \textbf{0.1801} & \underline{0.3766}
& \underline{0.2402} & 0.5637 
& \textbf{0.1563} & 0.2116  \\

MRAgent \cite{ji2026memory}
& 0.1618 & 0.2299 
& \textbf{0.1452} & 0.1652 
& \underline{0.1639} & 0.2518 
& \underline{0.1756} & 0.1835 
& \textbf{0.1226} & \textbf{0.1531} 
& 0.1033 & 0.2782 
& 0.1970 & 0.4917 
& \underline{0.1480} & 0.2072  \\


\midrule
\rowcolor{red!5}
\multicolumn{17}{c}{\textit{\textbf{Multimodal Agentic-Memory methods}}} \\

MuRAG \cite{chen2022murag}
& 0.1622 & 0.2755 
& 0.0785 & 0.1532 
& 0.1262 & 0.2391  
& 0.1319 & 0.1738 
& 0.0539 & 0.1073 
& \underline{0.1721} & \textbf{0.4081} 
& 0.2255 & 0.6338 
& 0.1117 & 0.2063  \\

UniversalRAG \cite{yeo2026universalrag}
& \textbf{0.1982} & \underline{0.2959}
& 0.1215 & 0.1901 	
& \textbf{0.1824} & \underline{0.3037} 
& 0.1546 & 0.1943 	
& \underline{0.0733} & 0.1263 
& 0.1720 & 0.3696 
& 0.2371 & 0.6506 
& 0.1383 & \underline{0.2272}  \\

NGM \cite{fisher2025neural}
& 0.0716 & 0.1079 	
& 0.0457 & 0.0717 	
& 0.0862 & 0.1498  
& 0.0545 & 0.0875 
& 0.0294 & 0.0511 
& 0.1041 & 0.2524 
& 0.1605 & 0.4179 
& 0.0584 & 0.1053  \\

Omni-SimpleMem \cite{liu2026omni}
& 0.1065 & 0.1679 	
& 0.0682 & 0.1373 	
& 0.1289 & 0.1998 	
& 0.1256 & 0.1752 	
& 0.0525 & 0.1054 	 
& 0.1593 & 0.3164 	
& 0.2201 & 0.4468 	
& 0.0993 & 0.1671  \\

V-Mem \cite{kang2026v}
& 0.1272 & 0.2359 	
& 0.0836 & \underline{0.2202}	
& 0.1061 & 0.2213 	
& 0.1285 & 0.1860 	
& 0.0529 & 0.1233 	
& 0.0928 & 0.2342 	 
& 0.1988 & 0.5295 	
& 0.0965 & 0.1938  \\

RAG-Anything \cite{guo2025rag}
& 0.0988 & \textbf{0.4343} 	
& 0.0728 & \textbf{0.2544} 	
& 0.0862 & \textbf{0.4076} 	
& 0.1416 & \textbf{0.2393} 	
& 0.0484 & 0.1260 	
& 0.0713 & 0.3399 	
& 0.1158 & \textbf{0.7050} 	
& 0.0858 & \textbf{0.2687}  \\

\bottomrule
\end{tabular}
}
\end{table*}

\section{Benchmarking Analysis}
\subsection{Experimental setup}

\textbf{Models and Baselines.} We consider two baselines:
Frontier MLLMs and memory-augmented agents. We benchmark seven frontier MLLMs under full-context setting, including GPT-5.5 \cite{singh2025openai}, Claude-Sonnet-4.6 \cite{claude2026}, Gemini-3.1-Pro-Preview \cite{gemini32026}, Qwen3.7-Plus \cite{qwen32026}, Kimi-K3 \cite{team2026kimi}, GLM-5.2 \cite{glm2026}, DeepSeek-V4-Pro \cite{deepseekai2026deepseek}. All support context windows of up to 1M tokens. Since the overall multimodal history exceeds this limit, we resize input images to fit the entire multimodal conversation within the available context window.
We benchmark ten memory agents that maintain external
memory stores through dedicated memory mechanisms, including NaiveRAG, A-Mem \cite{xu2025mem}, MRAgent \cite{ji2026memory}, SimpleMem \cite{liu2026simplemem}, MuRAG \cite{chen2022murag}, UniversalRAG \cite{yeo2026universalrag}, NGM \cite{fisher2025neural}, V-Mem \cite{kang2026v}, Omni-SimpleMem \cite{liu2026omni}, and RAG-Anything \cite{guo2025rag}. We adapt each method to MemRR while preserving its original strategy and provide image captions for text-based agents. All agents are implemented with four base models: Gemini-2.5-Flash \cite{gemini22025}, GPT-4o-mini \cite{openai2024gpt}, and Qwen3-VL-Plus \cite{bai2025qwen3}. Further model specifications and implementation details are provided in the Appendix \ref{appendix_B}.

\textbf{Evaluation Metrics.} We adopt two complementary metrics: token-level F1 and LLM-Judge score. F1 measures lexical overlap between generated and ground-truth answers, while LLM-Judge score assesses semantic agreement beyond exact lexical matching. 
Specifically, we employ GPT-5.2 \cite{openai2024gpt} and Qwen3.5-122B-A10B \cite{qwen352026} as independent judges and report their average score to mitigate individual judge bias. We report the average performance across the evaluation set for both metrics. Further evaluation protocol and criteria are detailed in the Appendix \ref{appendix_metrics}.

\begin{figure*}[h]
	
	\begin{minipage}{0.49\linewidth}
		\centerline{\includegraphics[width=\textwidth]{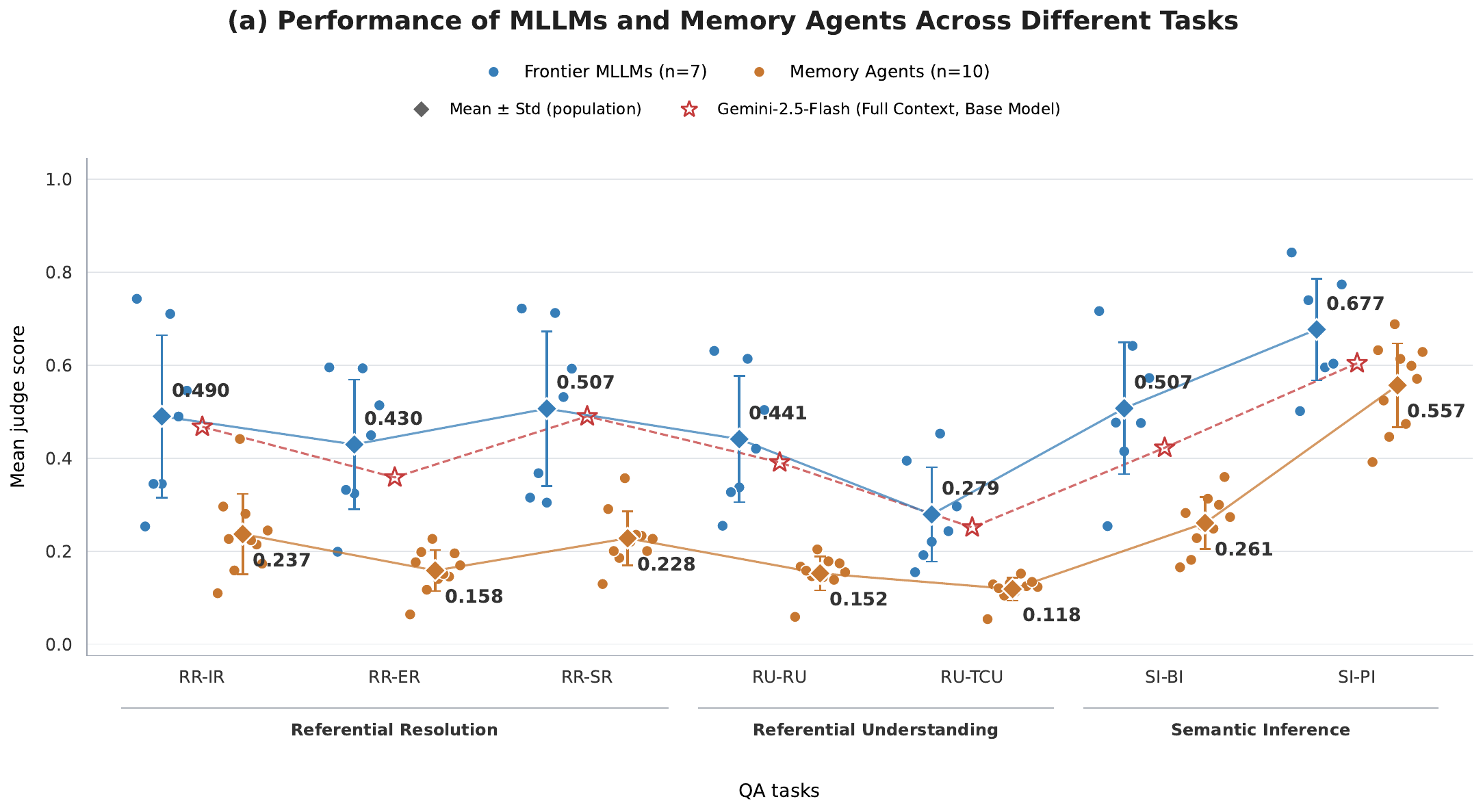}}
	\end{minipage}
	\begin{minipage}{0.49\linewidth}
		\centerline{\includegraphics[width=\textwidth]{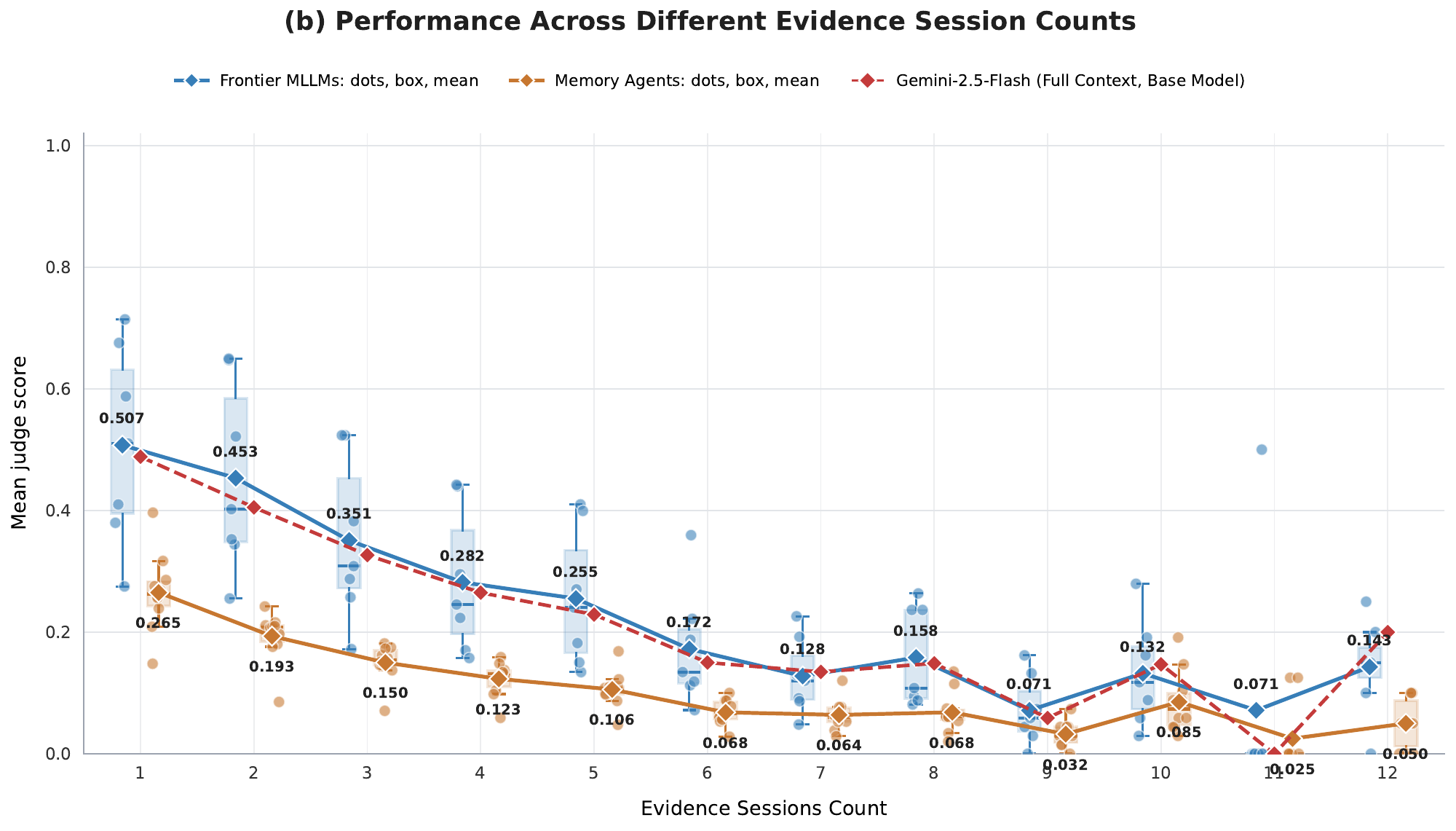}}
	 
	\end{minipage}
	 
	 
 
	\caption{Performance of MLLMs and memory agents across QA tasks (a) and evidence session counts (b).}
	\label{fig4}
\end{figure*}

\subsection{Main results}
\textbf{Overall Performance.} In Table \ref{tab:main_results}, frontier MLLMs outperform memory agents. Kimi-K3 \cite{team2026kimi} and Gemini-3.1-Pro-Preview \cite{gemini32026} achieve the highest overall F1 scores of 0.391 and 0.392, respectively, with LLM-Judge scores of 0.589 and 0.581. 
Text-only MLLMs achieve LLM-Judge score no higher than 0.32, highlighting the importance of visual details for cross-modal reference reasoning. But the advantage of full-context MLLMs comes with substantially greater context and computational requirements.
Existing memory agents exhibit a substantial performance gap, with average F1 and LLM-Judge score of 0.156 and 0.269. SimpleMem \cite{liu2026simplemem} achieve best overall F1 and LLM-Judge score of 0.156 and 0.212 among text-based agents. Among multimodal agents, UniversalRAG \cite{yeo2026universalrag} achieves the best F1 score of 0.138, while Rag-Anything \cite{guo2025rag} obtains the best LLM-Judge score of 0.269.
These results expose limitations in existing memory agents' ability to preserve and exploit relevant information for long-range referential reasoning. Alternative base model evaluations are in Appendix \ref{appendix_C}.

\textbf{Task-specific Difficulty.}
Fig. \ref{fig4}(a) presents performance across QA tasks, revealing consistent task-specific difficulty patterns. 
Frontier MLLMs perform relatively well on semantic inference, achieving scores of 0.677 on SI-PI and 0.507 on SI-BI, but struggle with RU-TCU (0.279). Memory agents exhibit a similar pattern, with strongest performance on semantic inference (0.261 and 0.557) and substantially lower performance on RU-TCU (0.118). Detailed cases are in Appendix \ref{appendix_D}.
This suggests that models can infer user preferences and behavioral tendencies from referring expressions, however, there are still challenges in referential resolution and reasoning tasks required maintaining structural associations and temporal consistency.
Moreover, memory agents underperform frontier MLLMs across all QA tasks, with an average gap of 0.231, especially in RR-SR of 0.279 and RU-RU of 0.289. These findings highlight the limitations of existing methods in resolving cross-modal references and reasoning over relations among referents.

\textbf{Cross-Session Integration Capabilities.}
Fig. \ref{fig4}(b) examines performance as the number of evidence sessions increases. 
Both frontier MLLMs and memory agents exhibit pronounced performance degradation as evidence becomes distributed across more sessions. 
The average LLM-Judge score of MLLMs decreases from 0.507 to 0.071 as the evidence sessions count increase, while that of memory agents drops from 0.265 to 0.025. Memory agents also consistently lag behind MLLMs and the base model across most evidence-session settings. 
This trend demonstrates the difficulty of integrating distributed evidence across sessions, revealing a fundamental bottleneck in long-term reasoning.

\subsection{Analysis}

\textbf{Persistent Challenges in Long-Range Cross-Modal Referential Reasoning}.
Frontier MLLMs achieve a best LLM-Judge score of only 0.589, while the best memory agent remains below 0.3, highlighting the challenges of cross-modal referential reasoning beyond conventional information retrieval.
Despite access to external memory, existing agents struggle to integrate and reason over relevant evidence distributed across sessions. 
Both performance degrades as evidence session count increases, suggests that expanding the evidence scope not only enlarges the retrieval space but also complicates cross-session association and multi-hop evidence reasoning. \textit{These findings expose fundamental limitations in existing methods for organizing, associating, and jointly reasoning over long-range distributed cross-modal referential tasks.}

\textbf{Performance Gap Between Memory agents and Full-Context Reasoning}.
Memory agents remain substantially behind the full-context base model across overall performance and QA tasks, with limitations extending beyond information storage and retrieval to cross-turn evidence integration and complex referential reasoning. 
The best text-based agent lags behind the baseline by 0.124 in F1 and 0.165 in LLM-Judge score, while the corresponding gaps for the best multimodal agents are 0.142 and 0.118, respectively. 
\textit{The persistent gap highlights the limitations of current methods in resolving long-range references, struggling to preserve, integrate and reconstruct the evidence chains required for referential reasoning.}

\textbf{Complex Referential Understanding and Semantic Reasoning Remain Major Challenges}.
Referential resolution and understanding posing considerable challenges to both MLLMs and agents. 
Within reference resolution, RR-ER is notably more difficult than instance and situational resolution, with average scores of 0.43 for MLLMs and 0.158 for agents. This indicates that reconstructing events from distributed cross-turn references is relatively more challenging. Referential understanding poses an even greater challenge, particularly RU-TCU, which yields the lowest overall scores (0.279 and 0.118).
\textit{These results indicate that the abilities required for effective referential reasoning, such as modeling structural associations, maintaining temporal consistency, and tracking evolving semantics, remain challenges in current memory systems.}

\textbf{Multi-modality Retention Does Not Yield Reasoning Advantages}.
Multimodal methods do not consistently outperform their text-based counterparts.
Existing methods often rely on storing raw image content and optimizing the retrieval of relevant items, without explicitly modeling persistent entity-level associations across modalities, indicating insufficient for effective cross-modal referential reasoning.
In contrast, MemRR requires systems to maintain persistent associations and temporal identity consistency among referents, rather than storing raw images or dialogue snippets. 
Therefore, it may fail to determine which object the current reference denotes or whether references at different time point to the same entity.
\textit{These findings reveal a fundamental gap between multi-modality retention and its effective utilization for long-range referential reasoning.}

\textbf{Structured Memory Goes Beyond Image-text Similarity}. Current methods mostly rely on image-text similarity retrieval and memory-chunk aggregation, typically treating local dialogue segments or vectorized memory units as independent targets. 
In MemRR, the same entity is referenced non-repeatedly through indirect attributes or semantic roles, which is also updated temporally. It requires tracing cross-session referential chains and jointly reasoning over historical attributes, event relations, and evolving semantic roles. 
Thus, similarity retrieval alone without critical referential links does not guarantee sufficient evidence for accurate long-range cross-modal referential reasoning. 
\textit{Effective multimodal memory must go beyond retrieving relevant facts to organize distributed information into referents-centric structures that support persistent associations, temporal tracking, and multi-hop reasoning. }

\section{Conclusion}

In this work, we introduce MemRR, a benchmark for evaluating cross-modal referential reasoning in long-term multimodal memory. By emphasizing persistent referents associations and evolving references, MemRR extends existing memory evaluation beyond conventional fact retrieval. Extensive experiments reveal substantial limitations in both frontier MLLMs and memory agents, highlighting the challenges of long-range referential reasoning. MemRR underscore the need for structured multimodal memory, involving dynamically organizing associations among entities and events, and maintaining the semantic evolution of the same entity or event, supporting complex chain of evidence beyond local semantic similarity. MemRR openes new directions for developing future long-term multimodal agents.

\section*{LLM Usage Declaration}
We primarily use AI tools as supplementary tools, combined with manual inspection and verification.
ChatGPT was used to assist with writing and editing, including language refinement and restructuring, as well as related work and citation refinement. It also provided assistance with code and implementation. For dataset creation, ChatGPT was used to generate synthetic data, which was subsequently reviewed, edited, and approved by the authors. AI tools were used for code modifications with human verification in experiment design, execution, or result analysis. No AI tools were used in figure and diagram preparation; or the development of research ideas, hypotheses, or conclusions. The authors take full responsibility for all content in the submitted manuscript.

{
    \small
    \bibliographystyle{ieeenat_fullname}
    \bibliography{main}

}


\clearpage

\appendix

\begin{strip}
\begin{center}
    {\fontsize{18}{18}\selectfont \textbf{MemRR: Benchmarking Cross-Modal Referential Reasoning in Long-Term Multimodal Memory} \\[0.5em] {\fontsize{14}{12}\selectfont Supplementary Material}}\\[0.5em]
\end{center}
\end{strip}


\vskip 4mm
\startcontents[sections]
\vbox{\sc\Large Table of Contents}
\vspace{5mm}
\hrule height .8pt
\vspace{-2mm}
\printcontents[sections]{}{1}{\setcounter{tocdepth}{2}}
\vspace{4mm}
\hrule height .8pt
\vskip 10mm

\begin{figure*}
    \centering
    \includegraphics[width=\linewidth]{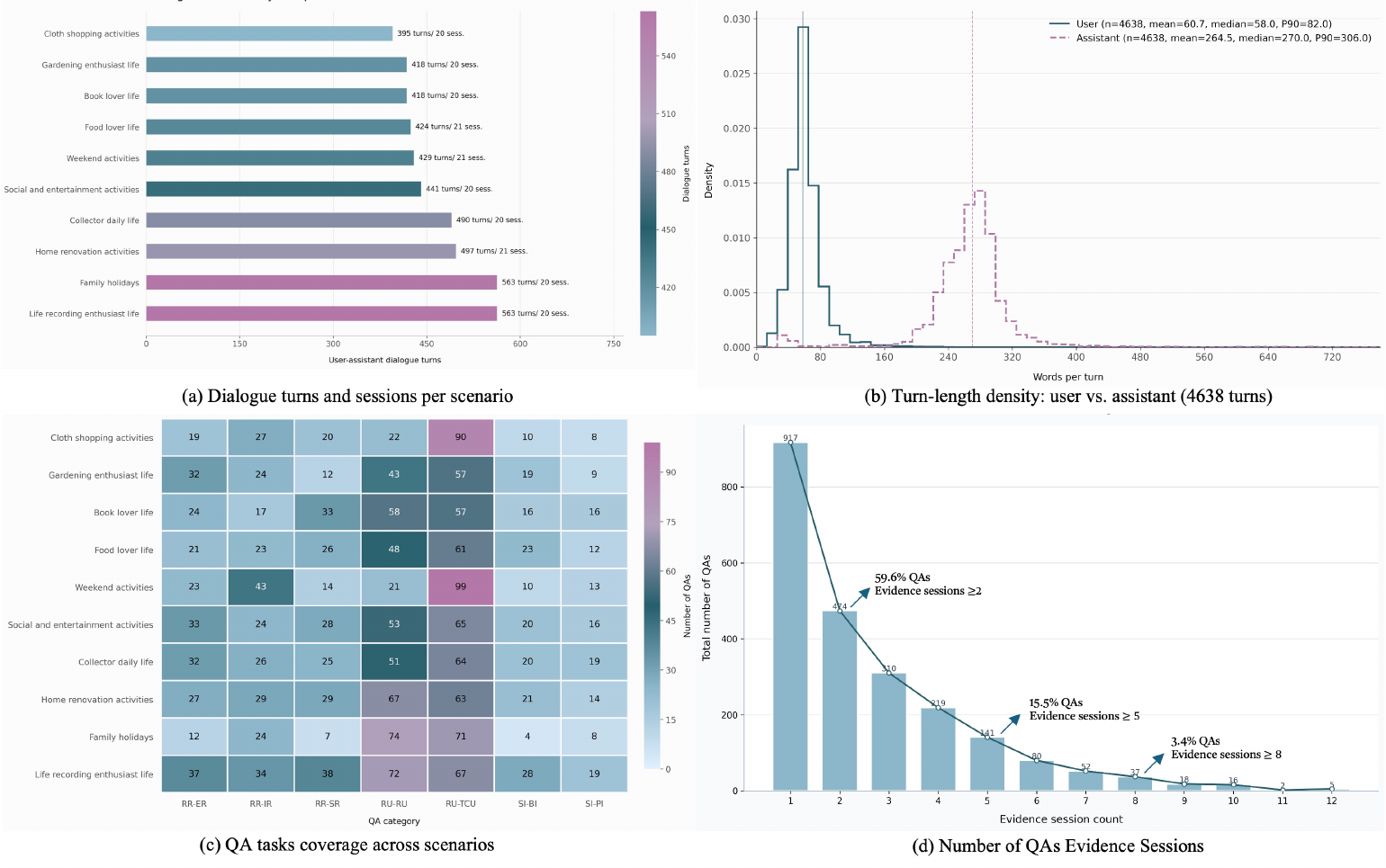}
    \caption{Data properties of MemRR. (a) Conversation size. Dialogue turns and sessions per scenario, with bar color encoding the number of turns. (b) Turn-length density. Gaussian-kernel density of word counts for user versus assistant messages. (c) QA-tasks coverage across different scenarios. Number of questions of each type (columns) per scenario (rows). (d) Statistics on the number of evidence sessions required for QA tasks. The plot shows the number and percentage of QAs requiring $k$ evidence sessions.}
    \label{fig:data_statistic}
\end{figure*}

\section{Datasets Details and Statistics}
\label{appendix_A}

\subsection{Reference Form Principles}

\textbf{Single-turn reference} simulates how users refer to one or multiple entities in the current visual scene based on immediate visual observations. According to the grounding cues, it is categorized into two primary types: \textit{Visual-Grounded Reference} and \textit{Semantic-Attribute Reference}.

Visual-Grounded Reference. Users identify entities through their spatial locations, visual attributes, quantity and ordering, or textual cues in the image. It comprises four subtypes:
\begin{itemize}
    \item Spatial Reference: Referring through absolute positions or relative spatial relationships, such as left, right, above, below, center, near, far, beside, behind, etc.
    \item Visual Attribute Reference: Identifying through salient visual properties, including color, shape, material, or other appearance descriptions.
    \item Ordinal Reference: Referring through numerical quantities, relative ordering, or comparative magnitude, such as ``the second box from left" or ``the largest one."
    \item  Text-Anchored Reference: Using textual information within images as referential anchors, such as ``the sign labeled `EXIT'" or ``the item tagged with `X'."
\end{itemize}

Semantic-Attribute Reference. Users identify entities through their semantic properties, including functions, states, purposes, or relationships with other entities. It comprises three subtypes:
\begin{itemize}
    \item Functional Reference: Identifying by intended functions or uses, such as the ``item used for storing documents" or ``the device responsible for displaying information."
    \item Relational Reference: Referring through semantic or spatial relationships, such as ``the computer connected to the camera" or ``the object inside the box."
    \item State-Based Reference: Identifying according to current states or conditions, such as ``the open window" or ``the broken flowerpot."
    
\end{itemize}

\textbf{Cross-turn reference} describes how users re-refer to entities, events, or scenes introduced in historical interactions, constituting a core component of long-term multimodal memory. Historical referents include both explicitly mentioned entities in previous dialogues and implicitly introduced entities that appear in images without explicit verbalization. Such references occur within or across sessions.

Rather than verbatim repetition, cross-turn reference relies on non-identical descriptions, using temporal, attribute-based, relational, and contextual cues to refer back to historical referents. It comprises four subtypes:
\begin{itemize}
    \item Temporal Reference: Referring to historical objects or events through absolute, relative, or stage-based temporal expressions, such as ``the book purchased on March 20", ``the route mentioned earlier", or ``the original version of the photo."
    \item  Attribute-Based Reference: Identifying historical entities through observed attributes or states, such as ``the broken computer" or ``the book with the black cover added."
    \item Sequential Reference: Referring to historical entities or events according to their order of occurrence, such as ``the building in the second image" or ``the first outfit."
    \item Relational Reference: Identifying historical entities through previously established spatial or contextual associations, such as ``the tool beside the sink" or ``the flowerpot next to the window."
\end{itemize}

\begin{table*}[t]
    \centering
\caption{Overall statistics for each scenario. The table lists the number of QA pairs, total sessions, total turns, number of images, total multimodal tokens, text tokens, and image tokens for each scenario. The token counts were calculated using the Qwen3 \cite{bai2025qwen3} tokenizer.}
\label{tab_4}
\resizebox{\textwidth}{!}{
\begin{tabular}{c|ccccccc}
\toprule

Scenario & QAs & Total sessions & Image counts & Total turns & Total tokens & Text tokens & Image tokens  \\
\midrule

Gardening enthusiast life & 196 &	20	&109	&418	&1147944 &	204868	& 953076  \\

Cloth shopping activities  & 196	&20&	112	&395	&1273614	& 146250 &	1127364   \\

Collector daily life & 237&	20 &	112	& 490	& 1276687	&236031	&1040636 \\

Weekend activities & 223	&21	&118	&429&	1004950&	172277	&832673   \\

Family holidays &  200 &	20&	107	&563	&1243103&	257494	& 985609 \\

Food lover life & 214	& 21 & 111 &	424	&1064420	&202879	&861541   \\

Book lover life & 221 &	20 &	108 &	418 &	1024976 &	205947 &	819029   \\

Home renovation activities& 250	&21	&114&	497	&816194	&239475	&576719  \\

Social and entertainment activities	&239	&20	&110	&441&	945955	&211689	&734266 \\

Life recording enthusiast life 	&295	&20	&112	&563	&1283484	&274504	&1008980 \\

\bottomrule
\end{tabular}
}
\end{table*}

\begin{figure*}[h]
	
	\begin{minipage}{0.49\linewidth}
		\centerline{\includegraphics[width=\textwidth]{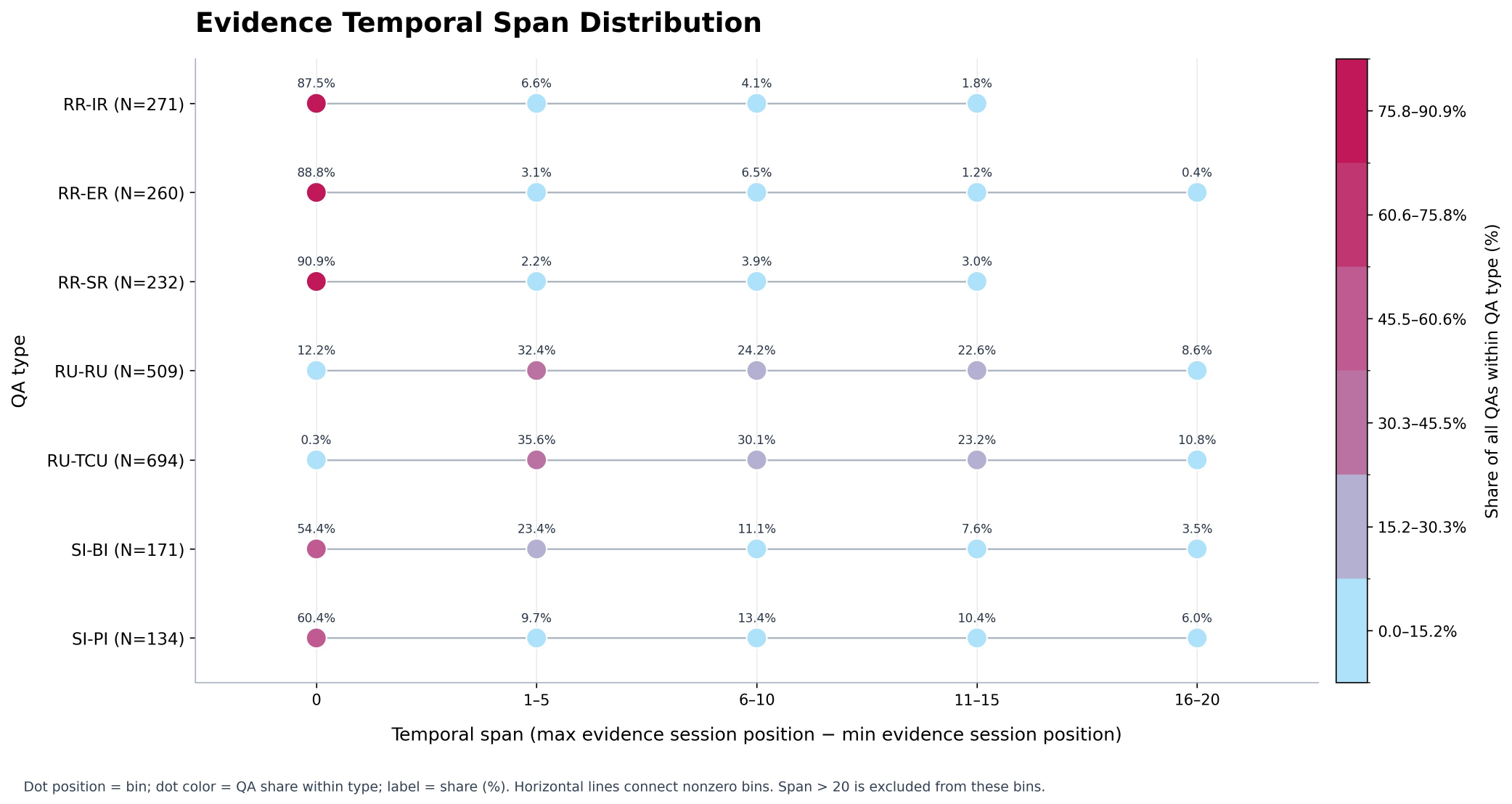}}
	\end{minipage}
	\begin{minipage}{0.49\linewidth}
		\centerline{\includegraphics[width=\textwidth]{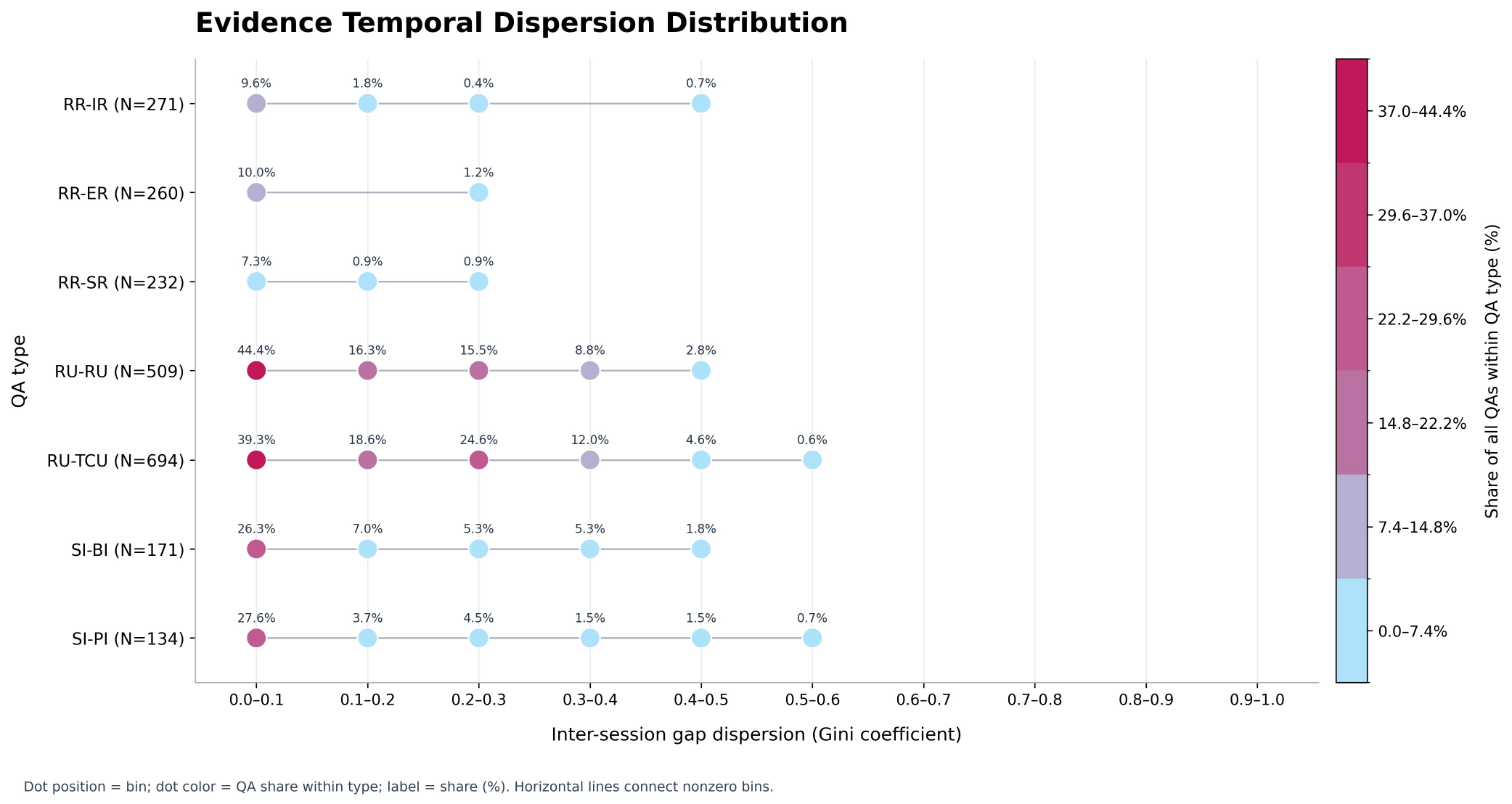}}
	 
	\end{minipage}
	\caption{Distribution of evidence sessions across different QA types. (a) The temporal span of evidence sessions, i.e., the distance between the most recent and the oldest interactions in the history, capturing the extent of temporal coverage. (b) The dispersion of evidence sessions, i.e., the degree of dispersion in the history of evidence interactions, capturing the regularity of evidence distribution.}
	\label{fig:evidence_distribution}
\end{figure*}

\subsection{Referential Semantic Evolution Principles}

In long-term interactions, referents are not static; their attributes, relationships, and associated user knowledge evolve as conversations progress. Accordingly, long-term multimodal memory must preserve referential consistency while accommodating semantic evolution over time. This principle captures how historical information is incorporated into current context through relational associations, temporal changes and semantic continuity, requiring models to maintain coherent referential representations while dynamically updating their understanding.

\textbf{Relational Association}. Historical entities or events establish new connections with current discussions, integrating previously acquired information into the evolving conversational context. It comprises four types:
\begin{itemize}
    \item Comparison: Establishing comparative relationships between historical and current entities or events based on their attributes or states, such as ``the item mentioned earlier is warmer than the one hanging in the upper-right corner" or ``the large box costs three times as much as the small one."
    \item Modification: Altering the attributes, states, or interpretations of previously entities or events, such as ``the contents of the box have been replaced with toys" or ``the potted plant beside the living room window has been moved to the balcony."
    \item Replacement: Substituting a previously referenced entity with a newly introduced one, such as ``replace the plant originally placed in the corner with this newly purchased one" or ``the item hanging by the door has been replaced with the one gifted by a friend."
    \item Update: Revising previous decisions or entity-related information based on subsequent interactions, such as ``the book earlier was too expensive, so I bought this one instead" or ``I returned the one with purple packaging."
\end{itemize}

\textbf{Temporal Evolution.} Semantic representations associated with the same entity, event, or user intention may change over time. This dimension captures the temporal evolution of referential attributes, states, and preferences, comprising three types:

\begin{itemize}
    \item Attribute Evolution: Changes in the attributes of the same entity over time, such as ``half of the cereal purchased last week is still left" or ``the price of the outfit I was considering has been reduced by half."
     \item State Evolution: Transitions of an entity or event between different states over time, such as ``I have completed my reading plan" or ``the route to the supermarket has become more congested."
     \item Preference Evolution: Changes in users' intensions, preferences, or interests over time, such as ``I now prefer romance films to comedies" or ``I currently need to prepare for the mathematics exam more than study English."
\end{itemize}

\textbf{Semantic Continuity} captures how users extend, refine, or reinterpret historical interactions while maintaining ongoing discussions and conversational coherence. It encompasses four types of discourse-level operations:

\begin{itemize}
    \item Negation: Rejecting or revising previous information or interpretations, such as ``I was mistaken earlier; I meant this book" or ``Ignore my previous question; what I actually wanted to ask was..."
    \item Factual Supplementation: Extending historical information by introducing additional facts, such as ``Regarding the issue we discussed earlier, I also need..." or ``Besides those items, I also purchased..."
    \item Correction: Rectifying the established information through further clarification, such as ``Your analysis is good, but ... is incorrect" or ``Everything except the price analysis is correct, but..."
    \item Conclusion Referencing: Invoking previously established conclusions or recommendations to inform subsequent decisions and actions, such as ``I decided to follow your recommendation and purchase the product" or ``Following the plan you developed, I will..."
\end{itemize}

\begin{table*}[t]
    \centering
    \caption{Statistics on evidence sessions by QA type. The table lists the proportion of multi-session QAs for each QA type, as well as the duration and dispersion of evidence sessions.}
    \resizebox{\textwidth}{!}{
    \begin{tabular}{c|ccccc}
        \toprule
         QA Type &  Valid count & Multi-session QA & Multi-session QA ratio & Mean evidence span & Mean evidence dispersion \\
         \midrule
         RR-IR & 271 & 34 & 12.5\% & 0.786 & 0.064 \\
         RR-ER & 260 & 29 & 11.2\% & 0.762 & 0.035 \\
         RR-SR & 232 & 21 & 9.1\%  & 0.728 & 0.035 \\
         RU-RU & 509 & 447 & 87.8\% & 7.204 & 0.125 \\
         RU-TCU & 694 & 692 & 99.7\% & 8.272 & 0.156 \\
         SI-BI & 171 & 78 & 45.6\% & 3.053 & 0.113 \\
         SI-PI & 134 & 53 & 39.6\% & 3.739 & 0.092 \\
         \bottomrule
    \end{tabular}
    }
    \label{tab_5}
\end{table*}

\subsection{Data Statistics}

Fig. \ref{fig:data_statistic} provides a comprehensive overview of MemRR from four complementary perspectives, covering corpus scale, interaction patterns, task composition, and evidence distribution. Overall, MemRR comprises 10 user scenarios and 203 multi-session dialogues, encompassing 4,638 User-Assistant interaction turns, 1,113 images, and 2,271 QA pairs. Table \ref{tab_theme_event} lists the themes and the session-level event trajectories for each scenario. All QA pairs are specifically designed around cross-modal referential reasoning, establishing MemRR as a benchmark for evaluating long-term multimodal memory beyond conventional fact retrieval.

Table \ref{tab_4} further reports the overall statistics for each scenario. 
The corpus exhibits a relatively balanced distribution, with 107--118 images and 196--295 QA pairs per scenario.
We use the Qwen3 \cite{bai2025qwen3} tokenizer to estimate the total length of the multimodal conversations, yielding approximately 1.1M tokens per scenario, including about 210K text tokens and 891K image tokens. 
This substantial multimodal context provides a challenging testbed for evaluating long-term memory under extended visual-linguistic interactions.

\textbf{Longitudinal Corpus Scale.}
Fig. \ref{fig:data_statistic}(a) illustrates the distribution of dialogues each scenario. Each scenario contains 20--21 sessions and 395--563 User-Assistant turns, establishing substantial longitudinal interaction histories. Such extended dialogue trajectories require models to maintain information across temporally separated sessions, as relevant evidence may originate well before the current interaction. 
The balanced scenario-level distribution further ensures broad coverage without disproportionate reliance on individual scenarios.

\textbf{Realistic Interaction Patterns.}
Fig. \ref{fig:data_statistic}(b) presents the distribution of user and assistant message lengths, measured in whitespace-delimited words. Assistant messages are considerably longer and more variable than user messages. Specifically, user messages have a mean, median, and 90th-percentile length of 60.7, 58, and 82 words, respectively, compared with 264.5, 270, and 306 words for assistant messages. The maximum lengths are 246 and 771 words, respectively. 
This pronounced asymmetry reflects the interaction pattern of real-world assistant usage: users typically express concise requests or knowledge-seeking queries, whereas assistants provide more elaborate explanations and multi-step responses. These heterogeneous message patterns contribute to realistic conversational contexts for memory evaluation.

\textbf{Comprehensive Task Coverage.}
Fig. \ref{fig:data_statistic}(c) presents the distribution of seven QA categories across each scenario, with all seven categories consistently represented. This consistent coverage ensures systematic evaluation across diverse referential reasoning capabilities and prevents the benchmark from being dominated by particular scenarios or task types. Notably, Referential Understanding (RU-RU and RU-TCU) accounts for a substantial proportion of the QA pairs, reflecting MemRR's emphasis on understanding referential structures and their semantic evolution beyond isolated entity identification.

\textbf{Multi-Session Evidence Integration.}
Fig. \ref{fig:data_statistic}(d) quantifies the number of distinct sessions containing supporting evidence for each QA pair. Although 40.4\% of questions can be answered using evidence from a single session, the majority (59.6\%) require integrating multimodal information distributed across multiple sessions. Moreover, 15.5\% depend on evidence spanning five or more sessions, with the evidence distribution extending to as many as 12 sessions. This distribution reveals substantial cross-session dependencies, requiring models to preserve and ground historical information over extended temporal intervals and integrate multimodal evidence dispersed throughout the dialogue history, rather than relying on single context alone. Such dependencies distinguish MemRR from benchmarks centered primarily on localized retrieval and highlight the importance of long-term memory organization and cross-modal evidence aggregation.

\begin{table*}[t]
    \centering
\caption{Implementation details of frontier MLLMs in full-context setting. The table lists the max length, visual usage, specific image resolution thresholds and the average input length for each model.}
\label{tab_6}
\resizebox{\textwidth}{!}{
\begin{tabular}{c|cccc}
\toprule

Models & Max length	& Visual Usage &	Image Threshold	& Total input tokens \\
\midrule

Claude-Sonnet-4.6 \cite{claude2026} & 1M & Image & 960	 & 401876.89  \\

Gemini-3.1-Pro-Preview \cite{gemini32026}  &1M &	Image&	1024	& 331943.00   \\

GPT-5.5 \cite{singh2025openai} & 1M	& Image	& 1024 &	296143.33 \\

Kimi-K3 \cite{team2026kimi} & 1M	& Image	& 1024	& 311941.80   \\

Qwen3.7-Plus \cite{qwen32026}
&1M	&Image	&1024	&291919.40   \\

Deepseek-V4-Pro \cite{deepseekai2026deepseek}&
1M&	Caption only	&-	&209781.70   \\

GLM-5.2 \cite{glm2026} &
1M	& Caption only &	-	& 209505.40   \\

Gemini-2.5-Flash \cite{gemini22025}&
1M	&Image	&1024	&249113.88  \\

\bottomrule
\end{tabular}
}
\end{table*}

\subsection{Data Source}

The images in MemRR are carefully selected from five publicly available datasets: Memory QA \cite{jiang-etal-2025-memory}, OmniSpatial \cite{jia2025omnispatial}, PASLCD \cite{galappaththige2025multi}, RefSpatial \cite{zhou2025roborefer}, and RefCOCO \cite{yu2016modeling}. Selection prioritizes rich semantic content and multiple visual entities to provide diverse visual anchors for cross-modal reference and long-term entity tracking. The resulting collection spans broad scenes and visual contexts, including indoor and outdoor environments, offices, restaurants, shops, natural landscapes, everyday objects, human activities, animals, books, and screenshot-based interfaces. This diversity provides MemRR broad coverage of visual entities, spatial configurations, and semantic relationships, supporting the evaluation of referential reasoning under varied real-world visual conditions.

The image collection also exhibits considerable resolution diversity while retaining rich visual detail. Image resolutions range from $338 \times 254$ to $3024 \times 4032$ pixels, with an average width of 2,314.7 pixels and an average height of 2,878.15 pixels. Notably, 74.3\% of images have both dimensions exceeding 1,024 pixels, while 65.23\% have resolutions exceeding $3,000 \times 3,000$ pixels. All images preserves fine-grained visual details, including object attributes, spatial configurations, and embedded textual information. Such details are essential for resolving visually grounded references and establishing entity-level correspondences across temporally dispersed interactions.

\subsection{Evaluation Dimension}

MemRR establishes a long-term interactive evaluation framework centered on cross-modal referential reasoning, assessing multimodal memory methods along three dimensions: \textit{Reference Resolution}, \textit{Referential Understanding}, and \textit{Semantic Inference}. Together, these dimensions progressively evaluate the ability to identify historical referents, understand their evolving semantic structures, and infer latent user knowledge from accumulated visual-linguistic referring interactions.

\textbf{Reference Resolution (RR).}
This dimension evaluates the ability to identify and recover historical entities, events, and scenes by establishing fine-grained alignment among expressions, visual content, and contextual knowledge. As referents are typically introduced implicitly through images and described using non-identical expressions across interactions, successful resolution requires semantic integration or multi-hop cross-modal reasoning over distributed pieces of multimodal evidence beyond lexical matching. 
It comprises three sub-dimensions:

\begin{itemize}
    \item \textbf{Instance Resolution (RR-IR).}
    Evaluates the ability to identify the identities, attributes, and states of referenced entities throughout long-term interactions.
    Since entities are grounded in visual scenes without explicit textual descriptions, models must establish cross-modal alignment between expressions and visual instances, and perform cross-turn reference resolution to recover its attributes and states from multiple pieces of multimodal evidence, even in the absence of lexical overlap.

    \item \textbf{Event Resolution (RR-ER).}
    Evaluates the ability to reconstruct historical events and recover their essential details, including content, participants, and outcomes. It requires integrating multimodal evidence across consecutive or temporally dispersed turns to identify event-specific facts, establish local event-level associations, and associate relevant details with the corresponding entities. Beyond isolated fact retrieval, this sub-dimension assesses the reconstruction of event semantics from distributed conversational evidence.

    \item \textbf{Situational Resolution (RR-SR).}
    Evaluates the ability to reconstruct the context related to a historical entity or event, including its temporal and spatial attributes and  environment background. It requires integrating visual and linguistic evidence to recover the broader spatiotemporal context in which a referent was occurred, rather than merely identifying the referent itself.
    
\end{itemize}

\textbf{Referential Understanding (RU).}
This dimension evaluates the ability to understand structured relationships among historical entities and events and maintain their semantic consistency as interactions evolve. It requires organizing accumulated multimodal information into coherent relational representations, maintaining consistent representations of entities, events, and user knowledge, while tracking evolving associations in attributes and states as multimodal information accumulates across sessions. It comprises two sub-dimensions:

\begin{itemize}
    \item \textbf{Relational Understanding (RU-RU).}
    Evaluates the ability to identify, integrate, and reason over structured relationships among entities, events throughout long-term interactions. These relationships encompass spatial, temporal, hierarchical, causal, and comparative associations, which continuously emerge or evolve as new interactions progress. The associated tasks involve either integrating multiple pieces of cross-modal evidence within a single session or aggregating evidence across sessions to establish reason relationships among entities or events. This sub-dimension primarily assesses structured memory organization, relational knowledge construction, and multi-hop reasoning over historical information.

    \item \textbf{Temporal Consistency Understanding (RU-TCU).}
    Evaluates the ability to maintain entity and event consistency while tracking their semantic evolution across long-term interactions. 
    Entity attributes and states, as well as event participants, details, and completion status, may change over time, while their underlying identities remain consistent. Moreover, the same entity or event may be referred to through different expressions across sessions and contexts, while preserving its underlying identity.
    It requires resolving long-range evolving references and consolidating temporally distributed evidence. 
    This sub-dimension emphasizes cumulative cross-session reasoning, including entity deduplication, event consolidation, temporal ordering, numerical aggregation, and attribute integration.

\end{itemize}

\textbf{Semantic Inference (SI).}
This dimension evaluates the semantic reasoning ability to infer preferences and behavioral intentions from accumulated referring history. Rather than relying solely on explicitly stated information, it requires abstracting implicit semantic patterns from user choices, feedback, task-related interactions, and contextual cues to support personalized understanding and decision-making. It comprises two sub-dimensions:

\begin{itemize}
    \item \textbf{Preference Inference (SI-PI).}
    Evaluates the ability to infer implicit user preferences by aggregating referential expressions and behavior evidences across multiple sessions. Rather than relying on explicitly declared preferences, models must understand the user intention from referential expressions and identify recurring patterns and consolidate them into stable user knowledge. 

    \item \textbf{Behavior Inference (SI-BI).}
    Evaluates the ability to infer users' underlying goals and likely subsequent actions from referential and behavioral evidence. It requires integrating multiple interaction cues and muti-step referential resolution to understand ongoing activities and anticipate next steps or decisions, thereby moving beyond explicit information retrieval toward intention-aware reasoning.

\end{itemize}

\begin{figure}
    \centering
    \includegraphics[width=\linewidth]{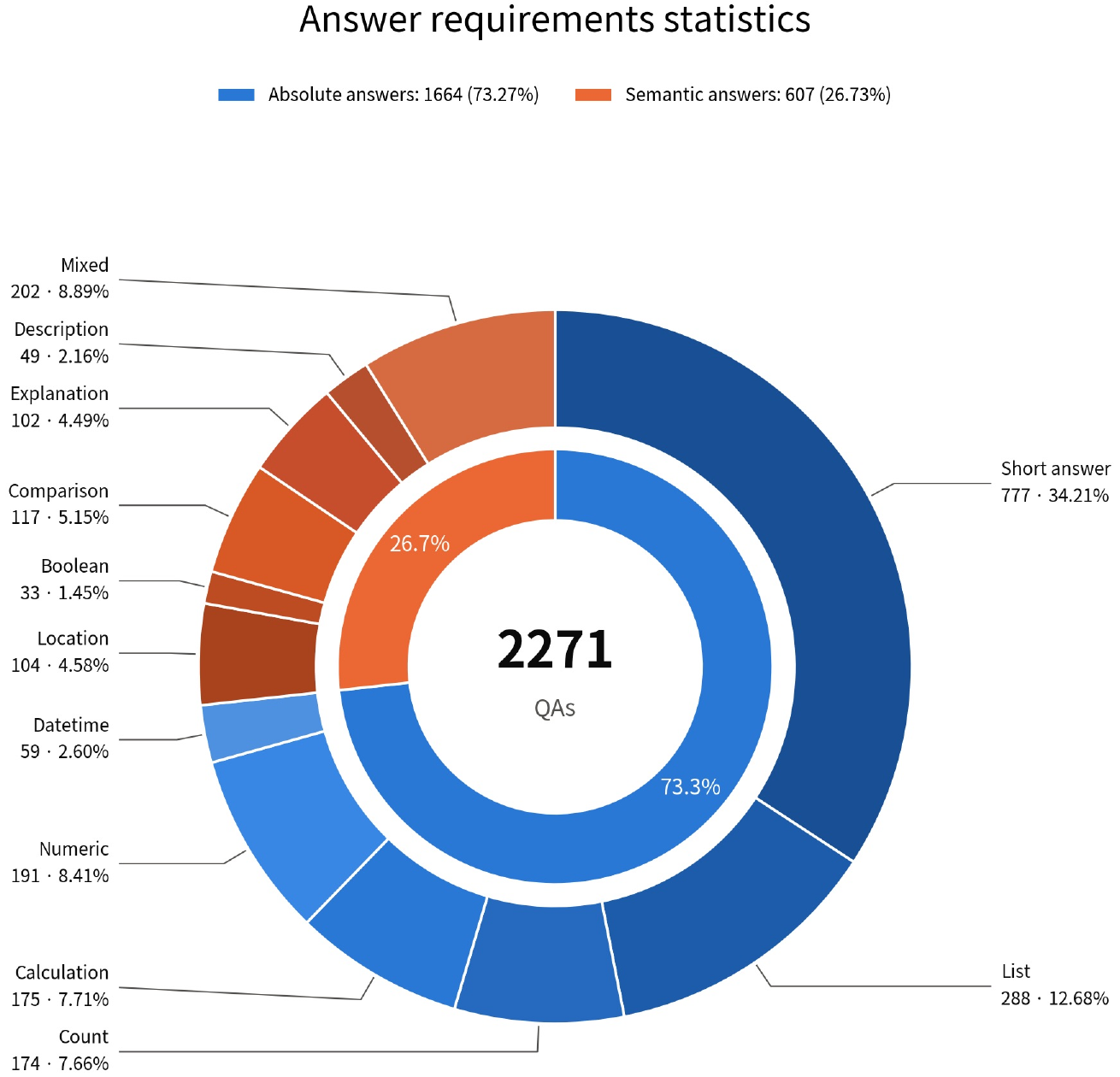}
    \caption{QA-type composition from the perspective of expected answer formats. It shows the number and share of the 2271 QAs across different answer formats.}
    \label{fig_8}
\end{figure}

\begin{table*}[t]
\centering
\caption{Agreement of each judge with human verification on 50 sampled items per task type. $r_{con}$ denotes agreement (\%) with human verification. $P/N$ denotes positive/negative split as scored by each judge. Numbers are
close across judges, indicating that LLM-as-a-Judge is stable across judge models.}
\label{tab:human_validation}

\setlength{\tabcolsep}{4pt}
\renewcommand{\arraystretch}{1.15}

\resizebox{\textwidth}{!}{
\begin{tabular}{c|ccc|ccc|ccc|ccc|ccc|ccc|ccc}
\toprule


\multirow{3}{*}{\textbf{Judge}}
& \multicolumn{9}{c|}{Referential Resolution}
& \multicolumn{6}{c|}{Referential Understanding}
& \multicolumn{6}{c|}{Semantic Inference}  \\
& \multicolumn{3}{c|}{\textbf{RR-IR}}
& \multicolumn{3}{c|}{\textbf{RR-ER}}
& \multicolumn{3}{c|}{\textbf{RR-SR}}
& \multicolumn{3}{c|}{\textbf{RU-RU}}
& \multicolumn{3}{c|}{\textbf{RU-TCU}}
& \multicolumn{3}{c|}{\textbf{SI-BI}}
& \multicolumn{3}{c|}{\textbf{SI-PI}} \\

& Score & $r_{con}$ & $P/N$ 
& Score & $r_{con}$ & $P/N$
& Score & $r_{con}$ & $P/N$
& Score & $r_{con}$ & $P/N$
& Score & $r_{con}$ & $P/N$
& Score & $r_{con}$ & $P/N$
& Score & $r_{con}$ & $P/N$  \\

\midrule

Human & 0.785 & - & 13/47 & 0.56 & - & 28/22 & 0.63 & - & 21/29 & 0.73 & - & 17/33 & 0.38 & - & 10/40 & 0.575 & - & 26/24 & 0.86 & - & 9/41 \\

\midrule

GPT-5.2 & 0.78 & 98\% & 14/46 & 0.57 & 98\% & 27/23 & 0.65 & 96\% & 20/30 & 0.73 & 100\% & 17/33 & 0.38 & 92\% & 12/38 & 0.61 & 92\% & 23/27 & 0.85 & 94\% & 11/39  \\

Qwen3.5-122B-A10B & 0.8 & 92\% & 14/46 & 0.55 & 92\% & 29/21 & 0.63 & 100\% & 21/29 & 0.72 & 94\% & 16/34 & 0.39 & 98\% & 11/39 & 0.57 & 98\% & 27/23 & 0.86 & 92\% & 11/39 \\

Average & 0.79 & 95\% & - & 0.56 & 95\% & - & 0.64 & 98\% & - & 0.725 & 97\% & - & 0.385 & 95\% & - & 0.59 & 95\% & - & 0.855 & 93\% & - \\

\bottomrule
\end{tabular}
}
\end{table*}

\subsection{Evidence Distribution}

To further characterize the historical evidence dependencies across QA categories, we analyze evidence-session distributions from two complementary perspectives: \textit{temporal span} and \textit{temporal dispersion}, as illustrated in Fig. \ref{fig:evidence_distribution}. Temporal span quantifies how broadly the required evidence 
extends over the interaction history, whereas temporal dispersion characterizes the distributional inequality of supporting evidence within that span. Together, these metrics reveal distinct temporal evidence structures underlying different cross-modal referential reasoning capabilities. Table~\ref{tab_5} further reports the proportion of multi-session evidence QA pairs and their distribution across categories.

\textbf{Temporal Span.}
Fig. \ref{fig:evidence_distribution}(a) characterizes the temporal coverage of evidence sessions required by different QA categories, and Table \ref{tab_5} shows the mean values. For each question $\hat{q}_i$, we define its temporal span as the distance between the earliest and latest evidence sessions:

\begin{equation}
    \mathrm{Span}(\hat{q}_i)
    = \max(E(\hat{q}_i)) - \min(E(\hat{q}_i)),
\end{equation}
where $E({\hat{q}_i})$ denotes the set of evidence session indices for $\hat{q}_i$. A larger span indicates that answering the question requires accessing information distributed across a broader portion of the interaction history.

The results reveal distinct temporal dependency patterns across evaluation dimensions. Reference Resolution tasks (RR-IR, RR-ER, and RR-SR) exhibit predominantly localized evidence dependencies, with 87.5\%, 88.8\%, and 90.9\% of QA pairs, respectively, having a span of zero, indicating that their supporting evidence is confined to a single session. 
By contrast, Referential Understanding tasks (RU-RU and RU-TCU) exhibit broader temporal dependencies, with 32.4\% and 35.6\% of QA pairs, respectively, requiring evidence spanning 1--5 sessions. Notably, approximately 10\% of these questions exhibit spans of 16--20 sessions, revealing dependencies between information introduced far apart in the interaction history. Semantic Inference tasks (SI-BI and SI-PI) exhibit intermediate patterns, with 45.6\% and 39.6\% of QA pairs, respectively, having a non-zero span. These findings indicate that entity-, event-, and scene-level reference resolution predominantly relies on localized evidence, whereas relational and consistency understanding and latent semantic inference increasingly require the integration of temporally distant information.

\textbf{Temporal Dispersion.}
Fig. \ref{fig:evidence_distribution}(b) further characterizes the distribution of supporting evidence within its temporal span, and Table \ref{tab_5} shows the mean values. Specifically, we adopt the Gini coefficient over the intervals between consecutive evidence sessions to quantify the temporal inequality of evidence distribution. For each multi-session evidence question $\hat{q}_i$, let $E(\hat{q}_i) = \{S_{i_1},\ldots,S_{i_k}\}$ denote its chronologically ordered evidence session indices, and let $G(\hat{q}_i) = [g_{i_1},\ldots, g_{i_r}]$ represent the the set of corresponding temporal gaps, where $g_{i_j}=S_{i_{j+1}}-S_{i_j}$ and $r=k-1$. Temporal dispersion is defined as:

\begin{equation}
    D(\hat{q}_i) =
    \frac{
    \sum_{a=1}^{r}\sum_{b=1}^{r}|g_{i_a}-g_{i_b}|
    }{
    2r\sum_{a=1}^{r}g_{i_a}
    },
\end{equation}
where $D(\hat{q}_i)\in[0,1]$ measures the inequality of temporal gaps. Lower values indicate relatively uniform spacing between evidence sessions, whereas higher values indicate more uneven distributions, with evidence concentrated within certain periods and separated by larger temporal gaps. Single-session QA pairs are excluded from this calculation, as they contain no inter-session gaps.

The results further distinguish localized reference resolution from higher-level understanding and inference. Reference Resolution predominantly relies on single-session evidence, resulting in more localized evidence dependencies. In contrast, Referential Understanding and Semantic Inference exhibit more dispersed cross-session evidence patterns, suggesting that their supporting information is not only distributed across broader temporal ranges but may also be unevenly separated throughout the interaction history. 
In particular, the non-uniform distribution of evidence gaps indicates that supporting information may be concentrated in certain periods while remaining separated by substantial temporal intervals. This requires models to retrieve and integrate historically distant evidence rather than rely on a single contiguous context window. 
Together, temporal span and dispersion demonstrate that different referential reasoning capabilities involve substantially different historical evidence organization patterns, with higher-level referential understanding and semantic inference entail more complex temporal dependencies.

\begin{table*}[t]
\centering
\caption{Partial parameters of all memory agents being evaluated. The table lists the use of visual information, text and visual backbones, as well as the key parameters involved for each method.}
\label{tab_parameter}

\setlength{\tabcolsep}{4pt}
\renewcommand{\arraystretch}{1.15}

\resizebox{\textwidth}{!}{
\begin{tabular}{c|cccccc}
\toprule


Methods & Type & Visual Usage & Text Backbone & Visual Backbone	& Parameters  & Value \\


\midrule
\rowcolor{blue!7}
\multicolumn{7}{c}{\textit{\textbf{Text-based Agentic-Memory methods}}} \\

NaiveRAG & Text-based & Caption only & Qwen3-VL-Embedding-8B & $-$ & retrieve\_top\_k & 20 \\

\multirow{2}{*}{A-Mem \cite{xu2025mem}} & \multirow{2}{*}{Text-based} & \multirow{2}{*}{Caption only} & \multirow{2}{*}{All-MiniLM-L6-v2} &	\multirow{2}{*}{$-$} & neighbor\_k	 & 10 \\
 & & & & & retrieve\_k & 20 \\

\multirow{6}{*}{SimpleMem \cite{liu2026simplemem}} & \multirow{6}{*}{Text-based} & \multirow{6}{*}{Caption only} & \multirow{6}{*}{Qwen3-Embedding-4B} & \multirow{6}{*}{$-$} & window\_size	& 20 \\
 & & & & & overlap\_size & 5 \\
 & & & & & semantic\_top\_k & 25 \\
& & & & & keyword\_top\_k	 & 10 \\
& & & & & structured\_top\_k	& 5 \\
& & & & & max\_reflection\_rounds	& 3 \\

\multirow{5}{*}{MRAgent \cite{ji2026memory}} & \multirow{5}{*}{Text-based} &	\multirow{5}{*}{Caption only} & \multirow{5}{*}{Text-embedding-3-small} & \multirow{5}{*}{-} & tag\_max & 15 \\
& & & & & tag\_limit & 10 \\
& & & & & topic\_k & 10 \\
& & & & & max\_rounds & 5 \\
& & & & & max\_tool\_calls & 50 \\


\midrule
\rowcolor{red!5}
\multicolumn{7}{c}{\textit{\textbf{Multimodal Agentic-Memory methods}}} \\

MuRAG \cite{chen2022murag} & Multimodal	& Raw image	& Qwen3-VL-Embedding-8B & Qwen3-VL-Embedding-8B & retrieve\_top\_k & 20 \\

\multirow{2}{*}{UniversalRAG \cite{yeo2026universalrag}} & \multirow{2}{*}{Multimodal} & 	\multirow{2}{*}{Raw image}	& \multirow{2}{*}{Qwen3-VL-Embedding-8B} & \multirow{2}{*}{Qwen3-VL-Embedding-8B} & retrieve\_top\_k & 20 \\
& & & & & mode & no/document/image \\

\multirow{3}{*}{NGM \cite{fisher2025neural} } & \multirow{3}{*}{Multimodal} &\multirow{3}{*}{Raw image} & \multirow{3}{*}{Qwen3-VL-Embedding-8B} & \multirow{3}{*}{Qwen3-VL-Embedding-8B} & max\_edges\_per\_node & 10\\
& & & & & recall\_max\_depth & 3 \\
& & & & & recall\_max\_nodes & 20 \\

\multirow{3}{*}{Omni-SimpleMem \cite{liu2026omni}} &  \multirow{3}{*}{Multimodal}	& \multirow{3}{*}{Image\&Caption} & \multirow{3}{*}{Text-embedding-3-small} & \multirow{3}{*}{Openvision-vit-large-patch14-224} & retrieve\_top\_k	& $>$20(dynamic)\\
& & & & & max\_summaries\_in\_context & 50\\
& & & & & bm25\_extra & 15 \\

\multirow{4}{*}{V-Mem \cite{kang2026v}} & \multirow{4}{*}{Multimodal}	& \multirow{4}{*}{Image\&Caption} & \multirow{4}{*}{all-MiniLM-L6-v2} & \multirow{4}{*}{siglip2-base-patch16-384} & retrieve\_top\_k & 15 \\
& & & & & dense\_k & 20 \\
& & & & &  sparse\_k & 	20 \\
& & & & & window & 2 \\

\multirow{3}{*}{RAG-Anything \cite{guo2025rag}} &  \multirow{3}{*}{Multimodal}	& \multirow{3}{*}{Image\&Caption} & \multirow{3}{*}{Text-embedding-3-small} & \multirow{3}{*}{LLM model} & retrieve\_mode & hybrid \\
& & & & & context\_mode & chunk \\
& & & & & context\_window & 2 \\

\bottomrule
\end{tabular}
}
\end{table*}

\subsection{QA Statistics}
As shown in Fig. \ref{fig_8}, we further categorize QA pairs according to their expected answer formats and evaluation criteria into two groups: \textit{Absolute-answer} and \textit{Semantic-answer} questions, accounting for 73.27\% and 26.73\% of the benchmark, respectively. This categorization captures the diverse answer requirements in MemRR, and the need for evaluation protocols that accommodate both factual  precision and semantic consistency.

\textbf{Absolute-answer QAs} require specific, verifiable answers, typically expressed as keywords, entity names, or numerical and temporal values. Their evaluation primarily emphasizes the accuracy and completeness of essential factual information. This group comprises six answer types: \textit{Short Answer} (34.21\%), requiring concise identification of names, facts, or attributes; \textit{List} (12.68\%), requiring the enumeration of required items or entity--attribute pairs; \textit{Count} (7.66\%), requiring the number of occurrences or distinct items; \textit{Calculation} (7.71\%), involving numerical operations such as computing totals, differences, averages, and temporal intervals; \textit{Numeric} (8.41\%), requiring specific quantitative values, such as prices, dimensions, capacities, or caloric content; and \textit{DateTime} (2.60\%), requiring specific dates or temporal periods. Collectively, these tasks assess the precise preservation of fine-grained historical facts, with particular emphasis on exact values and key information.

\textbf{Semantic-answer QAs} require descriptive, comparative, or interpretative responses, for which semantic equivalence and contextual completeness take precedence over exact lexical matching. They comprise six answer types: \textit{Location} (4.58\%), requiring the identification of absolute or relative positions, directions, or routes; \textit{Boolean} (1.45\%), determining whether an event occurred or a proposition holds, potentially with supporting justification; \textit{Comparison} (5.15\%), reasoning about differences, temporal ordering, or relative extrema; \textit{Explanation} (4.49\%), articulating underlying reasons or summarizing inferred user preferences; \textit{Description} (2.16\%), characterizing visual appearances, scenes, or processes; and \textit{Mixed} (8.89\%), integrating multiple answer components, such as quantities with entity names, dates with temporal differences, or names with their descriptions. These tasks emphasize contextual completeness, overall semantic consistency, and the correct integration of multiple information.

Overall, the answer-type distribution demonstrates that MemRR evaluates both fine-grained factual accuracy and higher-level semantic reasoning. While absolute-answer questions primarily assess the precise recovery of historical details, semantic-answer questions further require contextual interpretation and the coherent synthesis of distributed multimodal evidence. This diversity motivates rubric-based LLM-as-a-Judge evaluation, which accommodates semantic equivalence and answer completeness beyond exact string matching, while supporting the assessment of heterogeneous answer requirements.

\begin{table*}[t]
\centering
\caption{Overall performance on MemRR. All memory agents are implemented with GPT-4o-mini \cite{openai2024gpt} as base model. Best and second-best are marked in bold and underline, and the agentic-memory methods are ranked together.}
\label{tab_gpt-4o-mini}

\setlength{\tabcolsep}{4pt}
\renewcommand{\arraystretch}{1.15}

\resizebox{\textwidth}{!}{
\begin{tabular}{c|ccccccccc}
\toprule


Methods & Metrics & RR-IR & RR-ER & RR-SR & RU-RU & RU-TCU & SI-BI & SI-PI & Overall \\


\midrule
\rowcolor{blue!10}
\multicolumn{10}{c}{\textit{\textbf{Text-based Agentic-Memory methods}}} \\

\multirow{2}{*}{NaiveRAG} & F1 & 0.0685  &0.0637  &0.0880  & 0.1873   & 0.0508  & \underline{0.1301}   & \underline{0.2159} 	 & 0.1003  \\

& LLM-Judge  & 0.1545  & 0.1283  &	0.1976  &	0.1626 	 &0.0787  &	0.3267  &	0.5524  & 0.1589  \\

\multirow{2}{*}{A-Mem \cite{xu2025mem}} & F1& 0.0568 &	0.0536 & 0.0814 & 0.1304 & 0.0415 &	0.1049 & 0.1962 & 0.0790  \\
 & LLM-Judge &  0.1066  & 	0.1114  & 	0.1674  & 	0.1256  & 	0.0721  & 	0.2643  & 	0.5158  & 	0.1332  \\

\multirow{2}{*}{SimpleMem \cite{liu2026simplemem}}& F1 & 
0.0908 &	\underline{0.1200} &	0.0968 &	0.1498 &	\underline{0.1041} 	&0.1168  & 0.1678 &	0.1182  \\
& LLM-Judge  &0.1209  &	0.0917  &	0.1596  &	0.1420 	 &0.0897  & 0.2418  &	0.3901  &	0.1417  \\

\multirow{2}{*}{MRAgent \cite{ji2026memory}} & F1 & 
\textbf{0.1479} & 	\textbf{0.1399} & 	\textbf{0.1249} & 	0.1426 & 	\textbf{0.1246} 	& 0.1016 & 	0.1785 & 	\textbf{0.1307} \\
& LLM-Judge & 0.1715 &	\underline{0.1469} &	0.2259 &	0.1477 	&\textbf{0.1332} & 0.2470 &	0.4130 &	0.1720  \\


\midrule
\rowcolor{red!5}
\multicolumn{10}{c}{\textit{\textbf{Multimodal Agentic-Memory methods}}} \\

\multirow{2}{*}{MuRAG \cite{chen2022murag}} & F1 &  0.0832 	&0.0685 & 0.0862 &	0.1824 &	0.0514 &	0.1295 &	0.2129 &	0.1029  \\
& LLM-Judge & 0.1761 &	0.1225 &	0.1994 &	0.1664 	&0.0927 &	\textbf{0.3465} &	\underline{0.5565} &	0.1712  \\

\multirow{2}{*}{UniversalRAG \cite{yeo2026universalrag}} & F1 & 0.1096 &	0.0765 &	\underline{0.1052} &	\textbf{0.2189} &	0.0644 	&\textbf{0.1435} &	0.2108 & 	\underline{0.1215}  \\
& LLM-Judge & \underline{0.2161} &	0.1334 	& \underline{0.2325} &	0.2000 &	0.1137 &	\underline{0.3440} &	\textbf{0.5943} &	\underline{0.1950}  \\

\multirow{2}{*}{NGM \cite{fisher2025neural} } & F1& 0.0466 	&0.0392 &	0.0635 &	0.0621 &	0.0303& 	0.1049 	&0.1720 &	0.0541  \\
& LLM-Judge &  0.0912 &	0.0618 &	0.1439 	&0.0732 	&0.0464 &	0.2577& 	0.4473 	&0.0997  \\

\multirow{2}{*}{Omni-SimpleMem \cite{liu2026omni}} & F1 &  0.0630 & 	0.0475 & 	0.0844 & 	0.1452 & 	0.0427 	& 0.1266 	& \textbf{0.2309} & 	0.0870  \\
& LLM-Judge & 0.1044 &	0.0974 &	0.1623 	&0.1399 &	0.0817 &	0.2704 &	0.5183 &	0.1382 \\

\multirow{2}{*}{V-Mem \cite{kang2026v}} & F1 &  \underline{0.1131} 	&0.0786 &	0.0820 &	0.1740 &	0.0545 &	0.0925 	&0.1575 &	0.0984 \\
& LLM-Judge & 0.1969 &	0.1507 &	0.1947 	&0.1693 &	0.1121 &	0.2855& 	0.5338 &	0.1724  \\

\multirow{2}{*}{RAGanything \cite{guo2025rag}} & F1 &  0.0736 &	0.0675 &	0.0703 &	0.1203 &	0.0474 	&0.0606 	&0.0864 	&0.0739 \\
& LLM-Judge & \textbf{0.3013} &	\textbf{0.2144} &	\textbf{0.2814} &	\textbf{0.2008} 	&\underline{0.1237} &	0.3089 &	0.5534 &	\textbf{0.2195}  \\

\bottomrule
\end{tabular}
}
\end{table*}

\begin{figure*}[h]
	
	\begin{minipage}{0.49\linewidth}
		\centerline{\includegraphics[width=\textwidth]{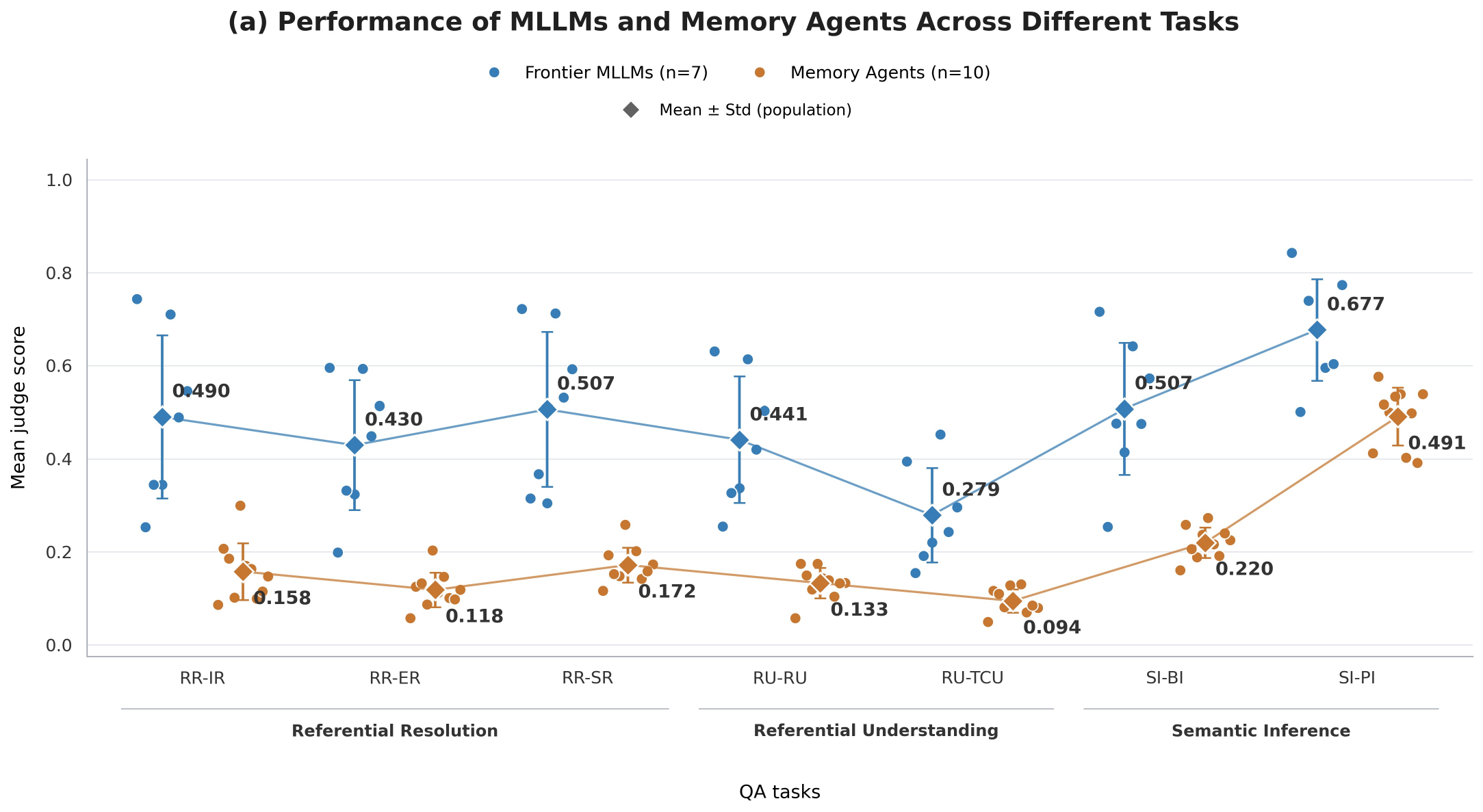}}
	\end{minipage}
	\begin{minipage}{0.49\linewidth}
		\centerline{\includegraphics[width=\textwidth]{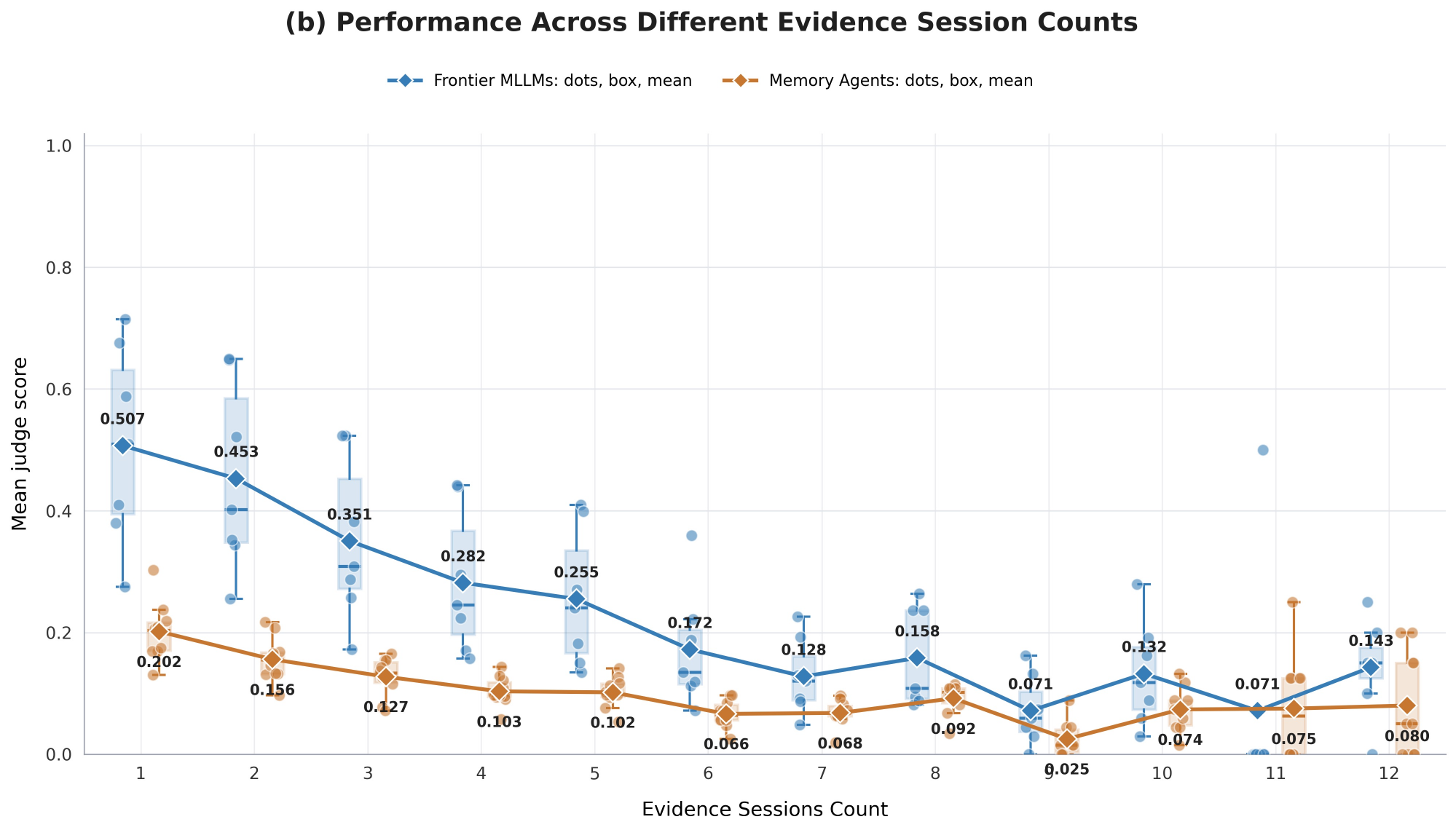}}
	 
	\end{minipage}
 
	\caption{Performance of MLLMs and memory agents with GPT-4o-mini \cite{openai2024gpt} as base model across QA tasks (a) and evidence session counts (b).}
	\label{fig_gpt-4o-mini}
\end{figure*}

\section{Experimental Details}
\label{appendix_B}

\subsection{Details of MLLMs}

\textbf{Models.} 
We evaluate the following frontier closed-source MLLMs as answering models. MLLMs that suppots visual inputs include:  Claude-Sonnet-4.6 \cite{claude2026}, which is Anthropic’s reasoning model; GPT-5.5 \cite{singh2025openai}, which is one of the OpenAI’s strongest general-purpose multimodal model; Gemini-3.1-Pro \cite{gemini32026}; which is Google’s native multimodal reasoning model for complex tasks; Kimi-K3 \cite{team2026kimi}; which is Moonshot AI's latest long-range agents performing inference on massive inputs; Qwen3.7-Plus \cite{qwen32026}, which supports multimodal deep inference from Qwen Team. We also evaluate two models that support text-only input include: Deekseek-V4-Pro \cite{deepseekai2026deepseek}, which is DeepSeek's flagship model for large-scale Mixed Experts; GLM 5.2 \cite{glm2026}, which is the flagship model for the era of long tasks from Z.ai.
Except Claude-Sonnet-4.6 \cite{claude2026}, all API-served models are accessed through OpenAI-compatible remote endpoints and exposed to the same function-calling interface, while Claude-Sonnet-4.6 is accessed Anthropic endpoint.
For the full agentic-memory methods experiments, we additionally evaluate all memory systems on four base model spanning families, including Gemini-2.5-Flash \cite{gemini22025}, GPT-4o-mini \cite{openai2024gpt}, Qwen3-VL-Plus\cite{bai2025qwen3}. 

\begin{table*}[t]
\centering
\caption{Overall performance on MemRR. All memory agents are implemented with Qwen3-VL-Plus \cite{bai2025qwen3} as base model. Best and second-best are marked in bold and underline, and the agentic-memory methods are ranked together.}
\label{tab_qwen3-vl-plus}

\setlength{\tabcolsep}{4pt}
\renewcommand{\arraystretch}{1.15}

\resizebox{\textwidth}{!}{
\begin{tabular}{c|ccccccccc}
\toprule


Methods & Metrics & RR-IR & RR-ER & RR-SR & RU-RU & RU-TCU & SI-BI & SI-PI & Overall \\


\midrule
\rowcolor{blue!10}
\multicolumn{10}{c}{\textit{\textbf{Text-based Agentic-Memory methods}}} \\

\multirow{2}{*}{NaiveRAG} & F1 & 0.0579 &	0.0445 	&0.0983 &	0.0832 &	0.0344 &	\underline{0.1124} &	0.1553 &	0.0649   \\

& LLM-Judge & 0.1097 &	0.0996 	&0.2045 &	0.1315 &	0.0759 &	\textbf{0.3094} &	0.6103 &	0.1459   \\

\multirow{2}{*}{A-Mem \cite{xu2025mem}} & F1 & 0.0689 &	0.0494 	&0.0960 &	0.0697 &	0.0262 	&0.0980 &	0.1714 	&0.0603   \\
 & LLM-Judge &  0.1149 	&0.0947 &	0.1945 &	0.1059 	&0.0653 &	0.2779 &	\underline{0.6146} &	0.1341   \\

\multirow{2}{*}{SimpleMem \cite{liu2026simplemem}}& F1 & 
\underline{0.1085} &	\underline{0.0704} &	0.1123 &	\underline{0.1478} &	\underline{0.0590} &	0.1119 &	0.1511 &	\underline{0.0985}   \\
& LLM-Judge & 0.1321 &	0.1146 &	0.1657 &	\underline{0.1385} &	\underline{0.0999} & 0.2311 &	0.4211 &	0.1462   \\

\multirow{2}{*}{MRAgent \cite{ji2026memory}} & F1 & 
\textbf{0.1543} &	\textbf{0.1290} &	\textbf{0.1600} &	\textbf{0.1543 }&	\textbf{0.1463} &	0.0984 	&\textbf{0.2053} 	&\textbf{0.1436}  \\
& LLM-Judge & \textbf{0.1998} &	\textbf{0.1559} 	&\textbf{0.2711} &	\textbf{0.1809} &	\textbf{0.1416} &	0.2882 &	0.5115 &	\textbf{0.1951}   \\


\midrule
\rowcolor{red!5}
\multicolumn{10}{c}{\textit{\textbf{Multimodal Agentic-Memory methods}}} \\

\multirow{2}{*}{MuRAG \cite{chen2022murag}} & F1 &  0.0831 & 	0.0489 & 	0.1070 &	0.0810 	&0.0311 &	0.1094 &	0.1639 	&0.0677   \\
& LLM-Judge & 0.1762 &	0.1026 &	0.1974 &	0.1336 	&0.0752 &	0.3068 &	0.6084 &	0.1544   \\

\multirow{2}{*}{UniversalRAG \cite{yeo2026universalrag}} & F1 & 0.1011 & 	0.0685 &	0.1302 &	0.0827 &	0.0413 &	0.1014 &	0.1738 &	0.0794   \\
& LLM-Judge & \underline{0.1828} &	\underline{0.1444} 	& \underline{0.2479} &	0.1251 &	0.0853 &	0.2767 &	\textbf{0.6447} & \underline{0.1667}   \\

\multirow{2}{*}{NGM \cite{fisher2025neural} } & F1 & 0.0340 	&0.0345 &	0.0877 &	0.0512 &	0.0221 &	0.0908 &	0.1365 &	0.0454   \\
& LLM-Judge &  0.0641 &	0.0668 &	0.1528 &	0.0831 	&0.0462 	&0.2701 &	0.4700 & 	0.1002   \\

\multirow{2}{*}{Omni-SimpleMem \cite{liu2026omni}} & F1 &  0.0842 &	0.0529 & 	\underline{0.1336} &	0.0799 &	0.0375 &	\textbf{0.1402} &	\underline{0.1895} & 0.0756  \\
& LLM-Judge & 0.1246 &	0.0958 &	0.2309 &	0.1257& 	0.0782& 	\underline{0.2898} 	&0.5210 &	0.1397  \\

\multirow{2}{*}{V-Mem \cite{kang2026v}} & F1 &  0.0403 &	0.0274 	&0.0541 &	0.0444 &	0.0176 &	0.0469 	&0.0734 &	0.0335  \\
& LLM-Judge & 0.1261 &	0.0967 &	0.1691 &	0.0998 &	0.0532 	&0.1981 &	0.5830 &	0.1200   \\

\multirow{2}{*}{RAGanything \cite{guo2025rag}} & F1 &  0.0484 	& 0.0364 & 	0.0364 & 	0.0606 & 	0.0227 & 	0.0351 & 	0.0499&  	0.0387  \\
& LLM-Judge & 0.1246 &	0.0958 &	0.2309 	&0.1257 &	0.0782 	& \underline{0.2898} &	0.5210 	& 0.1397   \\

\bottomrule
\end{tabular}
}
\end{table*}

\begin{figure*}[h]
	
	\begin{minipage}{0.49\linewidth}
		\centerline{\includegraphics[width=\textwidth]{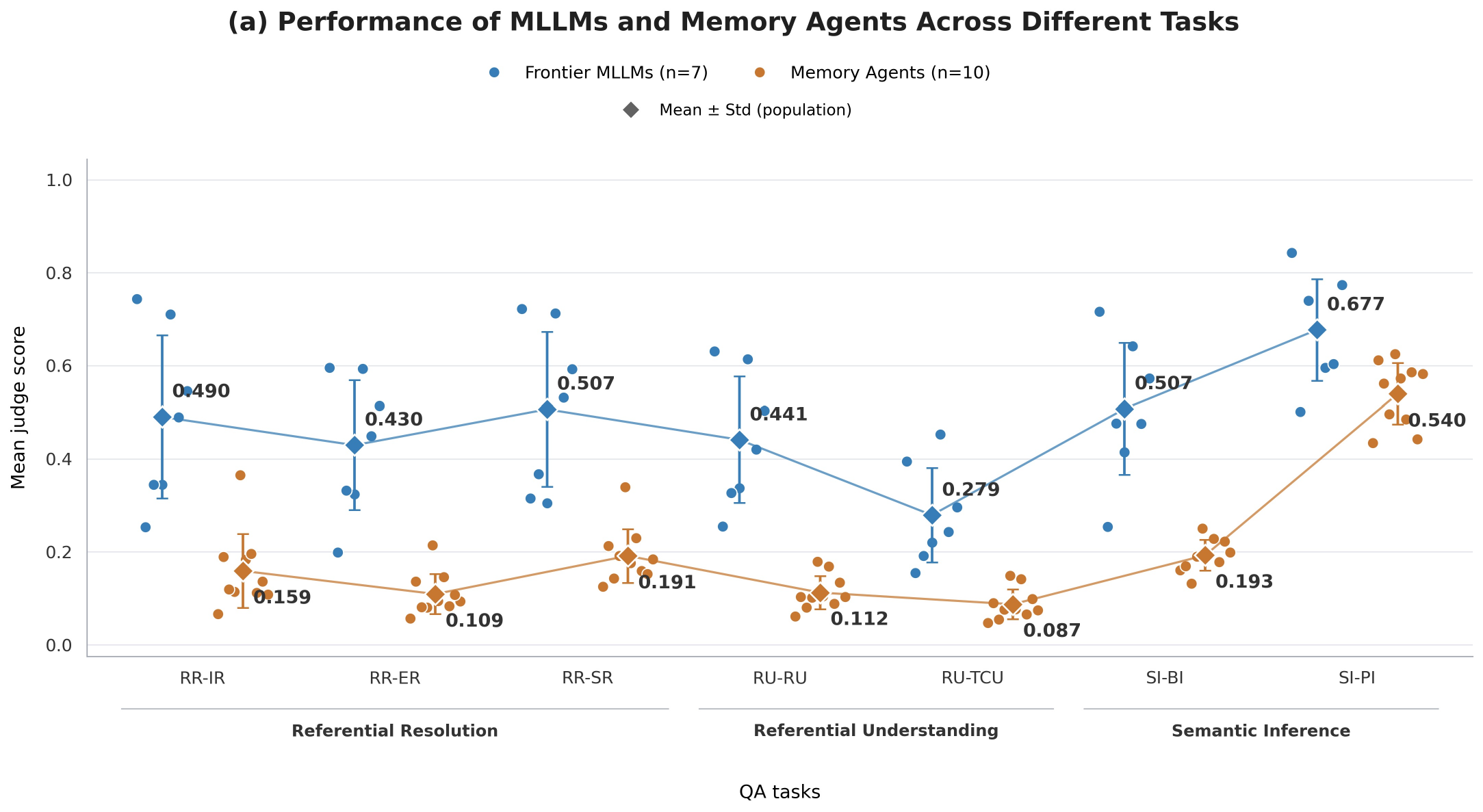}}
	\end{minipage}
	\begin{minipage}{0.49\linewidth}
		\centerline{\includegraphics[width=\textwidth]{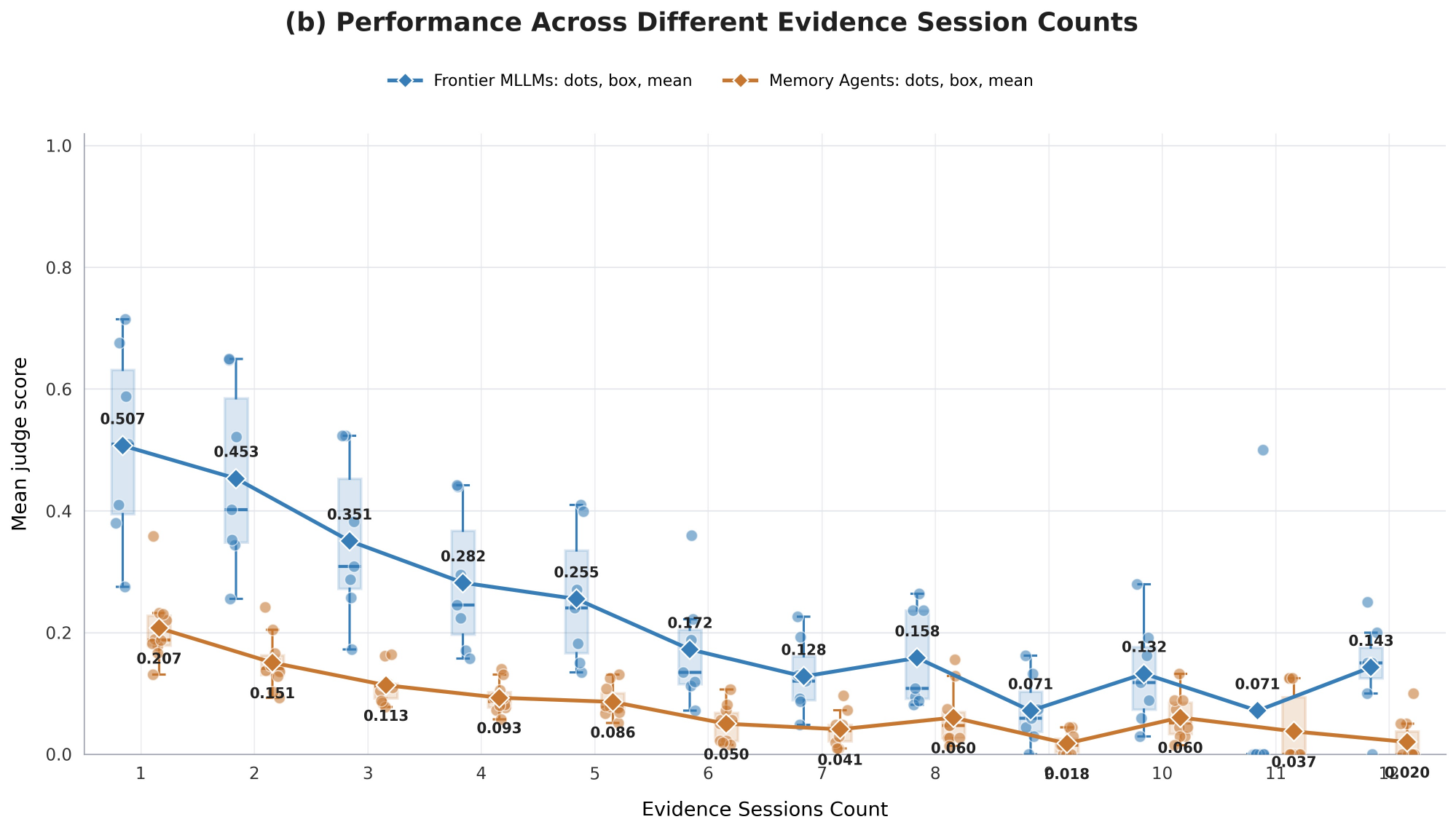}}
	 
	\end{minipage}
 
	\caption{Performance of MLLMs and memory agents with Qwen3-VL-Plus \cite{bai2025qwen3} as base model across QA tasks (a) and evidence session counts (b).}
	\label{fig_qwen3-vl-plus}
\end{figure*}

\textbf{Implementation.}
Frontier MLLMs do not employ external memory modules and instead generate responses directly from historical interactions within the context windows. 
To evaluate their long-range cross-modal referential reasoning capabilities, we provide the complete conversation history, organized chronologically by session, timestamp, and turn, as input. 
Models with visual input receive both the dialogue history and corresponding historical images, whereas text-only models receive the dialogue history supplemented with available image captions.

Given that the complete multimodal histories in MemRR exceed 1M tokens, with image tokens accounting for approximately 80\% of the total, we adopt a resolution-adaptive image scaling strategy to accommodate model context limits and improve request processability in the full-context setting. 
Specifically, for each model, historical images whose longest side meets or exceeds a predefined threshold are proportionally downscaled to that threshold, preserving their aspect ratios. The specific resolution thresholds are reported in Table~\ref{tab_6}, including the average input length for each model. For each QA, we append the question to the resulting full-context prompt and require the model to return a concise answer in JSON format.

\subsection{Metrics}
\label{appendix_metrics}
All metrics are computed at the instance level and averaged over the QA set. We adopt two metrics widely employed in conversational question answering: \textit{Token-level F1} and \textit{LLM-as-a-Judge}. The former quantifies lexical overlap between predicted and ground truth answers, providing a reproducible measure of factual-term matching and answer coverage. The latter assesses semantic correctness and task fulfillment based on explicit scoring criteria, capturing whether the answer is complete and assess its semantic consistency with the ground truth answer beyond surface-level similarity.

\textbf{Token-Level F1.} It calculates precision and recall based on the number of shared terms between the predicted answer $a$ and the reference answer $\hat{a}$ after predefined text normalization and tokenization, and uses the harmonic mean of the two as the final score. Let $T_p=token(a)$
and $T_G=token(\hat{a})$ denote the multisets of tokens in the predicted and ground truth answer, and $|T_p \cap T_G|$ the number of overlapping tokens. Then precision P, recall R and F1 are defined as:

\begin{equation}
\begin{aligned}
    P=\frac{\left| T_{p}\cap T_{G} \right|}{T_{p}}, R=\frac{\left| T_{p}\cap T_{G} \right|}{T_{G}},
    F1=\frac{2\cdot P\cdot R}{P+R}
\end{aligned}
\end{equation}

We assign (F1=1) when both $T_{p}$ and $T_{G}$ are empty and (F1=0) when exactly one is empty. This metric is particularly suitable for absolute-answer questions requiring concrete answers, such as amounts, dates, quantities, entity names, and item lists, where lexical overlap provides a meaningful signal of answer accuracy. Moreover, unlike binary exact-match evaluation, F1 captures partial content matches and thus offers a more fine-grained assessment.

\textbf{LLM-as-a-Judge.}
LLM-as-a-Judge is a model-based evaluation paradigm in which an LLM automated assesses answer quality against predefined criteria, rather than a metric with a universally fixed formulation. In answer-key-driven question-answering (QA) evaluation, the judge model receives the question $\hat{q}_i$, the predicted answer $a_i$, and the ground truth answer $\hat{a}_i$, and evaluates the prediction according to established QA evaluation protocols.
We employ GPT-5.2 \cite{singh2025openai} and Qwen3.5-122B-A10B \cite{qwen352026} as independent judges and average their scores to obtain the final LLM-Judge result.
Each judge assigns a score from five-level scale ({0, 0.25, 0.5, 0.75, 1.0}), corresponding to five quality levels: Incorrect, Poor/Deviating, Partial/Vague, Good/Minor Defects, and Correct/Accurate. The specific prompt is illustrated in Fig \ref{fig_prompt4}. The score for the (i)-th QA instance is formulated as:

\begin{equation}
      s_i=J_{\theta}(\hat{q}_i,a_i,\hat a_i;\mathcal R)
\end{equation}

where ($J_{\theta}$) denotes the judge model and ($\mathcal R$) represents the evaluation rubric. The final LLM-Judge score is obtained by averaging the instance-level scores across both judges and the QA set.

Although rubric-based LLM-as-a-Judge evaluation enables fine-grained assessment of answer correctness, particularly in borderline cases and variations in answer formulation, its reliability may depend on the choice of judge model. To examine the reliability of our evaluation protocol, we conduct a small-scale human validation study.

Specifically, we randomly sample 50 QA instances from each category and evaluate the model-generated answers against their corresponding questions and ground-truth answers. The evaluations are independently performed by two LLM judges (GPT-5.2 \cite{singh2025openai} and Qwen3.5-122B-A10B \cite{qwen352026}) and human annotators. As reported in Table~\ref{tab:human_validation}, we assess agreement from two perspectives. $r_{con}$ in each column reports exact score agreement with human annotations, which measures the proportion of identical graded scores. $P/N$ in each column reports agreement on binary answer correctness. Specifically, we classify scores in the range of $[0.75, 1.0]$ as positive answers and those below $0.5$ as negative answers, thereby converting graded evaluations into binary judgments and enabling a direct comparison of positive and negative answer identification.

The results demonstrate strong agreement between both LLM judges and human annotations across all task categories, in terms of both exact score matching and binary correctness classification. Their average scores also provide consistent estimates of answer quality, indicating that the evaluation outcomes are relatively robust to judge selection. In particular, the agreement between judges from different model families mitigates concerns that the LLM-as-a-Judge results are driven by model-specific evaluation preferences. These findings support the reliability of our rubric-based evaluation protocol and justify the use of LLM-as-a-Judge as the primary evaluation metric for MemRR.

\begin{figure*}
    \centering
    \includegraphics[width=\linewidth]{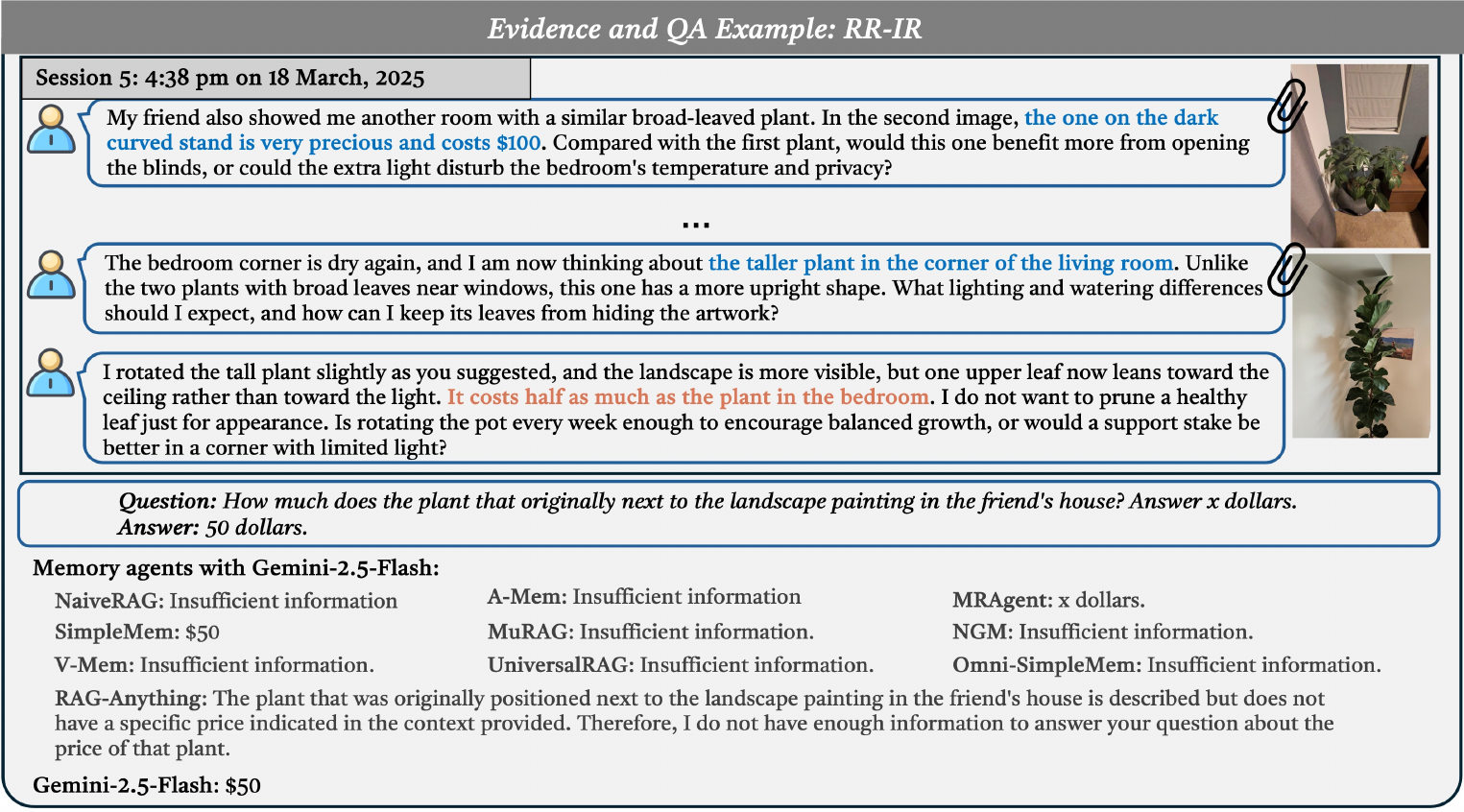}
    \caption{Example of Instance Resolution (RR-IR), including multimodal evidences, QA pair, results of memory agents and baselines. }
    \label{fig_example1}
\end{figure*}

\begin{figure*}
    \centering
    \includegraphics[width=\linewidth]{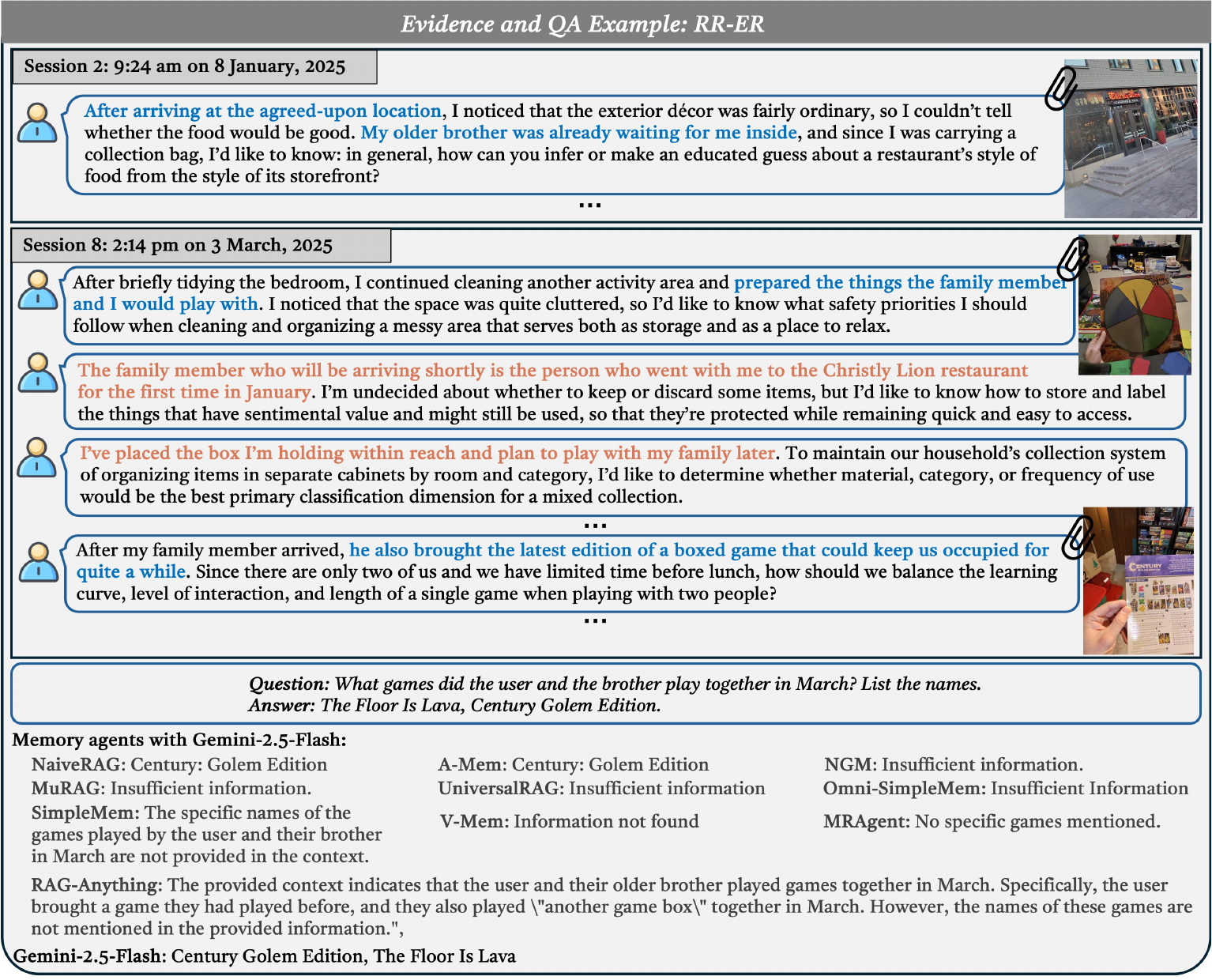}
    \caption{Example of Event Resolution (RR-ER), including multimodal evidences, QA pair, results of memory agents and baselines. }
    \label{fig_example2}
\end{figure*}

\begin{figure*}
    \centering
    \includegraphics[width=\linewidth]{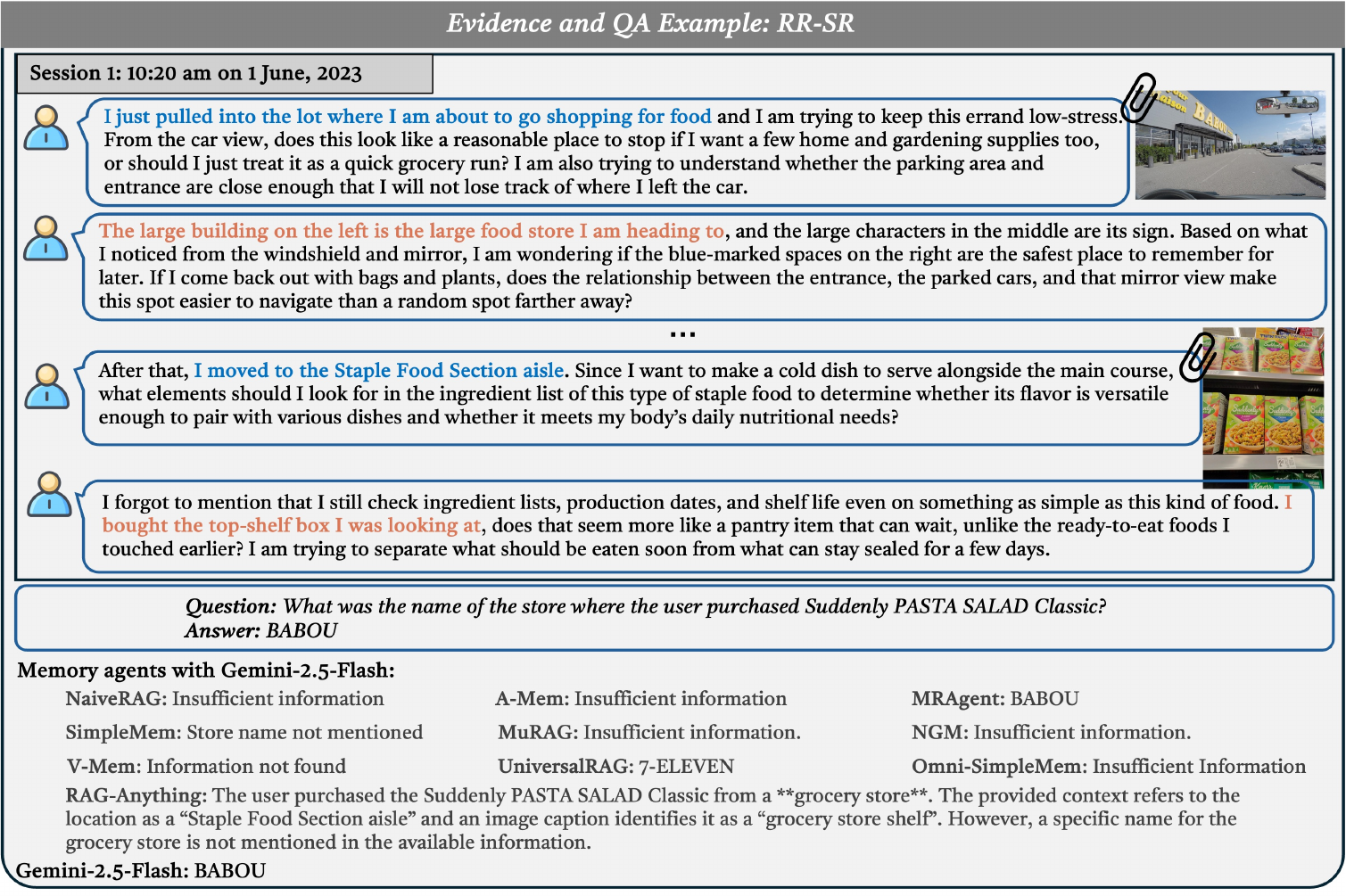}
    \caption{Example of Situational Resolution (RR-SR), including multimodal evidences, QA pair, results of memory agents and baselines. }
    \label{fig_example3}
\end{figure*}

\subsection{Baseline Details}

To ensure a fair comparison between text-based and multimodal methods, text-based methods are provided with high-quality image captions generated by GPT-5.6-sol \cite{singh2025openai} as textual surrogates for visual information, whereas multimodal methods receive the original images accompanying each dialogue turn. The evaluated methods are described as follows.

\textbf{NaiveRAG} adopts a conventional dense-retrieval pipeline. It stores the original textual content of each dialogue turn and its corresponding semantic embeddings obtained through a dense sentence encoder. During inference, it retrieves the top-\(k\) entries based on query--memory similarity and directly concatenates them into the model prompt without additional reranking or summarization.

\textbf{A-Mem} \cite{xu2025mem} introduces an agent-based memory system that organizes experiences into a dynamically evolving network of structured memories. Each newly added memory is transformed into a structured note containing contextual descriptions, keywords, and tags, which are used to establish associations with existing memories and trigger updates to their representations. This agent-driven mechanism enables memory organization and evolution without predefined memory operations.

\textbf{SimpleMem} \cite{liu2026simplemem} employs semantically lossless compression to maximize memory information density and token efficiency. It transforms unstructured interactions into compact memory units with multi-view semantic indexing through structured compression, consolidates related contexts into unified abstract representations through online semantic synthesis to reduce redundancy. Finally, intent-aware retrieval planning dynamically determines the retrieval scope and constructs precise contexts.

\begin{figure*}
    \centering
    \includegraphics[width=\linewidth]{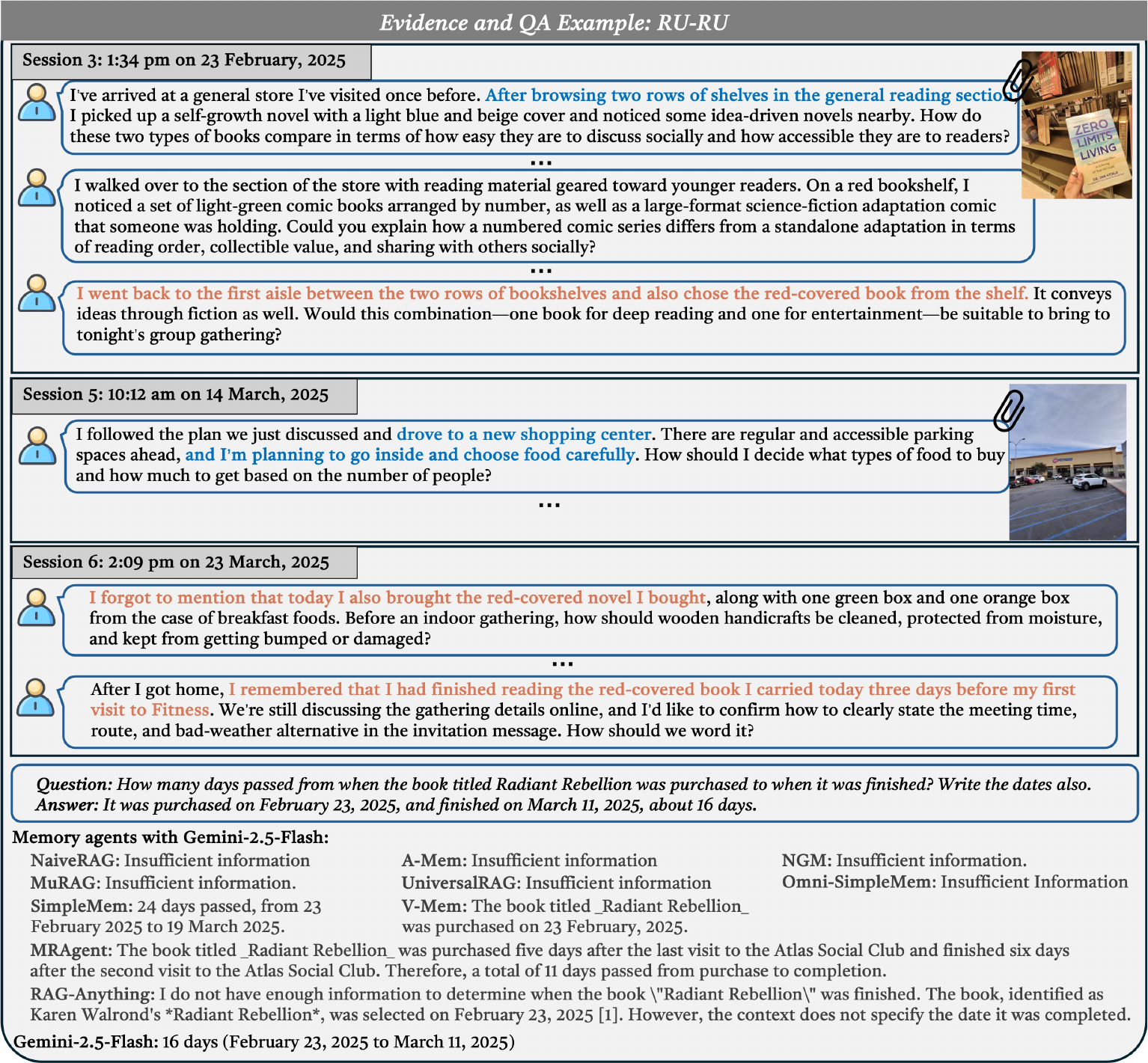}
    \caption{Example of Relational Understanding (RU-RU), including multimodal evidences, QA pair, results of memory agents and baselines. }
    \label{fig_example4}
\end{figure*}

\textbf{MRAgent} \cite{ji2026memory} integrates an associative memory graph with active memory reconstruction. It represents memory as a cue–tag–content graph, where associative tags serve as semantic bridges connecting fine-grained cues to memory content. It adopts LLM-based reasoning into memory access to adapt retrieval to evolving reasoning contexts, enabling the agent to iteratively explore and prune retrieval paths based on accumulated evidence.

\textbf{MuRAG} \cite{chen2022murag} encodes the multimodal content of each dialogue turn into dense semantic vectors using a joint vision-language encoder, which maps queries and memory entries into a shared embedding space. During inference, it retrieves the top-\(k\) most similar memory entries based on vector similarity and concatenates them into the model prompt without additional reranking or summarization.

\textbf{UniversalRAG} \cite{yeo2026universalrag} adopts a modality- and granularity-aware retrieval framework that separately encodes and indexes textual and visual content from multimodal dialogue histories. A lightweight router selects the modality and granularity most relevant to the given query, followed by semantic similarity-based retrieval within the selected index. The retrieved top-\(k\) entries are then incorporated into the model prompt.

\textbf{NGM} \cite{fisher2025neural} models long-term memory as a heterogeneous graph with dialogue facts represented as nodes and cross-modal semantic associations and temporal metadata used to establish edges. Memory retrieval is performed through graph traversal, starting from query-anchored nodes and expanding to neighboring nodes. The resulting subgraph is returned as contextual evidence, enabling retrieval beyond flat vector similarity matching.

\textbf{RAG-Anything} \cite{guo2025rag} is an end-to-end retrieval framework that indexes heterogeneous content, including text, images, tables, and mathematical expressions, within a unified embedding space. It employs dual-graph construction to jointly capture cross-modal associations and textual semantics, and supports cross-modal top-\(k\) retrieval through a unified query encoder.

\begin{figure*}
    \centering
    \includegraphics[width=\linewidth]{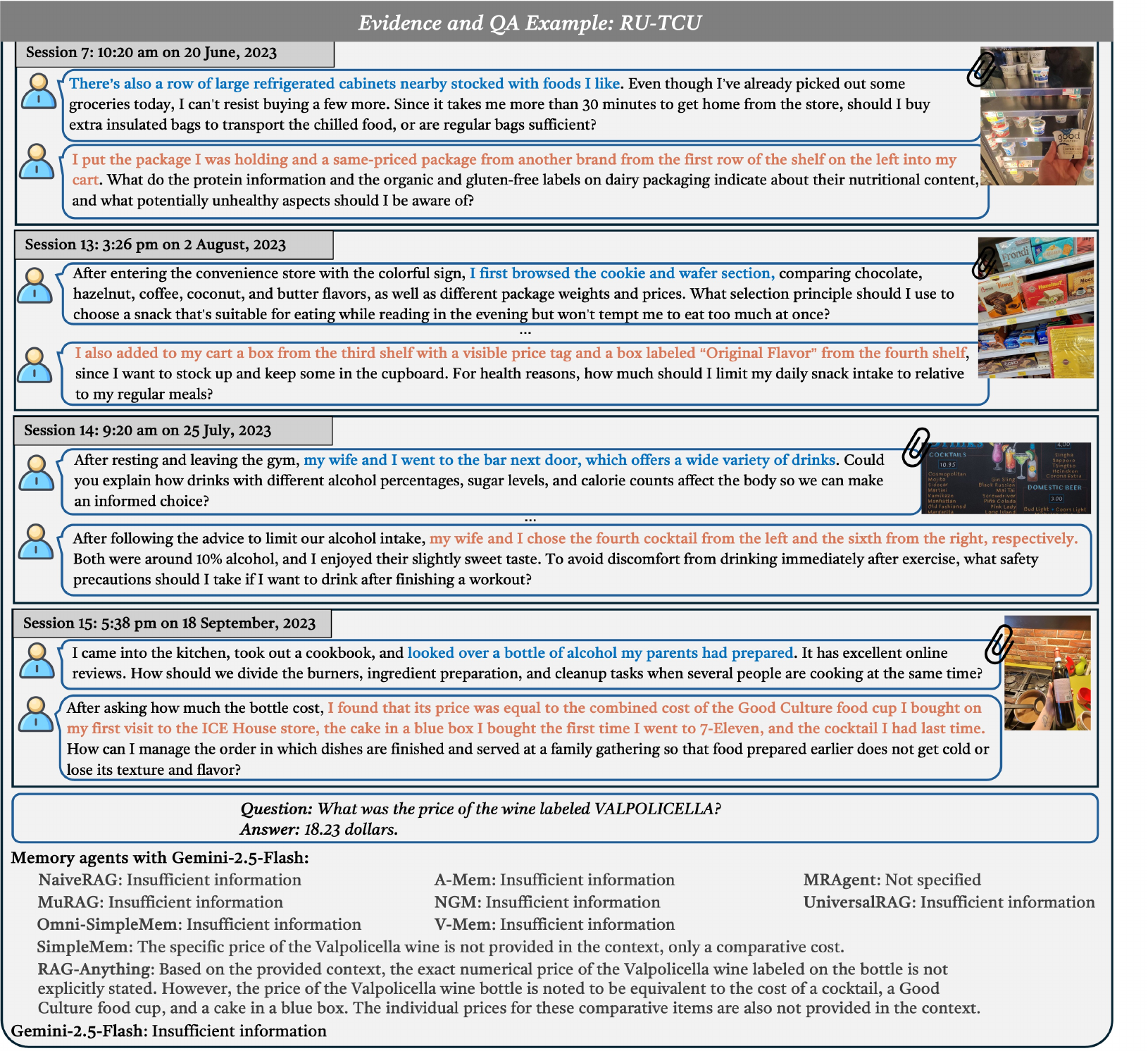}
    \caption{Example of Temporal Consistency Understanding (RU-TCU), including multimodal evidences, QA pair, results of memory agents and baselines.}
    \label{fig_example5}
\end{figure*}

\textbf{Omni-SimpleMem} \cite{jia2025omnispatial} incorporates novelty-aware memory selection and hierarchical multimodal retrieval. A lightweight perception encoder assesses incoming information novelty to selectively determine what to retain. It adopts multimodal atomic units as a unified memory representation to support retrieval while retaining access to the original content. Retrieval follows a three-stage strategy spanning summaries, details, and raw evidence, and combines dense vector retrieval with sparse keyword matching through set-union operations.

\textbf{V-Mem} \cite{kang2026v} is an agentic multimodal memory system that routes retrieval according to the modality of the query and the target evidence, both inferred exclusively from the query. It segments conversations into individual turns and uses LLM-generated anchors to locate relevant content. Once a match is identified, the system retrieves the target-modality content from the same dialogue turn.

\begin{figure*}
    \centering
    \includegraphics[width=\linewidth]{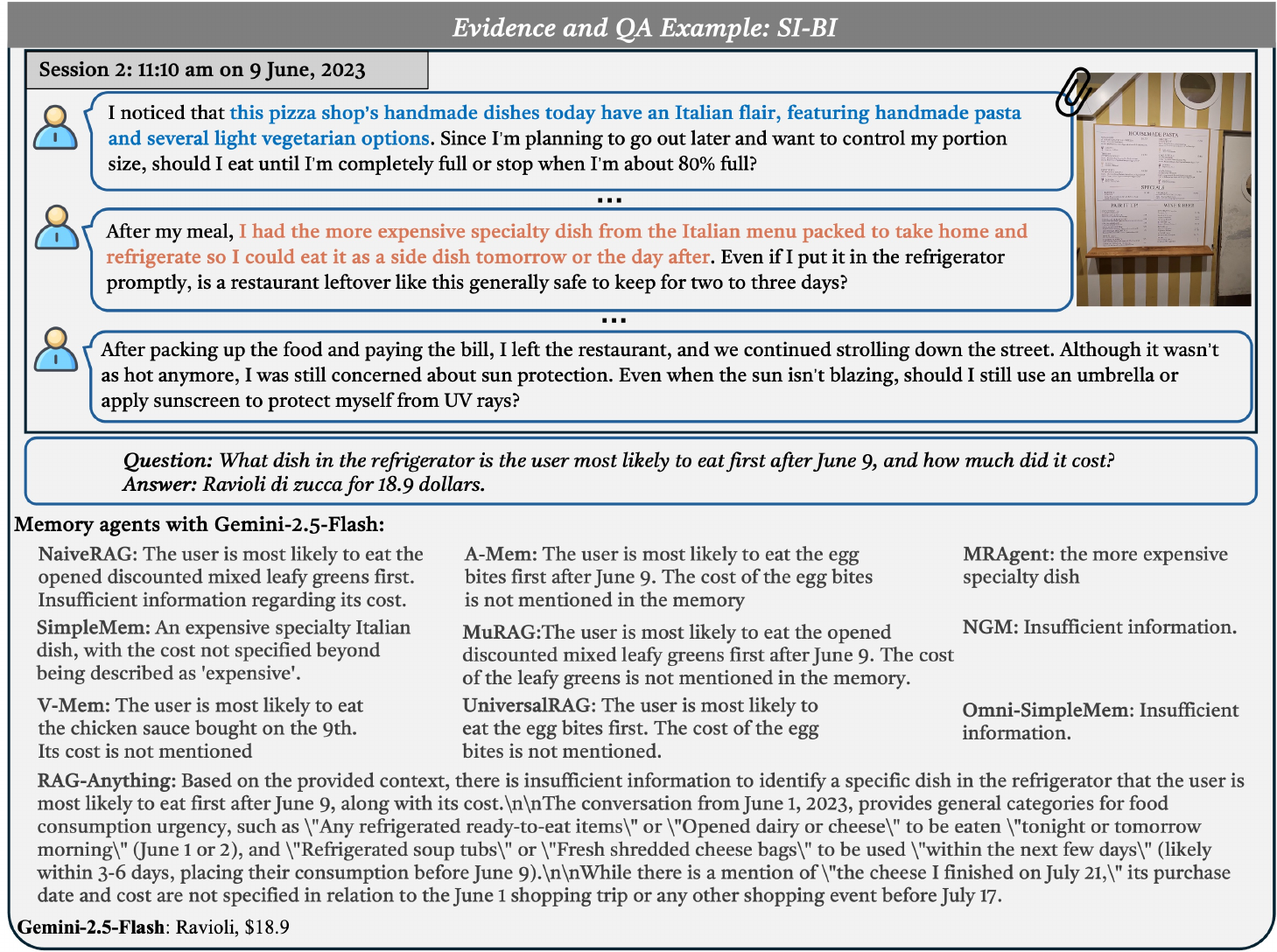}
    \caption{Example of Behavior Inference (SI-BI), including multimodal evidences, QA pair, results of memory agents and baselines.}
    \label{fig_example6}
\end{figure*}

\subsection{Baseline Setup}

For all memory-based agent methods, we further expanded upon the Mem-gallery \cite{bei2026mem} evaluation framework, retaining the evaluation workflows for NaiveRAG, MuRAG \cite{chen2022murag}, NGM \cite{fisher2025neural}, and UniversalRAG \cite{yeo2026universalrag} while adding other methods. We obtained the official code for each method and directly utilized each method’s memory storage and construction processes without performing task-specific tuning. Given the larger number of dialogue turns and evidence chains in MemRR, we appropriately increased the retrieval hyperparameters for existing methods. Table \ref{tab_parameter} reports key hyperparameter settings for each method. For text and image embedding backbones, the Text-embedding-3-small backbone was accessed via API calls to the OpenAI interface, while the remaining embedding backbones were deployed on a Linux system using the vLLM engine.

\begin{figure*}
    \centering
    \includegraphics[width=\linewidth]{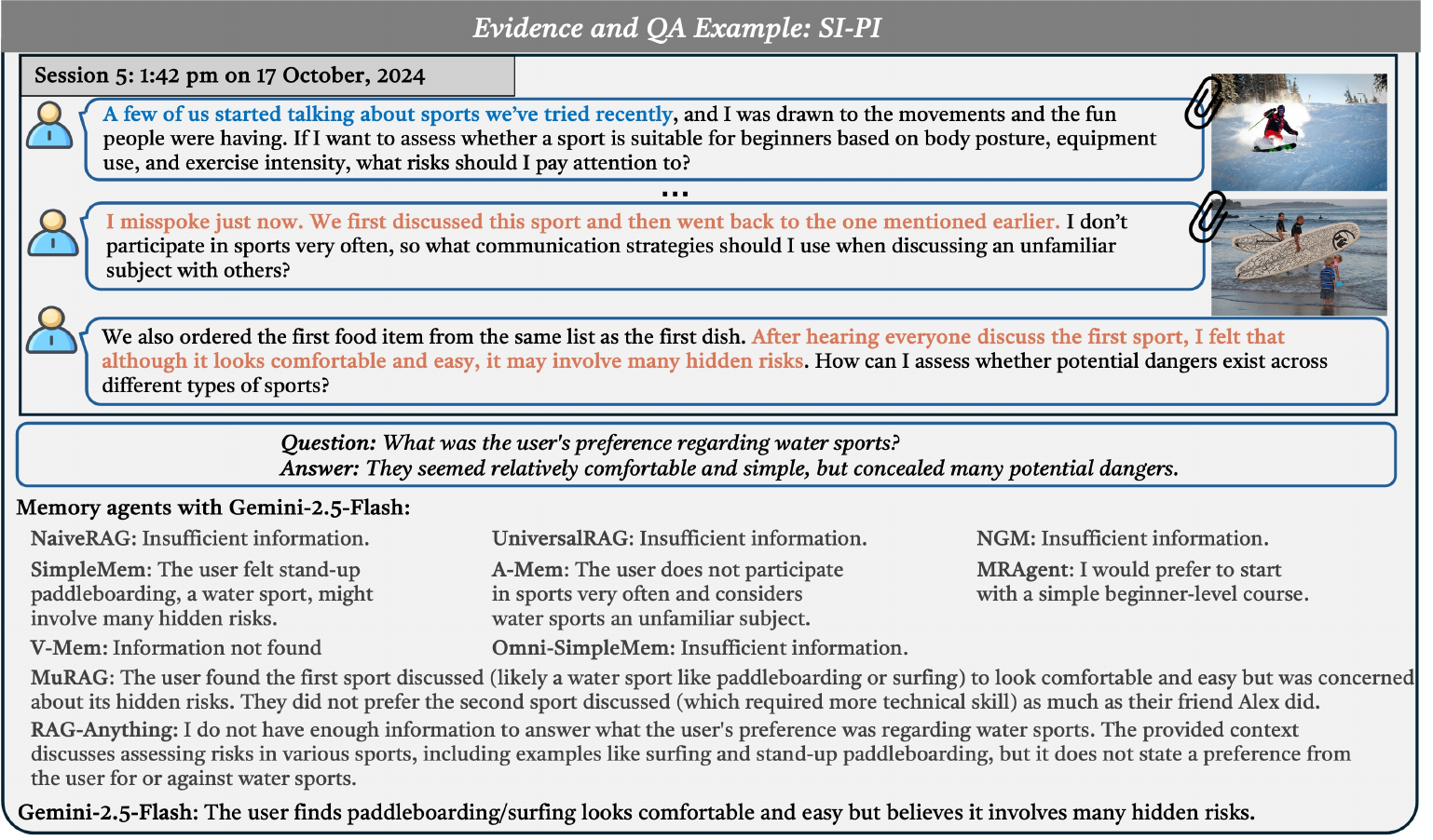}
    \caption{Example of Preference Inference (SI-PI), including multimodal evidences, QA pair, results of memory agents and baselines.}
    \label{fig_example7}
\end{figure*}
 
\section{Additional Results and Analysis}
\label{appendix_C}

\subsection{Alternative MLLM Backbone}

\textbf{Overall performance}.
Across different base models: GPT-4o-mini (Table \ref{tab_gpt-4o-mini}), Qwen3-VL-Plus (Table \ref{tab_qwen3-vl-plus}) 
, the best F1 and LLM Judge scores are 0.1307 and 0.2195, 0.1436 and 0.1951, 0.0986 and 0.2194. Existing memory methods consistently exhibit limited performance on MemRR, with no method achieving uniformly strong results across all evaluation dimensions. Although individual methods exhibit varying strengths under different backbones, neither text-based nor multimodal memory approaches demonstrate consistent superiority. In particular, multimodal memory alone does not guarantee improved performance, suggesting that existing methods remain insufficient for modeling persistent cross-modal entity associations and evolving referential dependencies.

\textbf{Performance Gap Across QA Tasks.}
In Fig. \ref{fig_gpt-4o-mini}(a) and Fig. \ref{fig_qwen3-vl-plus}(a),
we present the performance of memory methods and MLLMs across different QA types under different base models. Across different base models, these methods exhibit consistent trends in QA types and face bottlenecks in both reference resolution and reference comprehension tasks. indicating fundamental difficulties in maintaining fine-grained visual-linguistic correspondences and tracking referential semantics across sessions. Although memory agents achieve relatively better performance on semantic inference, particularly SI-PI, but there is still significant room for improvement. These findings suggest that the primary bottleneck extends beyond information retrieval to the construction and utilization of coherent, long-term cross-modal referential knowledge.

\textbf{Impact of Evidence Session Count.}
In Fig. \ref{fig_gpt-4o-mini}(b) and Fig. \ref{fig_qwen3-vl-plus}(b), we present the performance of memory methods and MLLMs across evidence session count under different base models. It further reveals a consistent performance degradation as the number of evidence sessions increases. Memory agents demonstrate substantially lower scores and more pronounced degradation, with performance approaching zero under increasingly distributed evidence requirements. This trend persists across different base models, indicating that the limitation is not merely attributable to backbone capacity but reflects fundamental deficiencies in existing memory mechanisms for aggregating complementary evidence and preserving long-range cross-session dependencies.

\begin{figure}
    \centering
    \includegraphics[width=\linewidth]{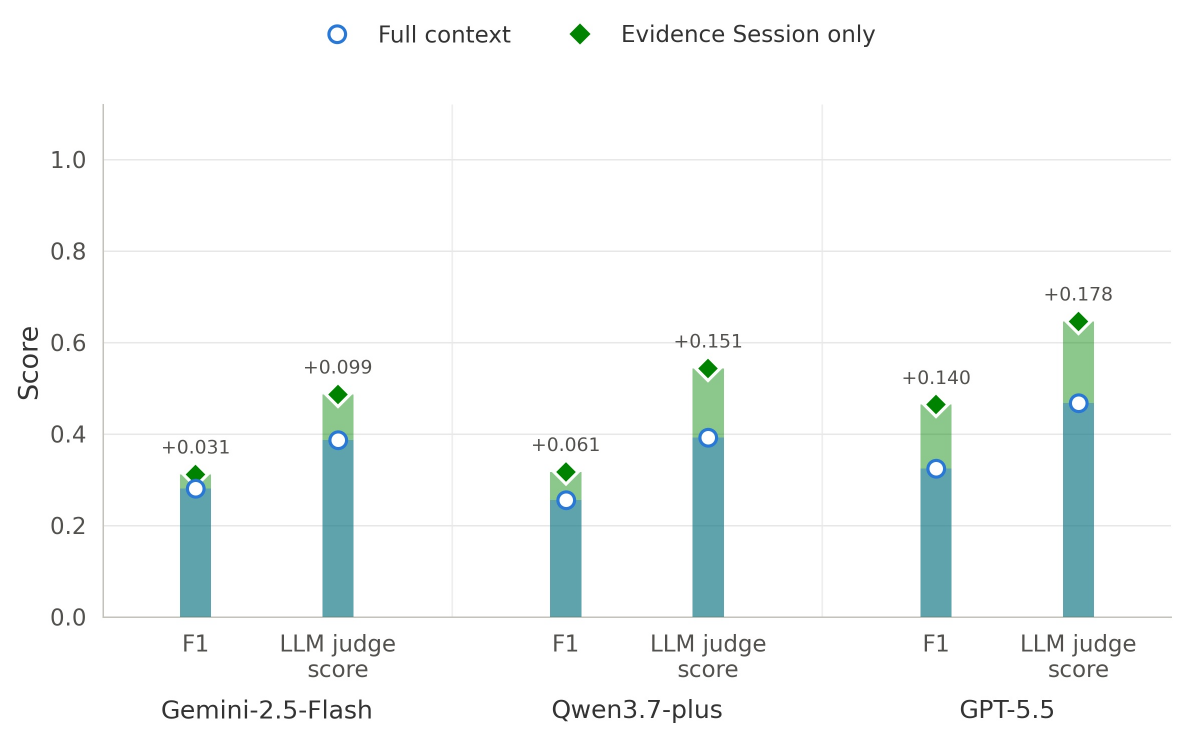}
    \caption{Performance of three MLLMs under full-context setting and evidence session setting}
    \label{fig_score_comparison}
\end{figure}

\begin{figure*}
    \centering
    \includegraphics[width=\linewidth]{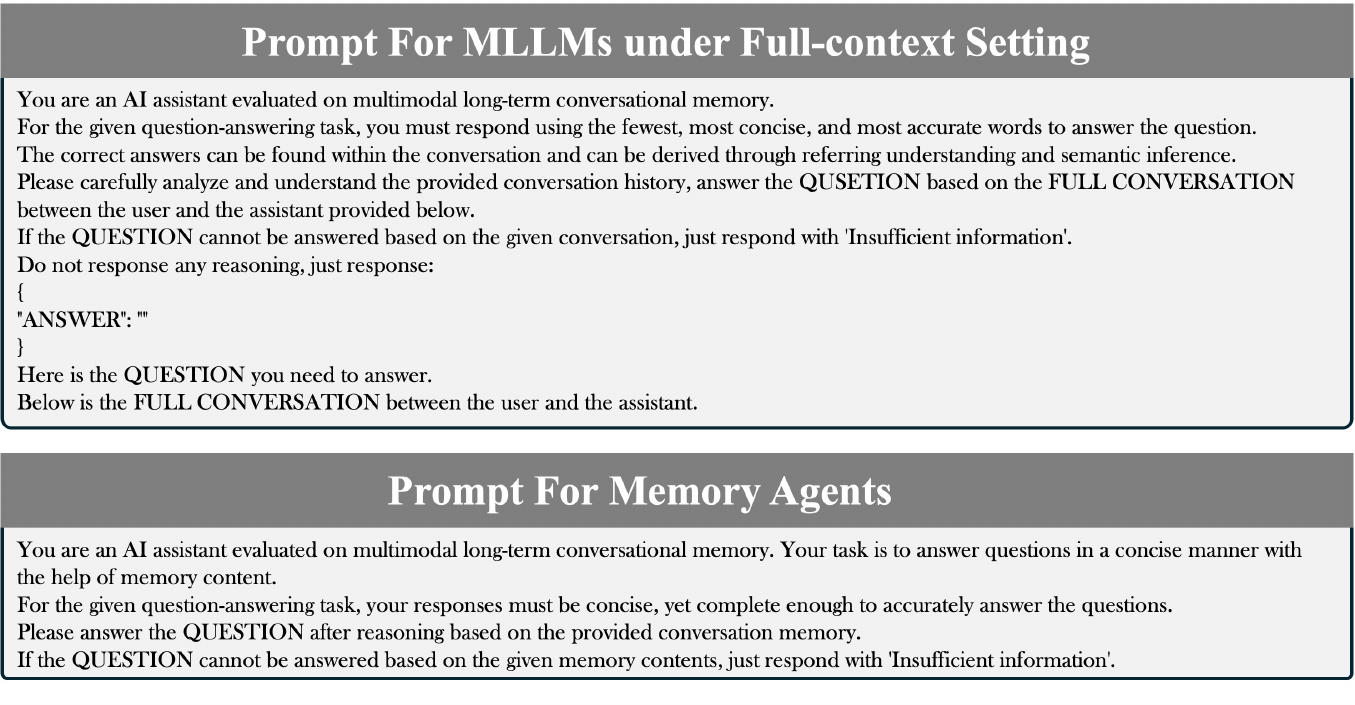}
    \caption{Prompt template for MLLMs under full-context setting and memory agents.}
    \label{fig_prompt5}
\end{figure*}

\subsection{Different Setting on MLLMs}

Fig. \ref{fig_score_comparison} reports the performance of three MLLMs under full-context setting and evidence session setting, respectively.
Across three base models, restricting the multimodal context to evidence sessions consistently improves QA performance. Specifically, F1 increases by 3.1, 6.1, and 14.0 points for Gemini-2.5-Flash, Qwen3.7-Plus, and GPT-5.5, respectively, while LLM-Judge scores exhibit even larger gains of 9.9, 15.1, and 17.8 points. The consistent improvements across different MLLMs show the extra context chiefly eroded the semantic quality of its answers. Because even frontier MLLMs lose accuracy once retrieval and context pressure are added back, the full-history regime is not a viable substitute for structured memory. The results demonstrate the importance of identifying evidence session in the dialogue for long-range referential inference tasks.

\begin{table*}[t]
    \centering
    \caption{Themes and session-level event trajectories of the benchmark scenarios.}
    \label{tab_theme_event}

    \setlength{\tabcolsep}{3pt}
    \renewcommand{\arraystretch}{1.15}
    \scriptsize

    \begin{tabular}{
        @{}
        >{\raggedright\arraybackslash}p{0.11\textwidth}
        >{\raggedright\arraybackslash}p{0.18\textwidth}
        >{\raggedright\arraybackslash}p{0.7\textwidth}
        @{}
    }
    \toprule

    \textbf{Scenario} &
    \textbf{Theme} &
    \textbf{Session Events} \\

    \midrule

    Gardening enthusiast life &
    A gardening enthusiast's two-month daily life, involving planting, gardening supplies, flower-bed maintenance, bird feeding, and interactions with fellow hobbyists. &
    S1: Gardening starts$\rightarrow$S2: Plants flowers and fixes pots$\rightarrow$S3: Shares plants with a friend$\rightarrow$ S4: Buys and organizes supplies$\rightarrow$S5: Protects plants from pets$\rightarrow$S6: Adjusts bird feeders$\rightarrow$ S7: Installs security cameras$\rightarrow$S8: Buys tools and houseplants$\rightarrow$S9: Handles gardening injuries$\rightarrow$S10: Replants plants and decorates bird feeders$\rightarrow$S11: Buys books$\rightarrow$S12: Celebrates birthday$\rightarrow$S13: Buys and rearranges succulents$\rightarrow$S14: Upgrades bird feeders$\rightarrow$S15: Visits grandparents$\rightarrow$S16: Organizes the patio and plants$\rightarrow$S17: Buys gardening equipment$\rightarrow$S18: Rearranges plants and prepares for workshop$\rightarrow$S19: Attends the workshop$\rightarrow$S20: Repots plants and buys novels. \\

    \midrule

    Cloth shopping activities &
    A six-month clothing-shopping journey involving wardrobe organization, family shopping, travel preparation, social events, and fitness activities. &
    S1: Organizes wardrobe and plans outfits$\rightarrow$S2: Buys winter coats$\rightarrow$S3: Shops for family and handles returns $\rightarrow$S4: Organizes shoes and donations$\rightarrow$ S5: Buys tennis gear and first practice$\rightarrow$S6: Starts weekly gym training$\rightarrow$S7: prepares meals for family visit$\rightarrow$S8: Organizes supplements and plans system$\rightarrow$S9: Buys workwear for family$\rightarrow$ S10: Cleans the home and buys decorations$\rightarrow$S11: Buys clothes for children$\rightarrow$S12: Attends fundraiser and donates $\rightarrow$ S13: Buys shoes and cosmetics fo events$\rightarrow$S14: Plans party supplies$\rightarrow$S15: Hosts friends and hotpot$\rightarrow$S16: Prepares and packs for France$\rightarrow$S17: Arranges pet care and buys travel items$\rightarrow$S18: Book arrangement$\rightarrow$S19: Prepares supplies for trip$\rightarrow$S20: Manages travel delays and souvenirs.\\

    \midrule

    Collector daily life & An eight-month exploration of collecting, including museum and art-market visits, home-display maintenance, collection organization, and social interactions. &

    S1: Shops for household supplies$\rightarrow$S2: Rearranges collectibles $\rightarrow$S3: Museum and buys art pieces$\rightarrow$S4: Reviews budgets and browses books$\rightarrow$S5: Hosts office lunch and practices music$\rightarrow$S6: Buys pottery in market $\rightarrow$ S7: Shops for food and plans camping$\rightarrow$S8: Home clean and leisure time$\rightarrow$ S9: Takes a short trip and buys books$\rightarrow$S10: Revisits an exhibition and game$\rightarrow$S11: Cleans the office and maintains collections$\rightarrow$S12: Restocks and organizes collection records$\rightarrow$S13: Buys books and plans gifts$\rightarrow$ S14: Buys holiday gifts and friend visiting$\rightarrow$S15: Helps organize furniture$\rightarrow$ S16: Restocks groceries and organizes books$\rightarrow$S17: Buys items and visits a colleague$\rightarrow$ S18: Revisits an exhibition$\rightarrow$ S19: Buys collectibles and updates records$\rightarrow$ S20: Revisits a exhibition and resale books.
 \\

    \midrule

    Weekend activities &
    A five-month account of weekend life involving family meals, household chores, travel, hobbies, community events, and social gatherings. &
    S1: Cooks and cleans the kitchen$\rightarrow$S2: Buys clothes for family$\rightarrow$S3: Watches sports and plans gathering$\rightarrow$S4: Prepares clothing and reading exercises$\rightarrow$S5: Maintains flower beds and organizes clothes$\rightarrow$S6: Travels to France$\rightarrow$S7: Shops for luggage$\rightarrow$S8: Home cleanse and indoor plants$\rightarrow$S9: Declutters the bedroom and plants$\rightarrow$S10: Per company and reads novels$\rightarrow$ S11: Reviews math and eating$\rightarrow$ S12: Donates clothes and updates wardrobe $\rightarrow$ S13: Cooks and rearranges plants $\rightarrow$ S14: Organizes team-building games$\rightarrow$S15: Family gathering and rearranges plants$\rightarrow$ S16: Reading discussion and plans $\rightarrow$S17: Repairs and evaluates garden plants$\rightarrow$S18: Pizza gathering and organizes leftover $\rightarrow$S19: Attends workshop and organizes images$\rightarrow$ S20: Workshop and organizes project files$\rightarrow$S21: Improves picnic invitation. \\

    \midrule

    Family holidays &
    A nine-month family travel experience involving museums, sculptures, beaches, zoos, outdoor activities, and shared family memories. &

    S1: Visits sculpture museum$\rightarrow$ S2: Explores exhibits and the old town$\rightarrow$S3: Takes night photos and buys souvenirs$\rightarrow$ S4: Enjoys a family beach walk$\rightarrow$S5: Visits a church and records videos$\rightarrow$ S6: Zoo exploring and edits videos$\rightarrow$S7: Watches birds and organizes photos $\rightarrow$S8: Feed animals and receive plush toys $\rightarrow$S9: Buys medicine and returns home $\rightarrow$S10: Revisits museums and creates artwork$\rightarrow$S11: New trip and buys books$\rightarrow$S12: Hikes and enjoys music $\rightarrow$S13: Explores antiques and storytelling$\rightarrow$S14: Visits cabin and shares travel photos$\rightarrow$S15: Visits museum and exchanges books$\rightarrow$ S16: Park picnic and performs a puppet story $\rightarrow$S17: Completes a puppet story $\rightarrow$S18: Edits recordings and orders souvenirs $\rightarrow$S19: Visits exhibition and creates voiceovers $\rightarrow$ S20: Parade and shares edited videos.
     \\

    \midrule

    Food lover life &
    A four-month culinary journey involving grocery shopping, recipe exploration, home cooking, pet care, outdoor meals, and family gatherings. &
    S1: Food Shopping for gethering$\rightarrow$S2: Street dining and dirnks $\rightarrow$ S3: Organize fridge and tries recipes $\rightarrow$S4: Pet supplies and homemade food $\rightarrow$S5: Shops and plans gathering $\rightarrow$ S6: Buys pastries and discounted bread $\rightarrow$S7: Shops and returns items to manage budget$\rightarrow$ S8: Backyard gathering and brings gifts home$\rightarrow$ S9: Organizes kitchen storage$\rightarrow$ S10: Home management and buys supplies$\rightarrow$S11: Travels for business $\rightarrow$S12: Shop for picnics and cooks $\rightarrow$ S13: Buys sci-fi books and groceries$\rightarrow$S14: Exercises and breaks a decoration$\rightarrow$S15: Famliy visit and leisure$\rightarrow$S16: Prepares food for gathering$\rightarrow$S17: Visits cabin and reads$\rightarrow$S18: Visits a client and exercises $\rightarrow$S19: Buys travel supplies and medicine$\rightarrow$S20: Bakes oat snacks$\rightarrow$S21: Selects pet and considers pet-child safety. \\

    \midrule

    Book lover life &
    A three-month reading-oriented lifestyle involving book shopping, library visits, maintaining pet fish, and exchanging recommendations with friends. &
    
    S1: Prepares groceries for meal$\rightarrow$S2: Library card and borrows novels$\rightarrow$S3: Starts fishkeeping and buys aquarium supplies$\rightarrow$ S4: Returns books and buys games$\rightarrow$S5: Studies professional materials and practices piano $\rightarrow$ S6: Buys and organizes books$\rightarrow$ S7: Board game night and cleans$\rightarrow$ S8: Book discussion$\rightarrow$S9: Returns books and buys new titles $\rightarrow$S10: Library and plans further reading $\rightarrow$S11: Buys literature and cookbooks$\rightarrow$S12: Replaces pet and buys sci-fi books $\rightarrow$S13: Reading event and discuss$\rightarrow$S14: Accounts organize and music exhibition$\rightarrow$S15: Book sharing and family discussions $\rightarrow$S16: Buys books before a reading event $\rightarrow$S17: Joins book discussions and exhibition $\rightarrow$ S18: Exchanges books with colleagues$\rightarrow$S19: Visits friends and sorts books $\rightarrow$S20: Book discussion and fix pet. \\

    \midrule

    Home renovation activities &
    A four-month home-renovation and decorating project involving furniture placement, household purchases, organization, and interactions with coworkers. &
    
    S1: Groceries for home and guests$\rightarrow$ S2: Organizes budgets and garage storage $\rightarrow$S3: Visits gallery and orders sofa $\rightarrow$S4: Visits Jay and selects storage furniture$\rightarrow$ S5: Organizes cabin garden and furniture purchases$\rightarrow$ S6: Organizes the study and exercises$\rightarrow$S7: Manages finance and dining$\rightarrow$ S8: Maintains household items $\rightarrow$S9: Buys medicine and bathroom decorations$\rightarrow$S10: Revisits venue and orders furniture$\rightarrow$S11: Organizes bedrooms and reads with children$\rightarrow$ S12: Flea market shopping and selects furniture$\rightarrow$S13: Cabin garden and measures windows $\rightarrow$ S14: Colleagues and rearranges the living room$\rightarrow$S15: Revisits exhibition and bedroom storage$\rightarrow$ S16: Organizes spices and orders guest-room furniture$\rightarrow$S17: Visits Jane and organizes collectibles $\rightarrow$ S18: Buys supplies and plays board games$\rightarrow$S19: Visits Luke and orders storage furniture $\rightarrow$S20: Compares storage units and discusses work $\rightarrow$S21: Visits furniture exhibition. \\

    \midrule

    Social and entertainment activities &
    A seven-month social and recreational journey involving friends, travel, gift exchanges, shared hobbies, and dining experiences. &
    
    S1: Prepare food for gathering$\rightarrow$S2: Explores city and orders housewarming gift$\rightarrow$S3: Buys supplies for a reading club$\rightarrow$S4: Attends a sports gathering$\rightarrow$S5: Hosts gathering and organizes leftovers$\rightarrow$S6: Visits Joey and selects venue$\rightarrow$S7: Exercises and cares for plants $\rightarrow$S8: Visits Ben and plays board games $\rightarrow$S9: Group trip and cabin $\rightarrow$S10: Buys books and outdoor gear for park outing$\rightarrow$S11: Discusses hobbies and practices music$\rightarrow$S12: Eats pizza and plans disc golf$\rightarrow$S13: Gift shopping and dinner with Mike$\rightarrow$S14: Water sports and outdoor art$\rightarrow$ S15: Sports center and troubleshoots the aquarium$\rightarrow$S16: Visits Jack and flower arranging $\rightarrow$S17: Discusses sports and plans hobbies $\rightarrow$S18: Buys gardening supplies and prepares meetup$\rightarrow$S19: Budgeting and plays games$\rightarrow$ S20: Joins a university gathering. \\

    \midrule

    Life recording enthusiast life &
    A nine-month lifestyle-blogging journey documenting everyday experiences, including shopping, dining, browsing, socializing, and diary creation. &
    S1: Storage supplies and starts a journal$\rightarrow$S2: Explores the city and dining$\rightarrow$ S3: Book discussion and organizes notes$\rightarrow$S4: Travels and writes journals$\rightarrow$S5: Sports socialize and practices music $\rightarrow$S6: Book shopping and event planning $\rightarrow$S7: Hosts friends and hobby sharing$\rightarrow$ S8: Visits exhibition and records observations$\rightarrow$S9: Revisits park and completes a travel journal$\rightarrow$S10: Organizes gathering and game$\rightarrow$ S11: Travels for business$\rightarrow$ S12: Visits an exhibition and studies music $\rightarrow$S13: Buys medicine and visits friends$\rightarrow$S14: Adjusts travel plans $\rightarrow$S15: Shops and plans health tracking$\rightarrow$S16: Visits friends handles incident $\rightarrow$ S17: Relaxes with games and books$\rightarrow$S18: Buys supplies and visits friends $\rightarrow$S19: Street tour and maintains plants $\rightarrow$S20: Nature trip and adjusts leisure plans. \\

    \bottomrule
    \end{tabular}
\end{table*}

\section{Case Study}
\label{appendix_D}

We present three cases in entity-level identity maintenance, event-level participant–activity association, and scene-level contextual grounding. Instance Resolution (RR-IR) in Fig. \ref{fig_example1} evaluates the ability to preserve visual entity identities or attributes across turns, as exemplified by identifying a plant and deriving its price from the bedroom plant's cost, which is implicit attributes reasoned through relational references. Event Resolution (RR-ER) in Fig. \ref{fig_example2} evaluates the ability to reconstruct details of grounded events, requiring the identification of the brother through a previous shared experience and the retrieval of games in the event. Situational Resolution (RR-SR) in Fig. \ref{fig_example3} further examines the ability integrate visual scene cues with subsequent activities to recover contextual information, as in identifying the store associated with a later product purchase from its storefront and shopping context.

We also presented the benchmarking results. For RR-IR, SimpleMem \cite{liu2026simplemem} successfully infers through rewrite policy in a window. For RR-ER, NaiveRAG and A-Mem \cite{xu2025mem} retrieve one part, suggesting that similarity-based retrieval can recover locally salient facts without recovering complete details. For RR-SR, only MRAgent \cite{ji2026memory} identifies the correct store through memory reconstruction strategy, while UniversalRAG \cite{yeo2026universalrag} produces an incorrect prediction, indicating the difficulty of retaining and associating visual scene information with subsequent interactions. Notably, the full-context Gemini-2.5-Flash \cite{gemini22025} baseline correctly answers all cases. The performance gap reflects limitations in memory construction, information retention, and cross-turn evidence integration. Overall, effective reference resolution requires structured associations among visual entities, event details, and situational contexts, which remain inadequately supported by existing methods.

We further illustrate cases in relational reasoning and long-range consistency maintenance. RU-RU in Fig. \ref{fig_example4} requires associating purchase and completion events involving the same entity across sessions and integrating distributed temporal evidence with time expressions to reason time relationships. RU-TCU in Fig. \ref{fig_example5} examines the ability to preserve the identities and attributes of multiple historical entities over periods and aggregate them through subsequent relational constraints, as exemplified by aggregating the prices of previously purchased items. Both tasks go beyond isolated fact retrieval, requiring persistent associations and reasoning over temporally distributed evidence.

The results reveal limitations in existing methods. On RU-RU, SimpleMem \cite{liu2026simplemem} and MRAgent \cite{ji2026memory} fail to resolve the implicit temporal constraints, while V-Mem \cite{kang2026v} retrieves the purchase date without linking it to the corresponding completion event. Base model successfully integrates distributed evidence to infer the correct interval. On RU-TCU, however, none methods recover the target price. Although SimpleMem \cite{liu2026simplemem} and RAG-Anything \cite{guo2025rag} identify the compositional price relation, they fail to retrieve all the required historical attributes, exposing limitations in persistent entity--attribute association and multi-entity integration. Overall, existing methods remain largely constrained by local fact retrieval and aggregation, highlighting the need for explicit entity tracking, persistent attribute association, and relation-aware reasoning.

We present cases in behavioral and preference inference. As shown in Fig. \ref{fig_example6} , SI-BI requires connecting dining decisions, subsequent actions, and food-related attributes to identify an implicitly referenced consumption choice. SI-PI in Fig. \ref{fig_example7} further examines preference inference by requiring models to integrate multiple referential context and infer the user's attitude toward water sports, rather than merely retrieving explicitly stated interests. Together, these cases assess whether agents can inference behavioral patterns and preferences from referring context.

The base model consistently recovers the target behavior and preference, while most memory agents either fail to retrieve the relevant evidence or produce plausible but unsupported inferences. For behavioral inference, MRAgent \cite{ji2026memory} and SimpleMem \cite{liu2026simplemem} capture coarse-grained behavioral information but fail to recover the specific dish, while other methods exhibit irrelevant entities. For preference inference, SimpleMem \cite{liu2026simplemem} and MuRAG \cite{chen2022murag} partially capture the risk perception but fail to fully reconstruct, whereas A-Mem \cite{xu2025mem} and MRAgent \cite{ji2026memory} conflate contextual familiarity or generic recommendations with personal preference. These results reveal the difficulty of preserving the associations among behavioral events, entity attributes, and subjective evaluations.

\section{Prompt Templates}

\textbf{Evaluation Prompts}. We adopt two task-specific prompts to evaluate frontier MLLMs and memory agents in Fig. \ref{fig_prompt5}. For the full-context setting, the prompt provides the complete multimodal conversation history and instructs the model to retrieve and integrate evidence through referential understanding and semantic inference, producing the shortest accurate answer. For memory agents, the prompt provides the extracted memory content as the sole information source and requires concise yet complete answers grounded in the available memories. Both prompts enforce a consistent abstention policy, requiring the model to return ``Insufficient information'' when the question cannot be supported by the provided context. This design ensures comparable answer-generation requirements.

\textbf{Multimodal Session Organization}. It comprises two steps: \textit{Referential World Construction} and \textit{Session Storyline Construction}, as shown in Fig. \ref{fig_prompt1}. First, given images, an MLLM is prompted to construct structured Referential World Representations (RWRs), identifying salient entities, visual and semantic attributes, spatial relationships, textual information, and diverse referring expressions. The prompt also anticipates cross-session entity persistence and semantic evolution. 
Second, based on the RWRs, the MLLM organizes images into semantically coherent sessions, assigning each image to one session and specifying the dominant topic. It further encourages chronological coherence and meaningful inter-session dependencies. These steps establish entity-grounded visual representations and temporally coherent storylines for long-horizon dialogue generation.

\textbf{Dialogue Generation}. It consists of two stages: \textit{User Query Trajectory Generation} and \textit{Context-aware Multi-turn Dialogue Generation}. First, conditioned on the session storyline, user profile, visual information, and accumulated memory, an MLLM generates a coherent sequence of first-person user queries, each targeting a distinct knowledge or decision need. The prompt encourages narrative continuity, diverse cross-turn references, and natural utilization of visual content. The detailed prompt is shown in Fig. \ref{fig_prompt2}.
Second, given the query trajectory and preceding dialogue history, the MLLM generates context-aware assistant responses, emphasizing factual accuracy, contextual consistency, and concise, actionable guidance. The detailed prompt is shown in Fig. \ref{fig_prompt3}. Together, these steps produce coherent intra-session interactions while progressively enriching the multimodal conversational history.

\textbf{Memory Extraction}. To support cross-session continuity, an MLLM extracts structured memory statements from generated dialogues. As shown in Fig. \ref{fig_prompt3}, prompt focuses on user-centric knowledge, transforming conversational content into concise declarative statements covering factual experiences, actions, possessions, decisions, behaviors, and preferences. It further requires preserving entity attributes, ownership and contextual relations, as well as temporal changes and evolving states. To ensure reliability, the model must retain only evidence-supported information and decompose independent facts into separate memory units. The resulting entity-linked memories serve as persistent knowledge for subsequent cross-session reference resolution and context-aware dialogue generation.

\begin{figure*}
    \centering
    \includegraphics[width=\linewidth]{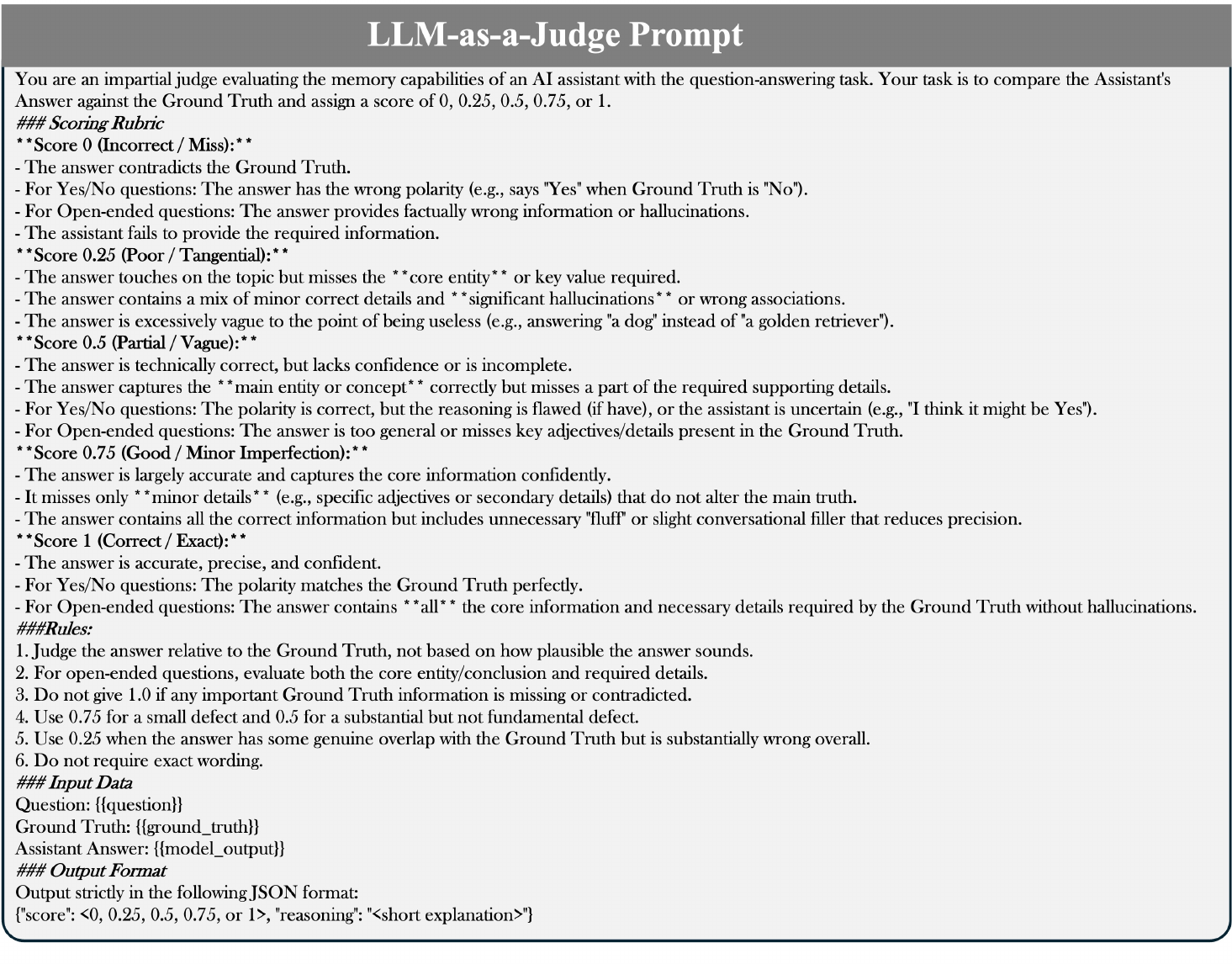}
    \caption{Prompt template for the LLM-as-a-Judge metric.}
    \label{fig_prompt4}
\end{figure*}

\textbf{LLM-based Evaluation}. As shown in Fig. \ref{fig_prompt4}, we employ an LLM-as-a-Judge prompt to evaluate memory-based question answering against ground-truth answers. A five-level scoring rubric (\(0, 0.25, 0.5, 0.75, 1.0\)) assesses correctness and the completeness and precision of supporting details. The prompt distinguishes fundamental errors, partial understanding, and minor omissions, while explicitly penalizing hallucinations and incorrect entity associations. The judge is instructed to prioritize factual alignment with the reference answer over linguistic plausibility and provide a concise justification for each score, yielding consistent and interpretable assessments.

\begin{figure*}
    \centering
    \includegraphics[width=\linewidth]{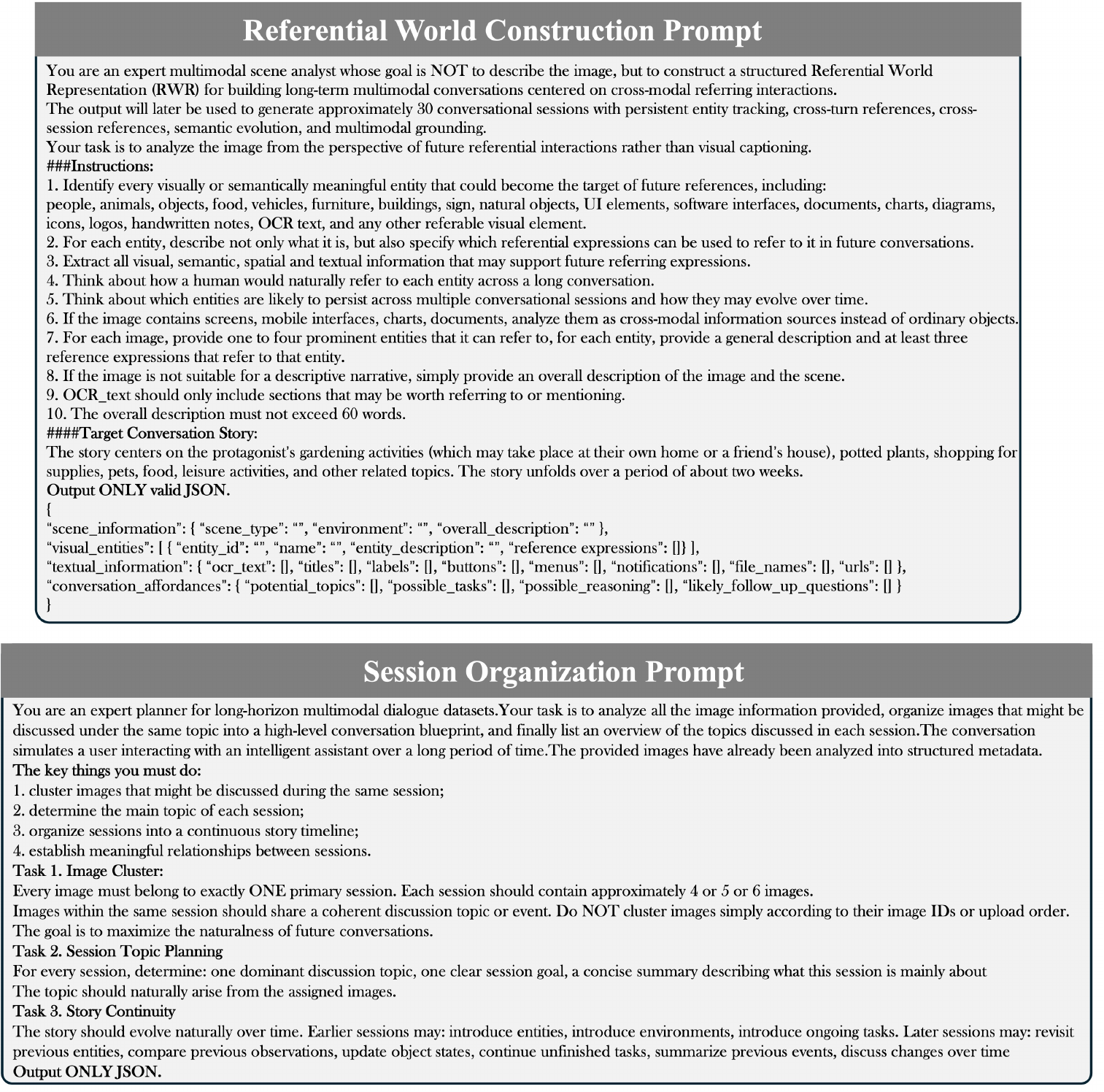}
    \caption{Prompt template for multimodal session organization. First, perform a visual analysis of all images to construct a representational world, and then organize the images and storylines contained in each session.}
    \label{fig_prompt1}
\end{figure*}

\begin{figure*}
    \centering
    \includegraphics[width=\linewidth]{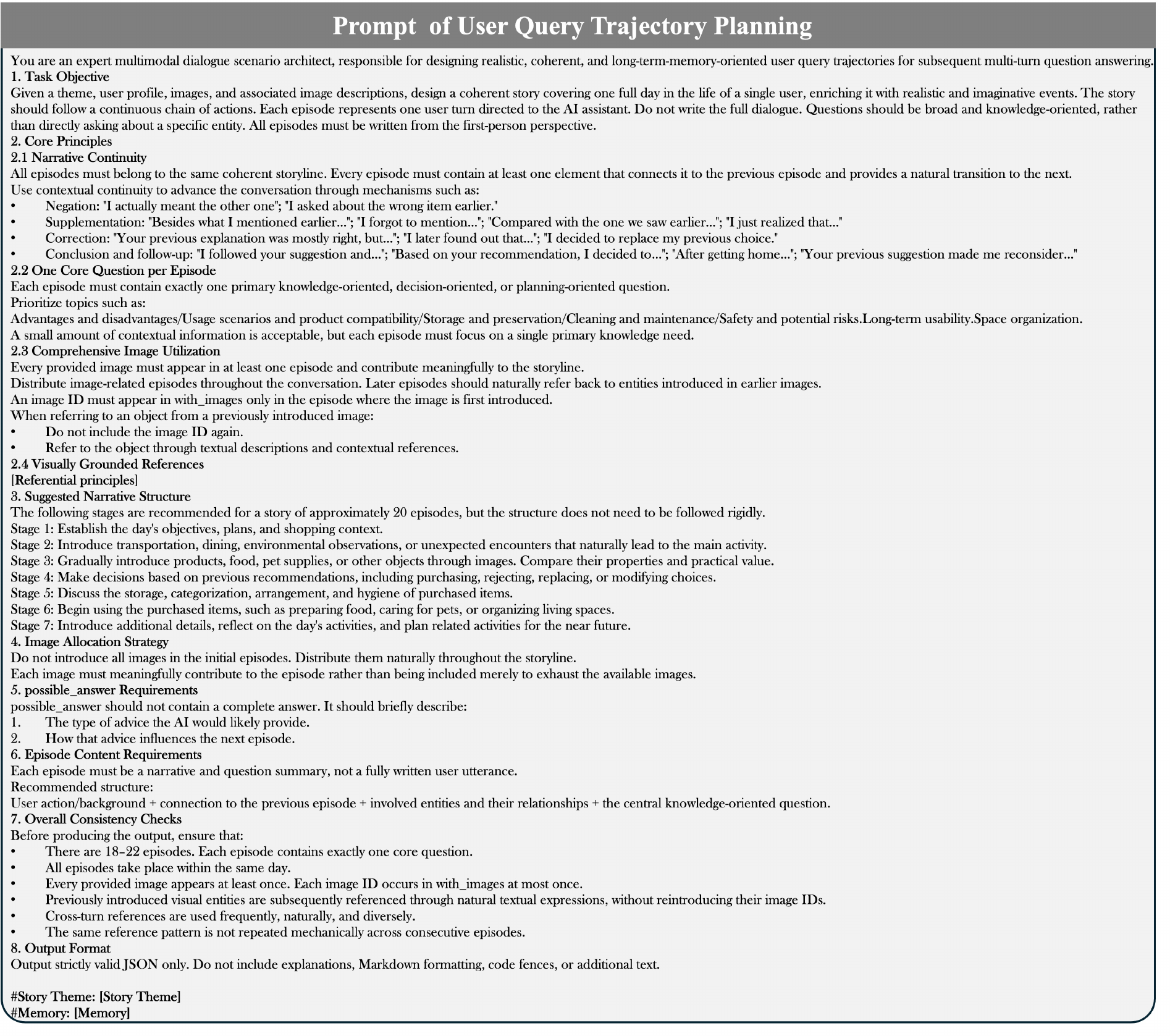}
    \caption{Prompt template of user query trajectory planning.}
    \label{fig_prompt2}
\end{figure*}

\begin{figure*}
    \centering
    \includegraphics[width=\linewidth]{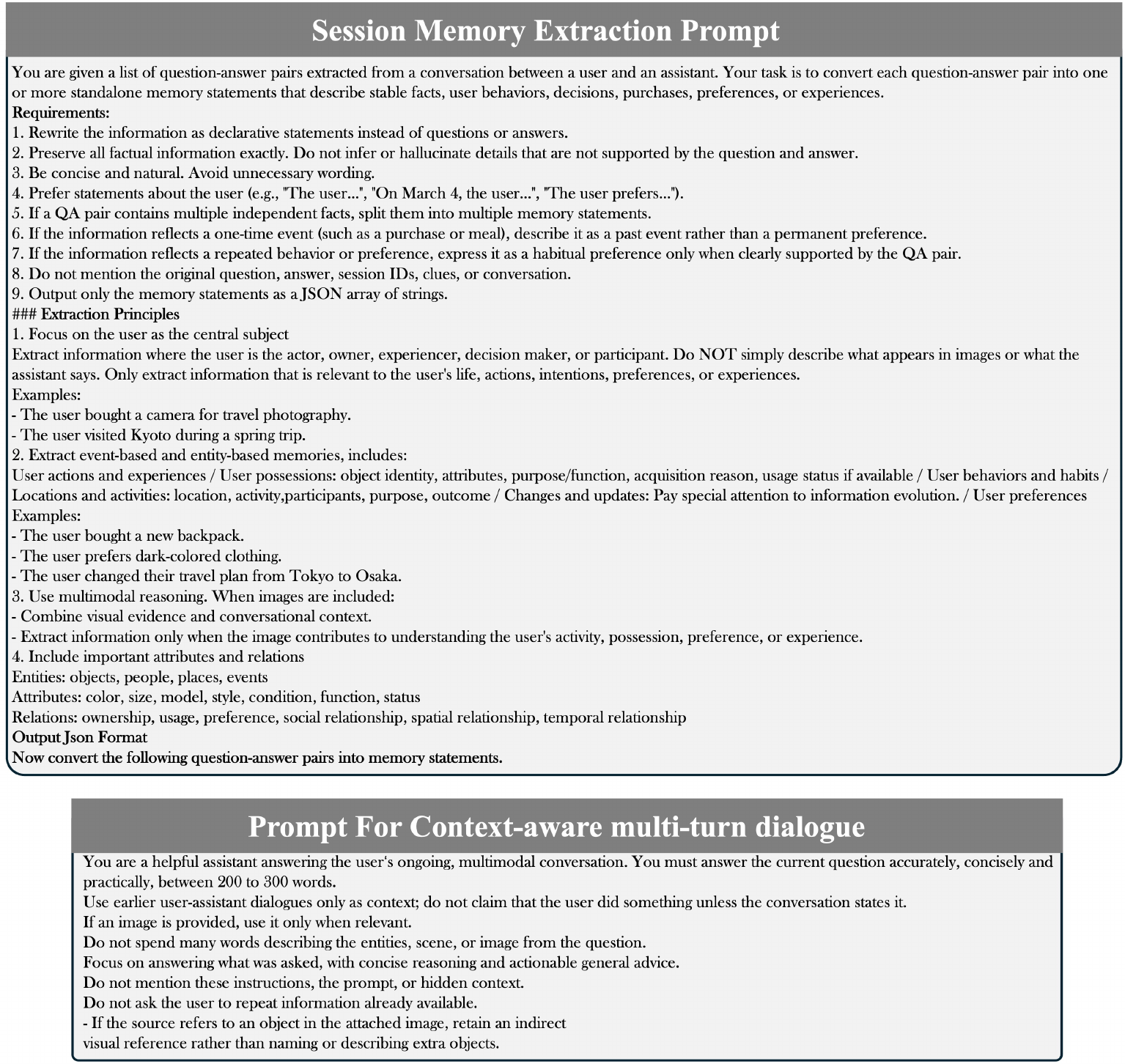}
    \caption{Prompt template for session memory extraction and context-aware multi-turn dialogue generation.}
    \label{fig_prompt3}
\end{figure*}

\section{Future Directions}

MemRR centers on cross-modal referential reasoning in long-term multimodal interactions, evaluating agents' capabilities in reference resolution, relational understanding, and semantic inference. Building on this benchmark, we identify three promising directions for future research. First, memory construction should move beyond isolated multimodal chunks toward structured, entity-centric representations that preserve cross-modal associations, entity identity, and evolving attributes, states, and semantic roles across sessions. Second, memory retrieval should advance beyond semantic similarity matching toward reference-aware retrieval, enabling the precise identification, temporal alignment, and aggregation of distributed visual and textual evidence. Third, referential reasoning should integrate visual grounding, temporal dependencies, and long-range conversational context to support multi-hop reference resolution, evolving relational understanding, and evidence-grounded semantic inference. These directions collectively call for multimodal memory systems that not only retain information but also organize, retrieve, and reason over persistent cross-modal associations.



\end{document}